\documentclass{article}

\usepackage[eandd, preprint]{neurips_2026}

\usepackage[utf8]{inputenc} %
\usepackage[T1]{fontenc}    %
\usepackage{hyperref}       %
\usepackage{url}            %
\usepackage{booktabs}       %
\usepackage{amsfonts}       %
\usepackage{nicefrac}       %
\usepackage{microtype}      %
\usepackage{xcolor}         %

\usepackage{graphicx}       %
\usepackage{subcaption}     %
\usepackage{algorithm, algorithmic}    %
\usepackage{tabularx}       %
\usepackage{tcolorbox}      %
\tcbuselibrary{breakable,skins} %
\usepackage{fvextra}        %
\usepackage{amsmath}        %
\usepackage{fontawesome5}   %
\usepackage{longtable}      %
\usepackage{soul}           %

\usepackage{amssymb}        %

  \definecolor{cHal}{RGB}{255,205,210}
  \definecolor{cAns}{RGB}{255,224,178}
  \definecolor{cDir}{RGB}{255,245,130}
  \definecolor{cAbs}{RGB}{200,230,201}
  \definecolor{cGen}{RGB}{225,190,231}
  \definecolor{cRef}{RGB}{187,222,251}
  \definecolor{cCau}{RGB}{175,238,238}
  \newcommand{\hHal}[1]{{\sethlcolor{cHal}\hl{#1}}}
  \newcommand{\hAns}[1]{{\sethlcolor{cAns}\hl{#1}}}
  \newcommand{\hDir}[1]{{\sethlcolor{cDir}\hl{#1}}}
  \newcommand{\hAbs}[1]{{\sethlcolor{cAbs}\hl{#1}}}
  \newcommand{\hGen}[1]{{\sethlcolor{cGen}\hl{#1}}}
  \newcommand{\kr}[1]{{\sethlcolor{cRef}\hl{#1}}}
  \newcommand{\hCau}[1]{{\sethlcolor{cCau}\hl{#1}}}

\newcommand{\benchmark}{\textsc{PreScience}}
\newcommand{\followupmetric}{\textsc{LACER}}

\title{\benchmark: A Dataset and Benchmark for Scientific Forecasting}

\author{
  Anirudh Ajith\textsuperscript{1,$\dagger$}\thanks{Correspondence to
    \texttt{anirudha@allenai.org}.}
  \And
  Amanpreet Singh\textsuperscript{1,$\dagger$}
  \And
  Jay DeYoung\textsuperscript{1,$\dagger$}
  \And
  Nadav Kunievsky\textsuperscript{2}
  \And
  Austin C. Kozlowski\textsuperscript{2}
  \And
  Oyvind Tafjord\textsuperscript{1}
  \And
  James Evans\textsuperscript{2}
  \And
  Daniel S. Weld\textsuperscript{1}
  \AND
  \mbox{
  Tom Hope\textsuperscript{1,3,$\dagger$} \qquad Doug Downey\textsuperscript{1,4,$\dagger$}
  }
  \\[0.75em]
  \textsuperscript{1}Allen Institute for Artificial Intelligence,
  Seattle, WA, USA
  \\
  \textsuperscript{2}Knowledge Lab, University of Chicago,
  Chicago, IL, USA
  \\
  \textsuperscript{3}School of Computer Science and Engineering,
  Hebrew University of Jerusalem, Israel
  \\
  \textsuperscript{4}Northwestern University,
  Evanston, IL, USA
  \\
  \textsuperscript{$\dagger$}Core contributor
}

\begin{document}

\maketitle

\begin{abstract}

Can AI systems trained on the existing scientific record forecast the advances that will follow?
We introduce \benchmark, a dataset and benchmark for scientific forecasting built around 98K recent AI research papers, together with companion papers covering author publication histories and citation links, yielding 502K papers in total.
The resulting paper records include titles, abstracts, disambiguated author identities, influential references, topic labels, citation trajectories, and metadata snapshotted to respect temporal cutoffs.
We instantiate seven exemplar tasks: five \textbf{paper-anchored} tasks---\textit{contribution generation}, \textit{collaborator prediction}, \textit{prior work selection}, \textit{citation count prediction}, and \textit{future combination prediction}---and two \textbf{aggregate} \textit{topic trend forecasting} variants.
We develop baselines ranging from simple heuristics and embedding methods to frontier language models and agentic systems, and introduce \followupmetric, an LLM-based metric for evaluating similarity of generated contribution descriptions that agrees better with human judgments than existing metrics.
Finally, we compose task models to generate a 12-month synthetic corpus and find that the resulting papers are systematically less diverse and less novel than human-authored research from the same period. \begin{center}
\href{https://huggingface.co/datasets/allenai/prescience}{\faDatabase\ \textbf{Dataset}}
\qquad 
\href{https://github.com/allenai/prescience}{\faGithub\ \textbf{Code}} 
\end{center}

\end{abstract}

\section{Introduction}
\label{sec:introduction}

Scientific research produces a large-scale chronological record: papers, authors, citations, and evolving topics. Predictive models trained on this record could help researchers identify relevant collaborators and prior work, anticipate emerging research directions, form hypotheses, and study how scientific interest evolves over time. Forecasting future scientific contributions also provides a natural challenge task for systems that aim to automate scientific discovery: success requires recovering the advances that were ultimately realized, given the prior art available at the time. These forecasting tasks demand integrating text, metadata, and relational structure; reasoning under temporal distribution shift; and identifying how prior work leads to new ideas and impact. Because it draws on a large historical record of science, the prediction task admits supervision at meaningful scale, offering a promising setting for evaluating and training models with deeper scientific understanding.

Previous work has examined individual science forecasting problems in isolation, including future collaborations~\citep{LibenNowell2003TheLPA,Sun2011CoauthorRPA,Kanakaris2021ShallIWA}, novel idea combinations~\citep{Sternlicht2025CHIMERAAK,Frohnert2024DiscoveringEC}, future-paper insights~\citep{heyueya2026giantsgenerativeinsightanticipation}, and publication impact~\citep{Wang2013QuantifyingLS}. However, these tasks are usually studied with disparate datasets and evaluation protocols, despite describing related facets of the same evolving scientific record. As a result, it remains difficult to compare methods across tasks, evaluate systems that use multiple kinds of scholarly evidence, or study scientific forecasting at both paper-level and aggregate scales. For large language models in particular, meaningful evaluation further requires forecasted papers to post-date model training cutoffs, a condition that many existing datasets and benchmarks do not satisfy.

\begin{figure}
    \centering
    \includegraphics[width=\linewidth]{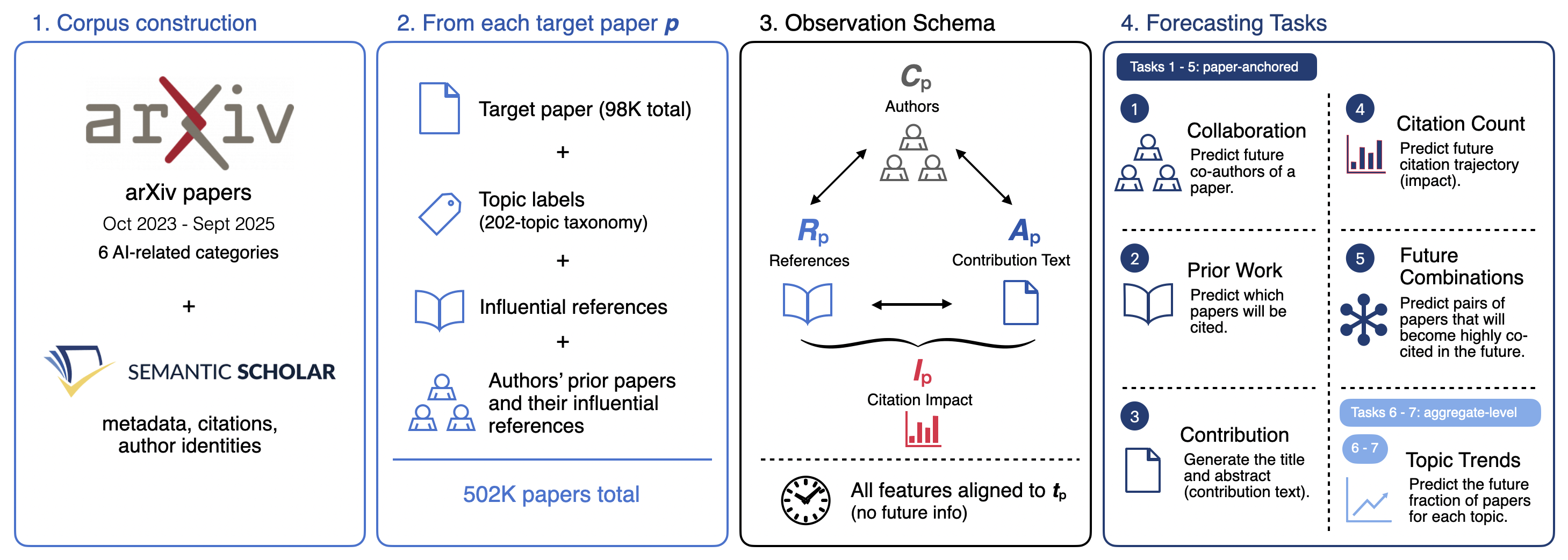}
    \caption{\benchmark~instantiates 7 tasks over a dataset built from 98K recent research papers along with author publication histories, citation links, and rich metadata spanning 502K total papers. }
    \label{fig:main_figure}
\end{figure}

We introduce \benchmark, a living benchmark for scientific forecasting built from 98K recent AI-related arXiv papers, featuring disambiguated author identities, influential references, per-paper topic labels, citation trajectories, temporally aligned scholarly metadata, and a structured companion corpus of author publication histories and citation links spanning 502K total papers.
In \benchmark, we treat each paper as an observable record of a scientific contribution, represented through a shared schema containing its authors, influential references, contribution text, citation trajectory, topics, and metadata.
This schema supports a broad family of forecasting tasks by varying which parts of the historical record are provided as inputs and which future quantities are predicted.
In this paper, we instantiate seven representative prediction problems of interest to machine learning and science-of-science: five \textbf{paper-anchored forecasting} tasks, where each prediction instance is tied to one future paper record---\textit{contribution generation}, \textit{collaborator prediction}, \textit{prior work selection}, \textit{citation count prediction}, and \textit{future combination prediction}---and two \textbf{aggregate forecasting} tasks, \textit{unary topic trend forecasting} and \textit{topic-pair trend forecasting}.

We develop and evaluate baselines spanning simple heuristics, embedding-based methods, frontier language models, and agentic systems. 
For contribution generation, we introduce an LLM-based metric called \followupmetric~that evaluates conceptual similarity between generated and ground-truth contribution descriptions, and find that \followupmetric~outperforms existing automated metrics when compared to expert human preferences, even approaching inter-annotator-agreement levels. 
Across tasks, our results show that current systems leave substantial headroom. 
Finally, we compose task models to run 12-month corpus-generation rollouts, showing that synthetic corpora generated by current methods are systematically less diverse and less novel than human-authored papers from the same period.

Our contributions are:
\begin{itemize}
    \item We introduce \benchmark, a living dataset for scientific forecasting, covering 98K recent AI-related papers and a companion corpus of prior work totaling 502K papers, with disambiguated authors and their publication histories, influential references, citation trajectories, topic labels, and temporally aligned metadata.
    \item We define seven forecasting tasks over a shared observation schema, including five paper-anchored tasks and two aggregate tasks, demonstrating the dataset's utility for controlled evaluation across multiple views of the scientific record. We benchmark a broad set of methods on these tasks: including heuristic, embedding-based, frontier LLM-based, and agentic approaches.
    \item We introduce \followupmetric, a metric for evaluating generated contribution descriptions that approaches inter-annotator-agreement level agreement with expert human preferences.
    \item We compose task-level models into corpus-level rollouts, and analyze how synthetic scientific corpora differ from real future literature, observing systematic degradations in diversity and novelty relative to human-authored research from the same period.
\end{itemize}

\section{The \benchmark~Dataset}
\label{sec:dataset}

\benchmark~is a living\footnote{Our scripts require only a date range and source arXiv categories, and can be used to periodically refresh the dataset.} dataset built from research papers posted to \href{http://info.arxiv.org/help/bulk_data/index.html}{arXiv} between October 2023 and September 2025 in six AI-related categories: \texttt{cs.CL}, \texttt{cs.LG}, \texttt{cs.AI}, \texttt{cs.CV}, \texttt{cs.IR}, and \texttt{cs.NE}. 
These 98K papers constitute the \emph{target papers} of our paper-anchored tasks.
We partition target papers into train (October 2023--September 2024) and test (October 2024--September 2025). %
To provide historical context for each target paper, we include a set of \emph{companion papers}: influential references of target papers, prior publications of target papers' authors, and influential references of those prior publications.
Each \emph{influential reference} is especially central to the citing paper's contribution \citep{ValenzuelaEscarcega2015IdentifyingMC} (see Appendix \ref{app:contribution_generation_reference_choice},~\ref{app:priorwork_reference_choice} for alternatives).
Together, target and companion papers form a 502K-paper corpus that can be used to construct the historical record $\mathbf{H}^{<t}$ available before any time $t$.
Summary statistics appear in Table~\ref{tab:dataset_stats}, with additional distributions in Appendix~\ref{app:dataset_statistics}.

\paragraph{Observation schema.}
Each target paper $p$ is represented by a temporally aligned paper record with four core \textit{components}: disambiguated authors $C_p$, influential references $R_p$, contribution text $A_p$ consisting of title and abstract, and citation trajectory $I_p$ (month over month cumulative citation counts).
The record also includes publication date $t_p$, topic labels (obtained via multilabel classification over 202 topics; Appendix~\ref{app:dataset_topics}), arXiv categories, Semantic Scholar and arXiv identifiers, and other temporally aligned metadata. Appendix~\ref{app:dataset_features} tabulates the features provided for each paper type.

\paragraph{Ensuring dataset quality.}
We take several measures to ensure that \benchmark~supports reliable modeling and evaluation. %
We source author identities and bibliographic metadata from the Semantic Scholar Graph~\citep{Wade2022TheSS}, and disambiguate author profiles using S2AND~\citep{subramanian2021s2and}.  In our manual examination, S2AND yielded more accurate author clusters than the current S2AG release. We restrict target papers to those with between 1 and 10 influential references, excluding instances with zero or unusually large influential-reference sets.
Finally, all author- and reference-level metadata (publication counts, citation counts, and $h$-indices) are temporally aligned to each paper's publication date to prevent leakage of future information into task inputs. We ensure the validity and consistency of topics assigned to papers; see Appendix~\ref{app:dataset_topics}. 

\begin{table}[tb]
\centering
\caption{Dataset statistics. Average and median statistics are computed over target papers.}
\label{tab:dataset_stats}
\small
\setlength{\tabcolsep}{3pt}

\begin{minipage}[t]{0.45\linewidth}
\vspace{0pt}
\centering
\begin{tabular}{lrrr}
\toprule
 & \textcolor{orange}{Train} & \textcolor{blue}{Test} & All \\
\midrule
Target Papers & 44{,}984 & 52{,}836 & 97{,}820 \\
All Papers & 373{,}707 & 464{,}942 & 501{,}866 \\
Avg. Words & 187.5 & 186.8 & 187.1 \\
\midrule
Avg. Infl. Refs & 3.13 & 3.04 & 3.08 \\
Med. Infl. Refs & 3 & 2 & 3 \\
\bottomrule
\end{tabular}
\end{minipage}
\hspace{0.01\linewidth}
\begin{minipage}[t]{0.50\linewidth}
\vspace{0pt}
\centering
\begin{tabular}{lrrr}
\toprule
 & \textcolor{orange}{Train} & \textcolor{blue}{Test} & All \\
\midrule
Unique Authors & 106{,}904 & 129{,}020 & 182{,}727 \\
Avg. Authors & 5.00 & 5.28 & 5.15 \\
Avg. Author Hist. & 22.5 & 27.8 & 25.5 \\
Med. Author Hist. & 7 & 9 & 8 \\
\midrule
Avg. Citations @ 12m & 5.53 & 5.77 & 5.57 \\
\bottomrule
\end{tabular}
\end{minipage}

\vspace{-0.5em}
\end{table}

\section{Scientific Forecasting Tasks} \label{sec:tasks}

The \benchmark~schema supports many forecasting tasks by varying input fields and prediction targets.
We instantiate seven representative tasks, spanning established problems from existing literature and new prediction problems enabled by the \benchmark~schema.
Together, they evaluate two scales of scientific forecasting: \textbf{paper-anchored} tasks which seek predictions for certain components of target paper records by conditioning on others (along with $\mathbf{H}^{<t_p}$), and \textbf{aggregate} tasks that predict corpus-level trends over longer time periods.
Three of our selected tasks---collaborator prediction, prior work selection, and contribution generation---can also be composed into a generative process that allows corpus-scale roll-outs of synthetic papers (as in Section~\ref{sec:corpus_generation}).

\subsection{Paper-anchored Forecasting Tasks} \label{sec:paper_level_tasks}

\paragraph{Contribution generation.}
Predicting future paper content from past literature is increasingly used to evaluate scientific ideation, hypothesis selection, and automated discovery.
Prior work studies future-aligned proposal generation from research questions and inspiring papers~\citep{wang2026learningpredictfuturealignedresearch}, or core-insight anticipation from a pair of parent papers~\citep{heyueya2026giantsgenerativeinsightanticipation}.
In \benchmark, we forecast the title and abstract $A_p$ of an actual future paper from its influential references $R_p$ and $\mathbf{H}^{<t_p}$; because $R_p$ may contain more than two prior papers, this evaluates generation from a richer evidence set tied to a ground-truth paper. %
This task requires measuring conceptual similarity rather than surface-level overlap.

We introduce \followupmetric~(Lattice of Automatically Constructed Exemplars for Reference) score, an LLM-as-judge~\citep{Zheng2023JudgingLW} metric calibrated to a 1--10 semantic-alignment scale using automatically constructed demonstrations.
These demonstrations anchor the scale between topically related prior work and semantic equivalence to the target contribution, yielding an interpretable dynamic range without human-written examples.
Against 250 expert similarity rankings, \followupmetric~approaches human inter-annotator agreement (Kendall's $\tau_b$: 0.57 vs. 0.53) and outperforms ROUGE-L~\cite{lin-2004-rouge}, BERTScore~\citep{Zhang2019BERTScoreET}, and ASPIRE Distance~\citep{mysore-etal-2022-multi} (0.27, 0.40, and 0.35); details appear in Appendix~\ref{app:followup_work_metric_validation}. We provide examples of \followupmetric~evaluations in Appendix~\ref{app:lacer_examples} along with two illustrations in Figure~\ref{fig:contribtuion_success_example}.

\paragraph{Collaborator prediction.}
Forecasting future collaborations is useful for collaborator recommendation~\citep{Zhang2023ScholarlyRS}, author discovery~\citep{Balog2012ExpertiseRetrieval}, and studying how research communities form around emerging problems~\citep{Fortunato2018ScienceOfScience}.
We formulate collaborator prediction as a link-prediction problem~\citep{LibenNowell2003TheLPA}: given a ``seed'' author (the first; Appendix~\ref{app:sub_seed_author}) of paper $p$ and $\mathbf{H}^{<t_p}$, rank candidate authors by whether they are among the remaining co-authors of $p$.
We further restrict candidate authors to those with a non-empty publication history, leaving the modeling of first-time authors to future work.
We evaluate rankings using normalized Discounted Cumulative Gain (nDCG)~\citep{Jrvelin2002CumulatedGE} and R-precision~\citep{trec_measures06}.

\paragraph{Prior work selection.}
To our knowledge, we introduce a new task: given a team of collaborators, surface prior work likely to be influential for their future paper.
Such a system could help scientific teams identify research directions to build on.
We formulate this as a team-conditioned ranking problem: given the authors $C_p$ of a target paper and $\mathbf{H}^{<t_p}$, rank prior papers by whether they appear among $p$'s influential references $R_p$.\footnote{Alternative definitions of influential prior work yield similar relative model rankings (Table~\ref{tab:key_reference_ablations}: Appendix~\ref{app:priorwork_reference_choice}).}
We evaluate rankings using nDCG and R-precision.

\paragraph{Citation count prediction.} 
Citation forecasting is useful for anticipating which papers are likely to receive early scholarly attention after publication, and can be a useful tool to researchers for prioritizing research directions.\footnote{Although citation counts are an imperfect proxy for scientific impact~\citep{Hicks2015LeidenManifesto}, they remain widely used in science-of-science research~\citep{Fortunato2018ScienceOfScience} and provide a scalable measure of early scholarly attention.}
Citation count prediction is a regression task: given $C_p$, $R_p$, $A_p$ and $\mathbf{H}^{<t_p}$, predict the number of citations $p$ will accumulate in the first 12 months after publication.
We evaluate using mean absolute error, $R^2$, Pearson correlation, and Spearman correlation.

\paragraph{Future combination prediction.}
The problem of predicting which lines of work can productively be combined to yield breakthroughs has long been of interest to the science-of-science community. 
Given a target paper $p$ from the first four months of the test period and $\mathbf{H}^{<t_p}$, we rank all prior papers $q \in \mathbf{H}^{<t_p}$ by how often $p$ and $q$ will later be jointly cited as influential references by future target papers during the eight months that follow $t_p$.
We evaluate rankings using nDCG and R-precision.

\subsection{Aggregate Forecasting}
\label{sec:aggregate_tasks}

\paragraph{Topic trend forecasting.}
Scientists and policymakers often need to track which research areas are gaining or losing attention, both individually and in combination.
This task poses the following problem: for each topic $z$, predict how much its share of papers changes from the train period to the test period.
Let $s_{\mathrm{train}}(z)$ and $s_{\mathrm{test}}(z)$ denote the fraction of evaluated papers assigned to $z$ in the train and test periods, respectively.
The prediction target is the difference between these two shares: $\Delta s(z)=s_{\mathrm{test}}(z)-s_{\mathrm{train}}(z)$.
We evaluate predictions using $R^2$ and Pearson correlation over topics.

We evaluate two variants: \textit{unary topic trends}, where each prediction unit is one of the 202 topics (Appendix~\ref{app:dataset_topics}), and \textit{topic-pair trends}, where each unit is an unordered topic pair.
For topic pairs, the target is the change in the fraction of evaluated papers assigned to both topics; we perform evaluations over pairs with at least 10 papers in either train or test, yielding 4,334 prediction units.

\section{Experiments}
\label{sec:experiments}

We provide implementation details in Appendix~\ref{app:exp_details} with computational costs appearing in Appendix~\ref{app:exp_details_compute}.
Additional per-task analyses and ablations appear in Appendix~\ref{app:analyses}.

\subsection{Contribution Generation}
\label{sec:contribution_generation_experiments}

\begin{figure}[t]
\centering
\scriptsize
\setlength{\fboxsep}{4pt}

\fbox{\begin{minipage}{0.95\linewidth}
\textbf{Influential references provided to the model} \;{\scriptsize(shared input; titles+abstracts provided; years indicate date of arXiv posting)}\\[3pt]
$\bullet$~\kr{SQuAD: 100,000+ Questions for Machine Comprehension of Text}~(2016)~\cite{rajpurkar-etal-2016-squad}\\
$\bullet$~The Curious Case of \hHal{Hallucinatory} \hAns{(Un)answerability}: Finding Truths in the \hDir{Hidden States} of Over-Confident LLMs~(2023)~\cite{slobodkin-etal-2023-curious}\\
$\bullet$~Do Large Language Models \hAns{Know What They Don't Know}?~(2023)~\cite{yin-etal-2023-large}\\
$\bullet$~The Geometry of Truth: Emergent \hDir{Linear Structure} in LLM \hDir{Representations} of True/False Datasets~(2023)~\cite{marks2024geometry}\\
$\bullet$~\kr{CREPE: Open-Domain Question Answering with False Presuppositions}~(2022)~\cite{yu-etal-2023-crepe}\\
$\bullet$~\kr{RepLiQA: A Question-Answering Dataset for Benchmarking LLMs on Unseen Reference Content}~(2024)~\cite{monteiro2024repliqa}
\end{minipage}}\\[2pt]

\parbox[t]{0.30\linewidth}{\centering$\big\downarrow$\\\emph{\scriptsize written by human authors \\ (26 Sep 2025)}}\hfill%
\parbox[t]{0.30\linewidth}{\centering$\big\downarrow$\\\emph{\scriptsize predicted by GPT-4o \\ (cutoff 01 Oct 2023)}}\hfill%
\parbox[t]{0.30\linewidth}{\centering$\big\downarrow$\\\emph{\scriptsize predicted by GPT-5 agent \\ (cutoff 30 Sep 2024)}}\\[3pt]

\fbox{\begin{minipage}[t]{0.30\linewidth}
\textbf{Ground-truth target paper}~\cite{lavi2025detecting}\\[3pt]
\textbf{Title:}~Detecting \hAns{(Un)answerability} in Large Language Models with \hDir{Linear Directions}\\[4pt]
\textbf{Abstract:}~Large language models (LLMs) often respond confidently to questions even when they lack the necessary information, leading to \hHal{hallucinated answers}. In this work, we study the problem of \hAns{(un)answerability detection}, focusing on extractive question answering (QA). We propose a simple approach for identifying \hDir{a direction in the model's activation space} that captures unanswerability and use it for classification. This direction is selected by applying activation additions during inference and measuring their impact on the model's \hAbs{abstention behavior}. We show that \hDir{projecting hidden activations onto this direction} yields a reliable score for (un)answerability classification. Experiments on two open-weight LLMs and four extractive QA benchmarks show that our method \hGen{generalizes better across datasets} than prompt-based and classifier-based approaches, and extends beyond extractive QA. Last, \hCau{causal interventions show that adding or ablating the directions effectively controls the} \hAbs{abstention behavior} of the model.
\end{minipage}}\hfill%
\fbox{\begin{minipage}[t]{0.30\linewidth}
\textbf{GPT-4o}~\colorbox{black!8}{\textbf{LACER 4/10}}\\[3pt]
\textbf{Title:}~PRESCIENCE: A Unified Framework for Handling False Presuppositions in Question Answering\\[4pt]
\textbf{Abstract:}~Despite progress in question answering (QA), addressing false presuppositions and \hAns{unanswerable questions} remains an open problem. We introduce PRESCIENCE, a unified framework that detects, handles, and corrects presupposition errors while \hAns{addressing answerability} in open-domain QA. Integrating \kr{CREPE}'s insights on false presuppositions and \kr{SQuAD}'s QA format, PRESCIENCE classifies questions as answerable, unanswerable, or containing false presuppositions. Our framework employs large language models, informed by recent studies on \hAns{query answerability} and the representation of truth in latent spaces, to detect and rectify presuppositional inaccuracies. We adapt \kr{RepLiQA} for unseen reference content to prevent data leakage. Experiments on multiple QA datasets including CREPE and SQuAD show that PRESCIENCE improves both answer accuracy and the identification of presuppositional errors.
\end{minipage}}
\hfill%
\fbox{\begin{minipage}[t]{0.30\linewidth}
\textbf{GPT-5 agent}~\colorbox{black!8}{\textbf{LACER 9/10}}\\[3pt]
\textbf{Title:}~\hAns{Answerability} \hDir{Directions}: Probing and Steering LLMs to \hAbs{Abstain} in Context-Conditional Question Answering\\[4pt]
\textbf{Abstract:}~Large language models frequently \hHal{hallucinate} in question answering, especially when questions contain false presuppositions or the answer is unsupported. We ask whether models \hAns{internally represent this context-conditional answerability}---even when they produce a confident but unsupported answer. Using \kr{SQuAD}-style answerable items, \kr{CREPE}-style presupposition failures, and \kr{RepLiQA} documents that eliminate training leakage, we show that LLMs form a robust, \hDir{approximately linear representation of answerability} expressed in the hidden state of the first decoded token. Simple \hDir{linear probes} predict answerability and presupposition failure, \hGen{transferring across datasets and models}. We propose \hAns{Answerability} \hDir{Directions} (AD): a probe-and-decoder gate that \hAbs{abstains} when the question is unsupported. We provide \hCau{causal evidence by intervening along the learned direction to shift model behavior from hallucination to} \hAbs{abstention} without fine-tuning, reducing hallucination while maintaining answer recall.
\end{minipage}}

\caption{
Examples of contribution generations from \textbf{GPT-4o} (center; the generated contribution PRESCIENCE is coincidental nomenclature and has no relation to our work) and \textbf{GPT-5 agent} (right) when provided the target's authors and influential references. Additionally, the GPT-5 agent is given leakage-safe access to $\mathbf{H}^{<t_p}$.
}
\label{fig:contribtuion_success_example}
\end{figure}

We evaluate contribution generation in the default setting: models receive titles and abstracts of a target paper's influential references $R_p$ and generate the target title and abstract (prompt in Appendix~\ref{app:prompt_contribution_generation}).
We compare frontier OpenAI and Anthropic models with finetuned OLMo~3~7B~\citep{olmo2025olmo3} and Qwen3~8B~\citep{qwen3technicalreport}, plus reference baselines: a gold paraphrase, a random influential reference from $R_p$, and a same-arXiv-category random paper.
We report \followupmetric~(using \texttt{gpt-5-2025-08-07} as judge) as the primary metric, with ROUGE-L~\citep{lin-2004-rouge} and BERTScore~\citep{Zhang2019BERTScoreET} for comparison.

\begin{table}[t]
\centering
\caption{Contribution generation results. Subscripts indicate bootstrap 95\% CIs. Unmarked errors are $0.00$. Left: model comparison using influential references as input; * indicates model cutoffs postdate test-period start (see cutoff ablation at Appendix~\ref{app:llm_cutoff_ablation}). Right: GPT-5 context ablations; \textit{citations} gives oracle access to 12-month citation counts.}
\label{tab:followup}
\footnotesize
\setlength{\tabcolsep}{3pt}

\begin{minipage}[t]{0.45\linewidth}
\vspace{0pt}
\centering
\begin{tabular}{@{}lccccc@{}}
\toprule
\textbf{Baseline} 
& \textbf{LACER}
& \multicolumn{2}{c}{\textbf{ROUGE-L}} 
& \multicolumn{2}{c}{\textbf{BERTScore}} \\
\cmidrule(lr){3-4} \cmidrule(lr){5-6}
& & \textbf{P} & \textbf{R} & \textbf{P} & \textbf{R} \\
\midrule
Same arXiv Category & $1.27_{(0.02)}$ & 0.13 & 0.12 & 0.14 & 0.13 \\
Random Infl. Ref. & $4.31_{(0.06)}$ & 0.16 & 0.16 & 0.19 & 0.18 \\
\midrule
OLMo 3 7B (FT) & $4.03_{(0.05)}$ & 0.19 & 0.17 & 0.24 & 0.19 \\
Qwen 3 8B (FT) & $3.99_{(0.05)}$ & \textbf{0.20} & 0.16 & 0.24 & 0.18 \\
\midrule
GPT 4o & $4.71_{(0.06)}$ & 0.17 & 0.16 & \textbf{0.25} & 0.23 \\
GPT o3 & $5.49_{(0.06)}$ & 0.12 & 0.16 & 0.15 & 0.22 \\
GPT 5 & $\mathbf{5.64}_{(0.06)}$ & 0.11 & 0.16 & 0.14 & 0.21 \\
GPT 5.2* & $5.60_{(0.06)}$ & 0.13 & 0.16 & 0.17 & 0.22 \\
Claude Sonnet 4.5* & $5.03_{(0.06)}$ & 0.14 & \textbf{0.18} & 0.21 & \textbf{0.24} \\
Claude Opus 4.5* & $5.04_{(0.06)}$ & 0.13 & 0.14 & 0.19 & 0.19 \\
\midrule
Gold Paraphrase & $10.00_{(0.00)}$ & 0.61 & 0.56 & 0.71 & 0.70 \\
\bottomrule
\end{tabular}
\end{minipage}
\hfill
\begin{minipage}[t]{0.45\linewidth}
\vspace{0pt}
\centering
\begin{tabular}{@{}lc@{}}
\toprule
\textbf{Input features} & \textbf{LACER} \\
\midrule
Infl. refs. & $5.64_{(0.06)}$ \\
Infl. refs. + related & $5.59_{(0.06)}$ \\
Infl. refs. + author papers & $5.72_{(0.06)}$ \\
Infl. refs. + citations & $5.37_{(0.06)}$ \\
Infl. refs. + related + author & $5.74_{(0.06)}$ \\
Infl. refs. + related + author + citations & $5.61_{(0.06)}$ \\
\midrule
Infl. refs. + $\mathbf{H}^{<t_p}$ (Agent) & $\mathbf{5.86}_{(0.06)}$ \\
\bottomrule
\end{tabular}
\end{minipage}
\vspace{-0.5em}
\end{table}

\paragraph{Results.}
Table~\ref{tab:followup} (left) shows that gold paraphrases reach the top of the \followupmetric~scale, while random same-category papers and influential references score much lower.
Finetuned open models improve over the same-category baseline but score lower than frontier models; even the best frontier model reaches only moderate similarity to the ground truth, reflecting the difficulty of forecasting specific future contributions.
Our ablations indicate that model pretraining overlap with the test period does not substantially affect \followupmetric~scores (Appendix~\ref{app:llm_cutoff_ablation}), and that relative rankings remain similar across \followupmetric~judge choices (Appendix~\ref{app:followup_work_metric_lacerscore_judges}). 

\paragraph{Context ablations.}
Table~\ref{tab:followup} (right) evaluates GPT-5 with richer inputs on the full test split.
All rows include influential references; related papers are retrieved from $\mathbf{H}^{<t_p}$ using the papers in $R_p$ as nearest-neighbor queries, and the \textit{citations} setting assumes oracle access to 12-month citation count.
Author histories provide a reliable improvement ($5.64 \rightarrow 5.72$), while related papers do not seem to help significantly, and oracle citation counts hurt clearly ($5.64 \rightarrow 5.37$), suggesting that scalar citation counts are not useful to GPT-5 in this setup.
An agentic variant ($\sim3\times$ in API costs compared to simple GPT-5 prompting) equipped with tool-access to the full historical record $\mathbf{H}^{<t_p}$ performs best ($5.86$), indicating that broader context can improve predictions. The agent baseline's implementation details appear in Appendix~\ref{app:exp_details_contribution_generation}. Figure~\ref{fig:contribtuion_success_example} shows an example of scored generations from GPT-4o and the GPT-5-based agent.

\subsection{Paper-anchored Ranking Tasks}

We evaluate three paper-anchored ranking tasks: collaborator prediction, prior work selection, and future combination prediction.
For collaborator prediction, models rank candidate authors given a seed author and $\mathbf{H}^{<t_p}$; for the other two tasks, models rank candidate papers from $\mathbf{H}^{<t_p}$.
We report nDCG@1000 and R-precision, using GRIT~\citep{muennighoff2025generative} embeddings in the main table and  GTR~\citep{ni-etal-2022-large} and Specter2~\citep{singh-etal-2023-scirepeval} variants in Appendices~\ref{app:collaborator_embedding_choice}, \ref{app:priorwork_embedding_choice}, and \ref{app:cocitation_embedding_choice}.

\paragraph{Baselines.}
For collaborator prediction and prior work selection, we evaluate frequency heuristics and embedding-based retrieval baselines.
\textit{Frequency} ranks candidates by historical co-authorship frequency with the seed author for collaborator prediction, and by previous citations from the target authors for prior work selection.
\textit{Rank Fusion} retrieves papers from $\mathbf{H}^{<t_p}$ using query embeddings from the seed author or author set, then ranks candidates by aggregated ranks.
\textit{Embedding Fusion} ranks authors or papers by similarity in embedding space.
\textit{Hierarchical Clustering} represents each author using multiple centroids to model heterogeneous interests, while \textit{Projection} learns a task-specific mapping over frozen embeddings using a Multi-Instance NCE objective~\citep{Miech2019EndtoEndLO}. 
For future combination prediction, \textit{Target Similarity} ranks prior papers by similarity to the target title and abstract; \textit{Reference Similarity} uses the mean embedding of the target paper's influential references; \textit{Citation Frequency} ranks papers by prior frequency as influential references; and \textit{Co-citation Frequency} ranks papers by prior influential co-citation with the target paper's influential references.
We also evaluate two agentic baselines: \textit{Autoresearch}, adapted from Karpathy's autoresearch framework~\citep{autoresearch}, which iteratively designs and tests ranking strategies under a 12-hour budget, and \textit{GPT-5 Agent}, which uses tools for paper search, paper inspection, author histories, references, and citation traversal, with all outputs filtered to $\mathbf{H}^{<t_p}$.
Implementation details for the static baselines, Autoresearch, and the GPT-5 agent appear in Appendices~\ref{app:exp_details_collaborator_prediction}, \ref{app:exp_details_prior_work_selection}, and \ref{app:exp_details_future_influential_co-citation}.

\begin{table*}[t]
\centering
\caption{Paper-anchored ranking task results. Subscripts indicate bootstrap 95\% CIs. Left: collaborator prediction and prior work selection. Right: future combination prediction. $\dagger$ indicates evaluation on $n=500$ random instances.}
\label{tab:paper_level_ranking}
\footnotesize
\setlength{\tabcolsep}{3pt}

\begin{minipage}[t]{0.50\linewidth}
\vspace{0pt}
\centering
\begin{tabular}{lcccc}
\toprule
 & \multicolumn{2}{c}{\textbf{Collab}} 
 & \multicolumn{2}{c}{\textbf{Prior Work}} \\
\cmidrule(lr){2-3} \cmidrule(lr){4-5}
\textbf{Baseline} & \textbf{nDCG} & \textbf{R-Prec} & \textbf{nDCG} & \textbf{R-Prec} \\
\midrule
Frequency & $0.40_{(0.01)}$ & $0.28_{(0.01)}$ & $0.10_{(0.01)}$ & $0.05_{(0.00)}$ \\
\midrule
Rank Fusion & $0.17_{(0.00)}$ & $0.08_{(0.00)}$ & $0.02_{(0.00)}$ & $0.01_{(0.00)}$ \\
Emb. Fusion & $0.28_{(0.00)}$ & $0.18_{(0.00)}$ & $0.11_{(0.04)}$ & $0.05_{(0.03)}$ \\
Hier. Clustering & $0.25_{(0.00)}$ & $0.15_{(0.00)}$ & $0.06_{(0.00)}$ & $0.02_{(0.00)}$ \\
\midrule
Emb. Fus. Refs & -- & -- & $0.06_{(0.00)}$ & $0.02_{(0.00)}$ \\
Emb. Fus Proj. & $0.24_{(0.01)}$ & $0.14_{(0.01)}$ & $0.13_{(0.01)}$ & $0.05_{(0.00)}$ \\
\midrule
Autoresearch & $\mathbf{0.49}_{(0.01)}$ & $\mathbf{0.34}_{(0.01)}$ & $\mathbf{0.15}_{(0.01)}$ & $0.06_{(0.00)}$ \\
GPT-5 Agent$^{\dagger}$ & $0.45_{(0.03)}$ & $0.32_{(0.03)}$ & $\mathbf{0.15}_{(0.02)}$ & $\mathbf{0.07}_{(0.02)}$ \\
\bottomrule
\end{tabular}
\end{minipage}
\hfill
\begin{minipage}[t]{0.44\linewidth}
\vspace{0pt}
\centering
\begin{tabular}{lcc}
\toprule
 & \multicolumn{2}{c}{\textbf{Future Comb.}} \\
\cmidrule(lr){2-3}
\textbf{Baseline} & \textbf{nDCG} & \textbf{R-Prec} \\
\midrule
Target Similarity & $0.26_{(0.01)}$ & $0.11_{(0.01)}$ \\
Ref. Similarity & $0.30_{(0.01)}$ & $0.16_{(0.01)}$ \\
Co-citation Freq. & $0.12_{(0.01)}$ & $0.05_{(0.01)}$ \\
Citation Freq. & $0.06_{(0.00)}$ & $0.01_{(0.00)}$ \\
\midrule
Autoresearch & $\mathbf{0.38}_{(0.01)}$ & $\mathbf{0.19}_{(0.01)}$ \\
GPT-5 Agent$^{\dagger}$ & $0.29_{(0.02)}$ & $0.17_{(0.02)}$ \\
\bottomrule
\end{tabular}
\end{minipage}

\vspace{-0.5em}
\end{table*}

\paragraph{Results.}
Table~\ref{tab:paper_level_ranking} shows that agentic methods perform best across all three ranking tasks.
For collaborator prediction, \textit{Autoresearch} improves over the strongest static baseline, \textit{Frequency}, from $0.40/0.28$ to $0.49/0.34$ nDCG@1000/R-precision.
For prior work selection, gains are smaller: \textit{Autoresearch} reaches $0.15/0.06$, only slightly above \textit{Embedding Fusion Proj.} ($0.13/0.05$), indicating that identifying influential references remains difficult.
For future combination prediction, \textit{Autoresearch} performs best ($0.38/0.19$), while \textit{Reference Similarity} is the strongest non-agentic baseline ($0.30/0.16$), suggesting that influential references provide useful signal for future combinations.

\subsection{Citation Count Prediction}
\label{sec:experiments_impact}

We evaluate citation forecasting with XGBoost regressors over three feature families: \textit{Target Text} (target title and abstract), \textit{Context Text} (influential references and author prior publications), and \textit{Bibliometrics} (reference citation counts and author $h$-index, total citations, and publication counts at publication time).
Text features use GRIT embeddings in the main table; GTR and Specter2 variants appear in Appendix~\ref{app:impact_embedding_choice}.
We predict 12-month citation counts and report metrics in both raw and log citation space. Similar to the previous tasks, we apply \textit{Autoresearch} to this task as well.

\begin{table*}[t]
\centering
\footnotesize
\caption{Citation count prediction results using GRIT embeddings.}
\label{tab:impact}
\setlength{\tabcolsep}{6pt}
\begin{tabular}{lccccc}
\toprule
\textbf{Baseline} 
& \textbf{MAE} 
& \textbf{MAE (log)} 
& \textbf{Pearson} 
& \textbf{Pearson (log)} 
& \textbf{Spearman} \\
\midrule
Target Text                 & $4.67$ & $0.71$ & $0.29$ & $0.49$ & $0.46$ \\
Bibliometrics                   & $4.79$ & $0.74$ & $0.36$ & $0.42$ & $0.37$ \\
\midrule
Target + Context               & $4.58$ & $0.69$ & $0.28$ & $0.54$ & $0.50$ \\
Target + Context + Bibliometrics & $4.52$ & $\mathbf{0.68}$ & $0.31$ & $0.56$ & $0.51$ \\
\midrule
Autoresearch & $\mathbf{4.31}$ & $\mathbf{0.68}$ & $\mathbf{0.44}$ & $\mathbf{0.57}$ & $\mathbf{0.53}$ \\
\bottomrule
\end{tabular}
\end{table*}

\paragraph{Results.}
Table~\ref{tab:impact} shows that adding \textit{Context Text} to \textit{Target Text} improves performance, and adding \textit{Bibliometrics} yields smaller further gains.
\textit{Autoresearch} performs best overall, reaching $4.31$ MAE, $0.44$ Pearson, and $0.53$ Spearman in raw citation space.
Prediction error remains substantial, reflecting the heavy-tailed nature of citation outcomes; SHAP analyses of bibliometric features and prediction scatter plots appear in Appendix~\ref{app:impact_prediction_analysis}. Implementation details appear in Appendix~\ref{app:exp_details_citation_count}.

\subsection{Topic Trend Forecasting}
\label{sec:experiments_topic_trends}

\paragraph{Baselines.}
We report simple extrapolative and momentum-based baselines as an initial benchmark for this new aggregate forecasting task. 
\textit{Mean} predicts the same average change for every topic.
\textit{Linear Extrapolation} fits a linear trend to monthly historical counts and extrapolates over the forecast horizon.
\textit{Citation Momentum} scores each topic by summing the observable citations of its historical papers, then converts the score to a paper-count forecast via a topic-agnostic linear scale calibrated on an inner train/test split of the history period.
\textit{Author Momentum} uses the same inner-split calibration but replaces citations with per-author metrics --- prior publication counts, $h$-indices, or total citations --- summed across each paper's authors.
Richer learned time-series forecasting methods are a natural direction for future work.

\begin{table}[t]
\centering
\caption{Topic trend forecasting results. We report $R^2$ and Pearson correlation for predicted changes ($\Delta$) in topic share. Unary trends are evaluated over 202 topics; topic-pair trends are evaluated over 4,334 topic pairs with at least 10 papers in either train or test.}
\label{tab:topic_trends}
\small
\setlength{\tabcolsep}{5pt}
\begin{tabular}{lrrrr}
\toprule
\textbf{Baseline} 
& \multicolumn{2}{c}{\textbf{Unary topics}} 
& \multicolumn{2}{c}{\textbf{Topic pairs}} \\
\cmidrule(lr){2-3} \cmidrule(lr){4-5}
& $\Delta$share $R^2$ & $\Delta$share Pearson 
& $\Delta$share $R^2$ & $\Delta$share Pearson \\
\midrule
Mean & $-11.29$ & $0.14$ & $-5.48$ & $0.25$ \\
Linear Extrapolation & $0.15$ & $0.52$ & $\mathbf{0.19}$ & $\mathbf{0.57}$ \\
Citation Momentum & $-2.37$ & $\mathbf{0.61}$ & $-2.74$ & $0.43$ \\
Author Momentum (papers) & $-0.05$ & $0.42$ & $0.00$ & $0.42$ \\
Author Momentum ($h$-index) & $\mathbf{0.17}$ & $0.50$ & $0.16$ & $0.47$ \\
\bottomrule
\end{tabular}
\end{table}

\paragraph{Results.}
Table~\ref{tab:topic_trends} shows that simple historical signals are predictive but incomplete.
For unary topics, author momentum based on $h$-index achieves the best $R^2$ ($0.17$), while citation momentum gives the highest Pearson correlation ($0.61$).
For topic pairs, linear extrapolation performs best on both metrics ($0.19$ $R^2$, $0.57$ Pearson).
Negative $R^2$ values for mean and citation momentum show that correlation alone does not imply calibrated share-change predictions.

\subsection{Corpus Generation}
\label{sec:corpus_generation}

The preceding experiments evaluate task models in isolation. 
We also compose them into corpus-level rollouts that synthesize a 12-month stream of future paper records. 
Starting from the literature state at the beginning of the test period ($\mathbf{H}^{<t_0}$, where $t_0=\text{October 1, 2024}$), the rollout proceeds day by day: sample the number of papers from a train-period daily publication distribution $P_{\text{daily}}$, sample team sizes, predict collaborators, select influential references, generate titles and abstracts, and fold generated papers back into the literature state for subsequent steps. 
This rollout policy is a modeling choice for corpus generation, not part of the benchmark definition; full pseudocode appears in Appendix~\ref{app:corpus_generation_algorithm}. 
We use one-shot \textit{GRIT + Embedding Fusion} for collaborator prediction and prior work selection, and GPT-5 for contribution generation.

\begin{figure*}[t]
    \centering
    \includegraphics[width=\linewidth]{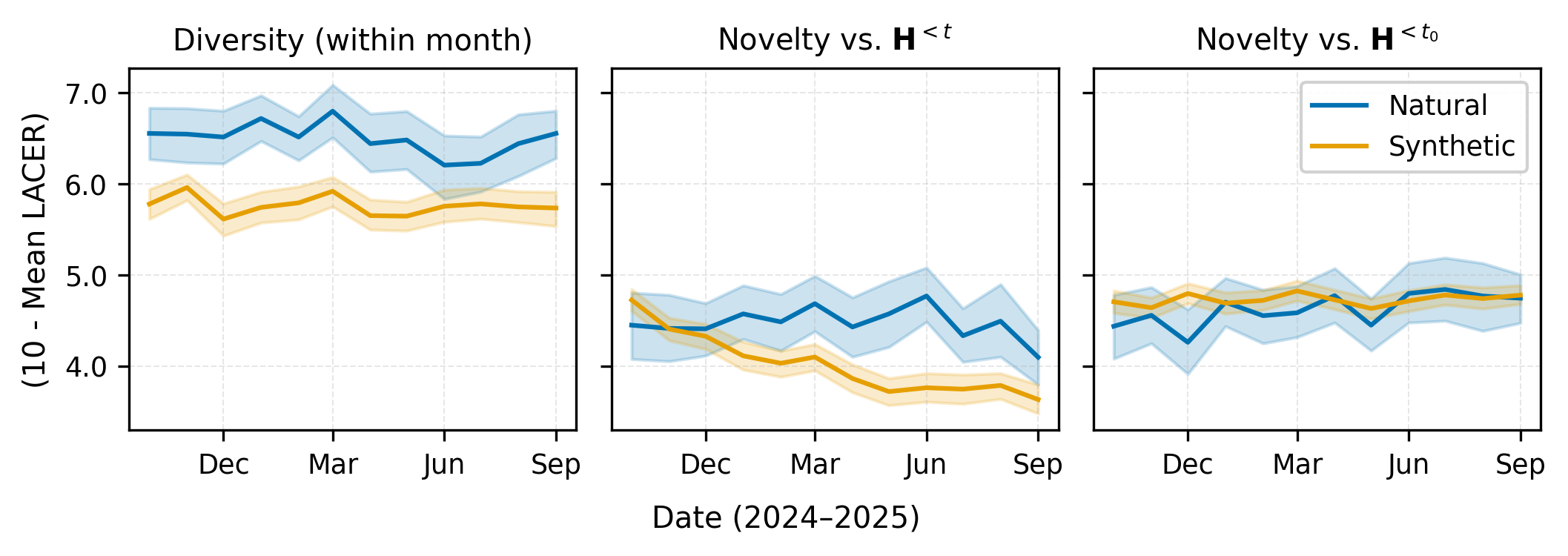}
    \caption{Corpus-generation rollouts produce synthetic papers that are less diverse than real future papers (left) and become less novel relative to the evolving literature state $\mathbf{H}^{<t}$ (middle). When novelty is measured against the fixed pre-rollout corpus $\mathbf{H}^{<t_0}$ (right), this decline largely disappears. Shading represents bootstrap 95\% CIs.}
    \label{fig:multiturn_figures_lacer_diversity_and_novelty}
\end{figure*}

\paragraph{Evaluation.}
We compare synthetic and real future corpora using \followupmetric-based diversity and novelty. 
For each month, we sample $n=100$ generated papers and retrieve their $k=10$ nearest neighbors in GRIT embedding space. 
Diversity uses same-month papers as the retrieval pool; novelty uses either the evolving literature state $\mathbf{H}^{<t}$ or the fixed pre-rollout corpus $\mathbf{H}^{<t_0}$. 
We size-match real-paper retrieval pools to synthetic pools, repeat the rollout six times, and report mean trends.

\paragraph{Results.}
Synthetic corpora are consistently less diverse than real papers from the same period and become less novel relative to the evolving literature state $\mathbf{H}^{<t}$ (Figure~\ref{fig:multiturn_figures_lacer_diversity_and_novelty}). 
This novelty decline largely disappears when measured against the fixed pre-rollout corpus $\mathbf{H}^{<t_0}$, suggesting that generations remain distant from the original historical corpus but become increasingly similar to earlier synthetic outputs. 
Predicted author and prior-work sets are more diverse than their real-world counterparts (Appendix~\ref{app:sub_corpus_gen_analysis}), suggesting that the diversity gap arises primarily from contribution generation rather than predicted collaborators or references.

\section{Related Work}
\label{sec:related_work}

\subsection{Forecasting Scientific Advances}

Several recent benchmarks evaluate predictive models against future scientific advances. A few forecast concept co-occurrence in future papers, albeit at different scales: dynamic word embeddings trained on 67k quantum-physics papers  \citep{Frohnert2024DiscoveringEC}, a knowledge graph built from 21M papers \citep{Gu2024ForecastingHR}, and 65k research sub-fields (1827--2024) from OpenAlex \citep{rezaee2025fos}, respectively.
Proof of Time~\citep{ye2026proofoftime} was created to evaluate agents on 4 disparate tasks - impact prediction, author research evolution, peer-review awards and benchmark performance. ScienceMeter~\citep{wang2025sciencemeter} measures \emph{knowledge projection}, the ability of an updated model to anticipate scientific claims in post-cutoff papers. 
Closer to our contribution-generation task, \citep{wang2026learningpredictfuturealignedresearch} train language models to produce future-aligned proposals scored by their Future Alignment Score, while \citep{heyueya2026giantsgenerativeinsightanticipation} introduce GiantsBench, a benchmark of parent-papers to downstream-insight pairs scored by an LLM judge.
In contrast, \benchmark~defines multiple paper-anchored and aggregate forecasting tasks under a shared, temporally aligned observation schema, and can compose task models into corpus-generation rollouts.

\paragraph{Collaborator prediction}
This task has been well-studied, with most efforts using graph-based modeling approaches~\citep{Kanakaris2021ShallIWA, Xi2021RecommendationOA, Tuninetti2021PredictionON, Ebrahimi2021DevelopingAM, Ho2019CoauthorRP, Li2024NovSRCAN}. Some methods explore alternative representations, such as authors conditioned on a research topic~\citep{DBLP:journals/apin/ChuanSAKHD18, Cheng2023KnowledgeGE}, or the temporal nature of their publication histories~\citep{OMadadhain2005PredictionAR, Munasinghe2012TimeSA, Koopmann2021CoBERTSC}. See \citep{Kong2019AcademicSN, Zhang2023ScholarlyRS} for surveys touching on author link prediction.

\paragraph{Prior work selection}
For a given set of authors, we forecast the literature they will build upon for creating a new advance. To our knowledge, this team-conditioned formulation is novel. In loosely related lines of work on scientific ideation, it is common to retrieve past papers to serve as inspirations~\citep{Wang2023SciMONSI, Chen2025StructuringSI, Radensky2024ScideatorHS, Sternlicht2025CHIMERAAK}; however, the objective in those works is ideation-focused rather than to forecast the choice of prior work conditioned on the collaborating authors and their expertise.

\paragraph{Contribution generation}
The closest analogy to this task in current literature is scientific hypothesis generation or ideation, which can be grounded in the interactions between different areas of the scientific literature~\citep{Swanson1986UndiscoveredPK}. Many modern approaches build on this insight~\citep{Wang2023SciMONSI, Radensky2024ScideatorHS, Lu2024TheAS, Wang2024SciPIPAL, baek-etal-2025-researchagent}, including recent benchmarks mentioned earlier \citep{wang2026learningpredictfuturealignedresearch, heyueya2026giantsgenerativeinsightanticipation, Guo2025IdeaBenchBL} and multi-agent systems ~\citep{su-etal-2025-many}.

\paragraph{Topic-trend forecasting}
Knowledge-graph approaches have been previously used to detect, analyze, and forecast topics from large-scale scholarly graphs~\citep{salatino2021detectionanalysisprediction}, and topic popularity has been predicted five years in advance from publication, language, and patent signals~\citep{ofer2023whatsnext}. Closer to our setting, link prediction over concept knowledge graphs forecasts future co-investigation at scale~\citep{Gu2024ForecastingHR}, and dynamic word embeddings can capture emergent cross-area connections~\citep{Frohnert2024DiscoveringEC}.

\paragraph{Citation count and future combination prediction}
Prior works have used varied measures for impact, ranging from citation accumulation~\citep{Uddin2013NetworkEO, Gu2024ForecastingHR}, novelty~\citep{shi2023surprising, zhang2025language}, to research grant success~\citep{Cole1981ChanceAC, Boyack2018TowardPR, Gyrffy2020ResearchFP}. Beyond predicting the citation count of a paper itself, \benchmark's future influential co-citation prediction task asks which prior works will appear as influential references jointly with a target paper. This sits in the same tradition of forecasting future conceptual connections---via knowledge-graph link prediction~\citep{Gu2024ForecastingHR} and dynamic concept embeddings~\citep{Frohnert2024DiscoveringEC}---but is evaluated against an explicit influential-reference signal from real future papers. \benchmark~conditions the citation predictions on the full generative context: the research team, their prior work, and the contribution itself. Our finding that author and reference features provide substantial predictive power aligns with the ``cumulative advantage'' literature~\citep{merton1968matthew, Wang2013QuantifyingLS}, while the residual variance points to other potentially helpful signals that remain unexplored.

\subsection{Generating Synthetic Scientific Corpora}

A complementary line of work generates scientific outputs at scale such as ideas, hypotheses, papers, or full research workflows with LLMs and compares these to empirical research. Some include a human in the loop~\citep{Cappello2025EAIRAEA, Jansen2024DISCOVERYWORLDAV, Jansen2025CodeScientistES}, while others operate fully autonomously~\citep{Lu2024TheAS, majumder2025discoverybench}.
Multi-agent simulators give agents specialized roles, access to relevant literature, and the ability to interact in lab-like setups, producing synthetic research artifacts as their primary output~\citep{su-etal-2025-many, Swanson2024TheVL, Chen2025BeyondBW, Yu2024ResearchTownSO}. AgentRxiv~\citep{Schmidgall2025AgentRxivTC} extends this further by letting agent labs share reports through a common preprint server, accumulating a synthetic record over time.
Comparing the resulting corpora against real research is itself an established framing: \citep{amancio2015comparingreal} contrast artificially generated and real manuscripts via topological features, and recent studies of LLM idea generation systems report limited diversity in generated ideas relative to human-authored ones~\citep{Si2024CanLG, hu2024nova}. 
Instead of solely focusing on ideation, \benchmark~ allows composing task-level forecasters for 12-month corpus-generation rollouts and quantifies how the resulting synthetic corpora differ from real future papers in diversity and novelty.

\section{Limitations}
\label{sec:limitations}

As a first step toward building a holistic benchmark for a complex, co-evolving real-world process, our work has several limitations. 
Our benchmark is limited to research from the AI domain and focuses on seven tasks from among many possibilities, one of which is predicting citations which is a limited signal of scientific impact. 
There exist variables (e.g. conferences, grants, institutions) we do not include in our dataset that influence the arc of science. 
Further, the methods evaluated in this paper are a partial set of all available techniques.  We hope that our openly released benchmark and dataset construction infrastructure facilitates addressing these limitations in future work.

\section{Conclusion}
\label{sec:conclusion}

We presented \benchmark, a living dataset and benchmark for scientific forecasting over a shared, temporally controlled corpus of recent AI research. \benchmark~provides an observation schema that supports multiple forecasting tasks over the same scientific record, rather than prescribing a single model of how science unfolds. We instantiate seven tasks spanning paper-level forecasting and aggregate topic trends, introduce \followupmetric~for contribution generation, and benchmark a broad set of heuristic, embedding-based, language-model and agentic baselines. Across tasks, current methods leave substantial headroom, and our corpus-generation rollouts show that today's composed systems produce synthetic corpora that are less diverse and less novel than real future literature. Future releases can extend \benchmark~to additional domains and richer observables, including institutions, funding, venues, peer-review signals, and multimodal paper content.

\paragraph{Broader Impact} We hope that \benchmark~  will spur the development of agents that can anticipate scientific activity and outcomes, helping scientists, policymakers, and others identify emerging opportunities and prioritize research directions. A potential risk of the work is that erroneous predictive outputs will be misused for decision-making. We emphasize that use of models trained and evaluated on \benchmark~carefully consider the limitations discussed in Section \ref{sec:limitations}.

\section*{Acknowledgements}
This work was supported in part by NSF Grant 2404109. We would also like to thank the Semantic Scholar team, UChicago APTO group, Sewon Min, and other members of Ai2 for their feedback and support.

\bibliography{sci_predict}

@string{acl = {Association for Computational Linguistics}}

@string{anth = {https://aclanthology.org/}}

@string{NAACL:2022:main = {Proceedings of the 2022 Conference of the North American Chapter of the Association for Computational Linguistics: Human Language Technologies}}

@string{EMNLP:2023:main = {Proceedings of the 2023 Conference on Empirical Methods in Natural Language Processing}}

@string{EMNLP:2022:main = {Proceedings of the 2022 Conference on Empirical Methods in Natural Language Processing}}

@string{NAACL:2025:long = {Proceedings of the 2025 Conference of the Nations of the Americas Chapter of the Association for Computational Linguistics: Human Language Technologies (Volume 1: Long Papers)}}

@inproceedings{lavi2025detecting,
    title = "Detecting (Un)answerability in Large Language Models with Linear Directions",
    author = "Lavi, Maor Juliet and Milo, Tova and Geva, Mor",
    editor = "Demberg, Vera and Inui, Kentaro and Marquez, Llu{\'i}s",
    booktitle = "Proceedings of the 19th Conference of the {E}uropean Chapter of the {A}ssociation for {C}omputational {L}inguistics (Volume 1: Long Papers)",
    month = mar,
    year = "2026",
    address = "Rabat, Morocco",
    publisher = "Association for Computational Linguistics",
    pages = "682--699",
    doi = "10.18653/v1/2026.eacl-long.29",
    url = "https://aclanthology.org/2026.eacl-long.29/",
    note = "arXiv:2509.22449"
}

@inproceedings{rajpurkar-etal-2016-squad,
    title = "{SQ}u{AD}: 100,000+ Questions for Machine Comprehension of Text",
    author = "Rajpurkar, Pranav  and
      Zhang, Jian  and
      Lopyrev, Konstantin  and
      Liang, Percy",
    editor = "Su, Jian  and
      Duh, Kevin  and
      Carreras, Xavier",
    booktitle = "Proceedings of the 2016 Conference on Empirical Methods in Natural Language Processing",
    month = nov,
    year = "2016",
    address = "Austin, Texas",
    publisher = "Association for Computational Linguistics",
    url = "https://aclanthology.org/D16-1264/",
    doi = "10.18653/v1/D16-1264",
    pages = "2383--2392"
}

@inproceedings{slobodkin-etal-2023-curious,
    title = "The Curious Case of Hallucinatory (Un)answerability: Finding Truths in the Hidden States of Over-Confident Large Language Models",
    author = "Slobodkin, Aviv  and
      Goldman, Omer  and
      Caciularu, Avi  and
      Dagan, Ido  and
      Ravfogel, Shauli",
    editor = "Bouamor, Houda  and
      Pino, Juan  and
      Bali, Kalika",
    booktitle = "Proceedings of the 2023 Conference on Empirical Methods in Natural Language Processing",
    month = dec,
    year = "2023",
    address = "Singapore",
    publisher = "Association for Computational Linguistics",
    url = "https://aclanthology.org/2023.emnlp-main.220/",
    doi = "10.18653/v1/2023.emnlp-main.220",
    pages = "3607--3625"
}

@inproceedings{yin-etal-2023-large,
    title = "Do Large Language Models Know What They Don't Know?",
    author = "Yin, Zhangyue  and
      Sun, Qiushi  and
      Guo, Qipeng  and
      Wu, Jiawen  and
      Qiu, Xipeng  and
      Huang, Xuanjing",
    editor = "Rogers, Anna  and
      Boyd-Graber, Jordan  and
      Okazaki, Naoaki",
    booktitle = "Findings of the Association for Computational Linguistics: ACL 2023",
    month = jul,
    year = "2023",
    address = "Toronto, Canada",
    publisher = "Association for Computational Linguistics",
    url = "https://aclanthology.org/2023.findings-acl.551/",
    doi = "10.18653/v1/2023.findings-acl.551",
    pages = "8653--8665"
}

@inproceedings{marks2024geometry,
    title = {The Geometry of Truth: Emergent Linear Structure in Large Language Model Representations of True/False Datasets},
    author = {Marks, Samuel and Tegmark, Max},
    booktitle = {First Conference on Language Modeling (COLM)},
    year = {2024},
    url = {https://arxiv.org/abs/2310.06824}
}

@inproceedings{yu-etal-2023-crepe,
    title = "{CREPE}: Open-Domain Question Answering with False Presuppositions",
    author = "Yu, Xinyan  and
      Min, Sewon  and
      Zettlemoyer, Luke  and
      Hajishirzi, Hannaneh",
    editor = "Rogers, Anna  and
      Boyd-Graber, Jordan  and
      Okazaki, Naoaki",
    booktitle = "Proceedings of the 61st Annual Meeting of the Association for Computational Linguistics (Volume 1: Long Papers)",
    month = jul,
    year = "2023",
    address = "Toronto, Canada",
    publisher = "Association for Computational Linguistics",
    url = "https://aclanthology.org/2023.acl-long.583/",
    doi = "10.18653/v1/2023.acl-long.583",
    pages = "10457--10480"
}

@inproceedings{monteiro2024repliqa,
    title = {RepLiQA: A Question-Answering Dataset for Benchmarking LLMs on Unseen Reference Content},
    author = {Monteiro, Joao and No{\"e}l, Pierre-Andr{\'e} and Marcotte, {\'E}tienne and Rajeswar, Sai and Zantedeschi, Valentina and V{\'a}zquez, David and Chapados, Nicolas and Pal, Christopher and Taslakian, Perouz},
    booktitle = {Advances in Neural Information Processing Systems (NeurIPS) Datasets and Benchmarks Track},
    year = {2024},
    url = {https://arxiv.org/abs/2406.11811}
}

@misc{autoresearch,
  title = {autoresearch},
  author = {Karpathy, Andrej},
  year = {2026},
  url = {https://github.com/karpathy/autoresearch},
  note = {https://github.com/karpathy/autoresearch. Accessed 2026-05-06}
}

@misc{wang2026learningpredictfuturealignedresearch,
      title={Learning to Predict Future-Aligned Research Proposals with Language Models}, 
      author={Heng Wang and Pengcheng Jiang and Jiashuo Sun and Zhiyi Shi and Haofei Yu and Jiawei Han and Heng Ji},
      year={2026},
      eprint={2603.27146},
      archivePrefix={arXiv},
      primaryClass={cs.CL},
      url={https://arxiv.org/abs/2603.27146}, 
}

@misc{heyueya2026giantsgenerativeinsightanticipation,
      title={GIANTS: Generative Insight Anticipation from Scientific Literature}, 
      author={Joy He-Yueya and Anikait Singh and Ge Gao and Michael Y. Li and Sherry Yang and Chelsea Finn and Emma Brunskill and Noah D. Goodman},
      year={2026},
      eprint={2604.09793},
      archivePrefix={arXiv},
      primaryClass={cs.CL},
      url={https://arxiv.org/abs/2604.09793}, 
}

@inproceedings{lin-2004-rouge,
    title = "{ROUGE}: A Package for Automatic Evaluation of Summaries",
    author = "Lin, Chin-Yew",
    booktitle = "Text Summarization Branches Out",
    month = jul,
    year = "2004",
    address = "Barcelona, Spain",
    publisher = "Association for Computational Linguistics",
    url = "https://aclanthology.org/W04-1013/",
    pages = "74--81"
}

@inproceedings{subramanian2021s2and,
      title={{S}2{AND}: {A} {B}enchmark and {E}valuation {S}ystem for {A}uthor {N}ame {D}isambiguation}, 
      author={Subramanian, Shivashankar and King, Daniel and Downey, Doug and Feldman, Sergey},
      year={2021},
      publisher = {Association for Computing Machinery},
      address = {New York, NY, USA},
      booktitle = {{JCDL} '21: Proceedings of the {ACM/IEEE} Joint Conference on Digital Libraries in 2021},
      series = {JCDL '21}
}

@misc{Sternlicht2025CHIMERAAK,
      title={CHIMERA: A Knowledge Base of Scientific Idea Recombinations for Research Analysis and Ideation}, 
      author={Noy Sternlicht and Tom Hope},
      year={2025},
      eprint={2505.20779},
      archivePrefix={arXiv},
      primaryClass={cs.CL},
      url={https://arxiv.org/abs/2505.20779}, 
}

@article{Frohnert2024DiscoveringEC,
  title={Discovering emergent connections in quantum physics research via dynamic word embeddings},
  author={Felix Frohnert and Xuemei Gu and Mario Krenn and Evert P L van Nieuwenburg},
  journal={Machine Learning: Science and Technology},
  year={2025},
  volume={6},
  url={https://api.semanticscholar.org/CorpusID:273963065}
}

@inproceedings{Wang2023SciMONSI,
  title={SciMON: Scientific Inspiration Machines Optimized for Novelty},
  author={Qingyun Wang and Doug Downey and Heng Ji and Tom Hope},
  booktitle={Annual Meeting of the Association for Computational Linguistics},
  year={2024},
  url={https://api.semanticscholar.org/CorpusID:258841365}
}

@article{Chen2025StructuringSI,
  title={Structuring Scientific Innovation: A Framework for Modeling and Discovering Impactful Knowledge Combinations},
  author={Junlan Chen and Kexin Zhang and Daifeng Li and Yangyang Feng and Yuxuan Zhang and Bowen Deng},
  journal={ArXiv},
  year={2025},
  volume={abs/2503.18865},
  url={https://api.semanticscholar.org/CorpusID:277313413}
}

@article{Kanakaris2021ShallIWA,
  title={Shall I Work with Them? A Knowledge Graph-Based Approach for Predicting Future Research Collaborations},
  author={Nikos Kanakaris and Nikolaos Giarelis and Ilias Siachos and N. Karacapilidis},
  journal={Entropy},
  year={2021},
  volume={23},
  url={https://api.semanticscholar.org/CorpusId:235301976}
}

@article{Sun2011CoauthorRPA,
  title={Co-author Relationship Prediction in Heterogeneous Bibliographic Networks},
  author={Yizhou Sun and R. Barber and Manish Gupta and C. Aggarwal and Jiawei Han},
  journal={2011 International Conference on Advances in Social Networks Analysis and Mining},
  year={2011},
  pages={121-128},
  url={http://ieeexplore.ieee.org/stamp/stamp.jsp?tp=\&arnumber=5992571}
}

@inproceedings{LibenNowell2003TheLPA,
  title={The link prediction problem for social networks},
  author={David Liben-Nowell and Jon M. Kleinberg},
  booktitle={International Conference on Information and Knowledge Management},
  year={2003},
  url={http://dl.acm.org/citation.cfm?id=956972}
}

@article{Radensky2024ScideatorHS,
  title={Scideator: Human-LLM Scientific Idea Generation Grounded in Research-Paper Facet Recombination},
  author={Marissa Radensky and Simra Shahid and Raymond Fok and Pao Siangliulue and Tom Hope and Daniel S. Weld},
  journal={ArXiv},
  year={2024},
  volume={abs/2409.14634},
  url={https://api.semanticscholar.org/CorpusID:272827497}
}

@inproceedings{Jansen2025CodeScientistES,
  title={CodeScientist: End-to-End Semi-Automated Scientific Discovery with Code-based Experimentation},
  author={Peter Jansen and Oyvind Tafjord and Marissa Radensky and Pao Siangliulue and Tom Hope and Bhavana Dalvi Mishra and Bodhisattwa Prasad Majumder and Daniel S. Weld and Peter Clark},
  editor={Che, Wanxiang and Nabende, Joyce and Shutova, Ekaterina and Pilehvar, Mohammad Taher},
  booktitle={Findings of the Association for Computational Linguistics: ACL 2025},
  month={jul},
  year={2025},
  address={Vienna, Austria},
  publisher={Association for Computational Linguistics},
  pages={13370--13467},
  doi={10.18653/v1/2025.findings-acl.692},
  url={https://aclanthology.org/2025.findings-acl.692/}
}

@article{Cheng2023KnowledgeGE,
  title={Knowledge Graph-based Embedding for Connecting Scholars in Academic Social Networks},
  author={Xiyao Cheng and Yuanxun Zhang and Harsh Joshi and Mayank Kejriwal and Prasad Calyam},
  journal={2023 IEEE 10th International Conference on Data Science and Advanced Analytics (DSAA)},
  year={2023},
  pages={1-10},
  url={https://api.semanticscholar.org/CorpusID:265054862}
}

@inproceedings{Xi2021RecommendationOA,
  title={Recommendation of Academic Collaborators: A Methodology Incorporating Word Embedding and Network Embedding},
  author={Xiaowen Xi and Ying Guo and Weiyu Duan},
  booktitle={AII@iConference},
  year={2021},
  url={https://api.semanticscholar.org/CorpusID:235259334}
}

@article{Koopmann2021CoBERTSC,
  title={CoBERT: Scientific Collaboration Prediction via Sequential Recommendation},
  author={Tobias Koopmann and Konstantin Kobs and Konstantin Herud and Andreas Hotho},
  journal={2021 International Conference on Data Mining Workshops (ICDMW)},
  year={2021},
  pages={45-54},
  url={https://api.semanticscholar.org/CorpusID:246081502}
}

@article{Zhang2023ScholarlyRS,
  title={Scholarly recommendation systems: a literature survey},
  author={Zitong Zhang and Braja Gopal Patra and Ashraf Yaseen and Jie Zhu and Rachit Sabharwal and Kirk Roberts and Tru Hoang Cao and Hulin Wu},
  journal={Knowledge and Information Systems},
  year={2023},
  volume={65},
  pages={4433-4478},
  url={https://api.semanticscholar.org/CorpusID:259081885}
}

@article{Munasinghe2012TimeSA,
  title={Time Score: A New Feature for Link Prediction in Social Networks},
  author={Lankeshwara Munasinghe and Ryutaro Ichise},
  journal={IEICE Trans. Inf. Syst.},
  year={2012},
  volume={95-D},
  pages={821-828},
  url={https://api.semanticscholar.org/CorpusID:30012200}
}

@article{OMadadhain2005PredictionAR,
  title={Prediction and ranking algorithms for event-based network data},
  author={Joshua O'Madadhain and Jon Hutchins and Padhraic Smyth},
  journal={SIGKDD Explor.},
  year={2005},
  volume={7},
  pages={23-30},
  url={https://api.semanticscholar.org/CorpusID:3343116}
}

@article{DBLP:journals/apin/ChuanSAKHD18,
  author       = {Pham Minh Chuan and
                  Le Hoang Son and
                  Mumtaz Ali and
                  Tran Dinh Khang and
                  Le Thanh Huong and
                  Nilanjan Dey},
  title        = {Link prediction in co-authorship networks based on hybrid content
                  similarity metric},
  journal      = {Appl. Intell.},
  volume       = {48},
  number       = {8},
  pages        = {2470--2486},
  year         = {2018},
  url          = {https://doi.org/10.1007/s10489-017-1086-x},
  doi          = {10.1007/S10489-017-1086-X},
  bibsource    = {dblp computer science bibliography, https://dblp.org}
}

@article{Uddin2013NetworkEO,
  title={Network Effects on Scientific Collaborations},
  author={Shahadat Uddin and Liaquat Hossain and Kim J. R. Rasmussen},
  journal={PLoS ONE},
  year={2013},
  volume={8},
  url={https://api.semanticscholar.org/CorpusID:7633781}
}

@article{Tuninetti2021PredictionON,
  title={Prediction of new scientific collaborations through multiplex networks},
  author={Marta Tuninetti and Alberto Aleta and Daniela Paolotti and Yamir Moreno and Michele Starnini},
  journal={EPJ Data Science},
  year={2021},
  volume={10},
  url={https://api.semanticscholar.org/CorpusID:234489207}
}

@article{Kong2019AcademicSN,
  title={Academic social networks: Modeling, analysis, mining and applications},
  author={Xiangjie Kong and Yajie Shi and Shuo Yu and Jiaying Liu and Feng Xia},
  journal={J. Netw. Comput. Appl.},
  year={2019},
  volume={132},
  pages={86-103},
  url={https://api.semanticscholar.org/CorpusID:86850665}
}

@article{Lu2024TheAS,
  title={The AI Scientist: Towards Fully Automated Open-Ended Scientific Discovery},
  author={Chris Lu and Cong Lu and Robert Tjarko Lange and Jakob Nicolaus Foerster and Jeff Clune and David Ha},
  journal={ArXiv},
  year={2024},
  volume={abs/2408.06292},
  url={https://api.semanticscholar.org/CorpusID:271854887}
}

@article{Jrvelin2002CumulatedGE,
  title={Cumulated gain-based evaluation of IR techniques},
  author={Kalervo J{\"a}rvelin and Jaana Kek{\"a}l{\"a}inen},
  journal={ACM Trans. Inf. Syst.},
  year={2002},
  volume={20},
  pages={422-446},
  url={https://api.semanticscholar.org/CorpusID:1981391}
}

@inproceedings{ValenzuelaEscarcega2015IdentifyingMC,
  title={Identifying Meaningful Citations},
  author={Marco Antonio Valenzuela-Escarcega and Vu A. Ha and Oren Etzioni},
  booktitle={AAAI Workshop: Scholarly Big Data},
  year={2015},
  url={https://api.semanticscholar.org/CorpusID:2538517}
}

@inproceedings{Zhang2019BERTScoreET,
  title     = {{BERTScore}: Evaluating Text Generation with {BERT}},
  author    = {Tianyi Zhang and Varsha Kishore and Felix Wu and Kilian Q. Weinberger and Yoav Artzi},
  booktitle = {8th International Conference on Learning Representations (ICLR 2020), Addis Ababa, Ethiopia, April 26--30, 2020},
  year      = {2020},
  publisher = {OpenReview.net},
  url       = {https://openreview.net/forum?id=SkeHuCVFDr}
}

@inproceedings{Pramanick2024TheNO,
  title={The Nature of {NLP}: Analyzing Contributions in {NLP} Papers},
  author={Aniket Pramanick and Yufang Hou and Saif M. Mohammad and Iryna Gurevych},
  editor={Che, Wanxiang and Nabende, Joyce and Shutova, Ekaterina and Pilehvar, Mohammad Taher},
  booktitle={Proceedings of the 63rd Annual Meeting of the Association for Computational Linguistics (Volume 1: Long Papers)},
  month={jul},
  year={2025},
  address={Vienna, Austria},
  publisher={Association for Computational Linguistics},
  pages={25169--25191},
  doi={10.18653/v1/2025.acl-long.1224},
  url={https://aclanthology.org/2025.acl-long.1224/}
}

@article{Kendall1938ANM,
  title={A NEW MEASURE OF RANK CORRELATION},
  author={M. G. Kendall},
  journal={Biometrika},
  year={1938},
  volume={30},
  pages={81-93},
  url={https://api.semanticscholar.org/CorpusID:120478295}
}

@inproceedings{Zheng2023JudgingLW,
  title={Judging {LLM}-as-a-Judge with {MT}-Bench and Chatbot Arena},
  author={Zheng, Lianmin and Chiang, Wei-Lin and Sheng, Ying and Zhuang, Siyuan and Wu, Zhanghao and Zhuang, Yonghao and Lin, Zi and Li, Zhuohan and Li, Dacheng and Xing, Eric P. and Zhang, Hao and Gonzalez, Joseph E. and Stoica, Ion},
  booktitle={Advances in Neural Information Processing Systems 36 (NeurIPS 2023) Datasets and Benchmarks Track},
  volume={36},
  year={2023},
  url={https://proceedings.neurips.cc/paper_files/paper/2023/hash/91f18a1287b398d378ef22505bf41832-Abstract-Datasets_and_Benchmarks.html},
  note={arXiv:2306.05685}
}

@article{Swanson1986UndiscoveredPK,
  title={Undiscovered Public Knowledge},
  author={Don R. Swanson},
  journal={The Library Quarterly},
  year={1986},
  volume={56},
  pages={103 - 118},
  url={https://api.semanticscholar.org/CorpusID:267792818}
}

@article{Wang2024SciPIPAL,
  title={SciPIP: An LLM-based Scientific Paper Idea Proposer},
  author={Wenxiao Wang and Lihui Gu and Liye Zhang and Yunxiang Luo and Yi Dai and Chen Shen and Liang Xie and Binbin Lin and Xiaofei He and Jieping Ye},
  journal={ArXiv},
  year={2024},
  volume={abs/2410.23166},
  url={https://api.semanticscholar.org/CorpusID:273695165}
}

@article{Boyack2018TowardPR,
  title={Toward predicting research proposal success},
  author={Kevin W. Boyack and Caleb Smith and Richard Klavans},
  journal={Scientometrics},
  year={2018},
  volume={114},
  pages={449-461},
  url={https://api.semanticscholar.org/CorpusID:46804654}
}

@article{Cole1981ChanceAC,
  title={Chance and consensus in peer review.},
  author={Stephen Cole and Jonathan R. Cole and Gary A. Simon},
  journal={Science},
  year={1981},
  volume={214 4523},
  pages={
          881-6
        },
  url={https://api.semanticscholar.org/CorpusID:11183533}
}

@article{Gyrffy2020ResearchFP,
  title={Research funding: past performance is a stronger predictor of future scientific output than reviewer scores},
  author={Bal{\'a}zs Győrffy and P{\'e}ter Herman and Istv{\'a}n Szab{\'o}},
  journal={J. Informetrics},
  year={2020},
  volume={14},
  pages={101050},
  url={https://api.semanticscholar.org/CorpusID:219933512}
}

@inproceedings{
majumder2025discoverybench,
title={DiscoveryBench: Towards Data-Driven Discovery with Large Language Models},
author={Bodhisattwa Prasad Majumder and Harshit Surana and Dhruv Agarwal and Bhavana Dalvi Mishra and Abhijeetsingh Meena and Aryan Prakhar and Tirth Vora and Tushar Khot and Ashish Sabharwal and Peter Clark},
booktitle={The Thirteenth International Conference on Learning Representations},
year={2025},
url={https://openreview.net/forum?id=vyflgpwfJW}
}

@inproceedings{Jansen2024DISCOVERYWORLDAV,
  title={DiscoveryWorld: A Virtual Environment for Developing and Evaluating Automated Scientific Discovery Agents},
  author={Peter Jansen and Marc-Alexandre C{\^o}t{\'e} and Tushar Khot and Erin Bransom and Bhavana Dalvi Mishra and Bodhisattwa Prasad Majumder and Oyvind Tafjord and Peter Clark},
  booktitle={Advances in Neural Information Processing Systems 37: Annual Conference on Neural Information Processing Systems 2024, NeurIPS 2024, Datasets and Benchmarks Track, Vancouver, BC, Canada, December 10--15, 2024},
  editor={Amir Globerson and Lester Mackey and Danielle Belgrave and Angela Fan and Ulrich Paquet and Jakub M. Tomczak and Cheng Zhang},
  year={2024},
  url={http://papers.nips.cc/paper_files/paper/2024/hash/13836f251823945316ae067350a5c366-Abstract-Datasets_and_Benchmarks_Track.html}
}

@article{Cappello2025EAIRAEA,
  title={EAIRA: Establishing a Methodology for Evaluating AI Models as Scientific Research Assistants},
  author={Franck Cappello and Sandeep Madireddy and Robert Underwood and Neil Getty and Nicholas Chia and Nesar Ramachandra and Josh Nguyen and Murat Keceli and Tanwi Mallick and Zilinghan Li and Marie Claver Nd{\'e}bane Ngom and Chenhui Zhang and Angel Yanguas-Gil and Evan R. Antoniuk and Bhavya Kailkhura and Minyang Tian and Yu Du and Yuan-Sen Ting and Azton Wells and Bogdan Nicolae and Avinash Maurya and M. Mustafa Rafique and E. A. Huerta and Bo Li and Ian Foster and Rick Stevens},
  journal={ArXiv},
  year={2025},
  volume={abs/2502.20309},
  url={https://api.semanticscholar.org/CorpusID:276647576}
}

@article{Swanson2024TheVL,
  title={The Virtual Lab of AI agents designs new SARS-CoV-2 nanobodies},
  author={Swanson, Kyle and Wu, Wesley and Bulaong, Nash L. and Pak, John E. and Zou, James},
  journal={Nature},
  volume={646},
  number={8085},
  pages={716--723},
  year={2025},
  publisher={Springer Science and Business Media LLC},
  doi={10.1038/s41586-025-09442-9},
  url={https://doi.org/10.1038/s41586-025-09442-9}
}

@misc{Chen2025BeyondBW,
  title={Beyond Brainstorming: What Drives High-Quality Scientific Ideas? Lessons from Multi-Agent Collaboration},
  author={Nuo Chen and Yicheng Tong and Jiaying Wu and Minh Duc Duong and Qian Wang and Qingyun Zou and Bryan Hooi and Bingsheng He},
  year={2025},
  url={https://api.semanticscholar.org/CorpusID:280540858}
}

@inproceedings{Yu2024ResearchTownSO,
  title={ResearchTown: Simulator of Human Research Community},
  author={Yu, Haofei and Hong, Zhaochen and Cheng, Zirui and Zhu, Kunlun and Xuan, Keyang and Yao, Jinwei and Feng, Tao and You, Jiaxuan},
  booktitle={Proceedings of the 42nd International Conference on Machine Learning},
  series={Proceedings of Machine Learning Research},
  volume={267},
  pages={73051--73096},
  year={2025},
  publisher={PMLR},
  url={https://proceedings.mlr.press/v267/yu25i.html},
  note={arXiv:2412.17767}
}

@article{Schmidgall2025AgentRxivTC,
  title={AgentRxiv: Towards Collaborative Autonomous Research},
  author={Samuel Schmidgall and Michael Moor},
  journal={ArXiv},
  year={2025},
  volume={abs/2503.18102},
  url={https://api.semanticscholar.org/CorpusID:277271871}
}

@article{Ebrahimi2021DevelopingAM,
  title={Developing a mathematical model of the co-author recommender system using graph mining techniques and big data applications},
  author={Fezzeh Ebrahimi and Asefeh Asemi and Amin Nezarat and Andrea Ko},
  journal={Journal of Big Data},
  year={2021},
  volume={8},
  url={https://api.semanticscholar.org/CorpusID:232133644}
}

@article{Ho2019CoauthorRP,
  title={Co-author Relationship Prediction in Bibliographic Network: A New Approach Using Geographic Factor and Latent Topic Information},
  author={Thi Kim Thoa Ho and Quang Vu Bui and Marc Bui},
  journal={Proceedings of the 10th International Symposium on Information and Communication Technology},
  year={2019},
  url={https://api.semanticscholar.org/CorpusID:209450869}
}

@article{Li2024NovSRCAN,
  title={NovSRC: A Novelty-Oriented Scientific Collaborators Recommendation Model},
  author={Xiuxiu Li and Mingyang Wang and Chaoran Wang and Yujia Fu and Xianjie Wang},
  journal={International Journal of Advanced Computer Science and Applications},
  year={2024},
  url={https://api.semanticscholar.org/CorpusID:268818672}
}

@article{Gu2024ForecastingHR,
  title={Forecasting high-impact research topics via machine learning on evolving knowledge graphs},
  author={Xuemei Gu and Mario Krenn},
  journal={Machine Learning: Science and Technology},
  year={2025},
  volume={6},
  url={https://api.semanticscholar.org/CorpusID:267636723}
}

@inproceedings{
muennighoff2025generative,
title={Generative Representational Instruction Tuning},
author={Niklas Muennighoff and Hongjin SU and Liang Wang and Nan Yang and Furu Wei and Tao Yu and Amanpreet Singh and Douwe Kiela},
booktitle={The Thirteenth International Conference on Learning Representations},
year={2025},
url={https://openreview.net/forum?id=BC4lIvfSzv}
}

@article{Si2024CanLG,
  title={Can LLMs Generate Novel Research Ideas? A Large-Scale Human Study with 100+ NLP Researchers},
  author={Chenglei Si and Diyi Yang and Tatsunori Hashimoto},
  journal={ArXiv},
  year={2024},
  volume={abs/2409.04109},
  url={https://api.semanticscholar.org/CorpusID:272463952}
}

@article{Wade2022TheSS,
  title={The Semantic Scholar Academic Graph (S2AG)},
  author={Alex D Wade},
  journal={Companion Proceedings of the Web Conference 2022},
  year={2022},
  url={https://api.semanticscholar.org/CorpusID:251597885}
}

@article{merton1968matthew,
   title={The Matthew Effect in Science},
   author={Merton, Robert K.},
   journal={Science},
   volume={159},
   number={3810},
   pages={56--63},
   year={1968}
}

@article{shi2023surprising,
  title={Surprising combinations of research contents and contexts are related to impact and emerge with scientific outsiders from distant disciplines},
  author={Shi, Feng and Evans, James},
  journal={Nature Communications},
  volume={14},
  number={1},
  pages={1641},
  year={2023},
  publisher={Nature Publishing Group UK London}
}

@article{zhang2025language,
  title={Language Model Perplexity Predicts Scientific Surprise and Transformative Impact},
  author={Zhang, Zhen and Evans, James},
  journal={arXiv preprint arXiv:2509.05591},
  year={2025}
}

@misc{trec_measures06,
  title        = {Common Evaluation Measures},
  author       = {{NIST TREC}},
  year         = {2006},
  note         = {Appendix CE.MEASURES06},
  howpublished = {TREC 2006 Proceedings (Appendix)},
  url          = {https://trec.nist.gov/pubs/trec15/appendices/CE.MEASURES06.pdf}
}

@article{Miech2019EndtoEndLO,
  title={End-to-End Learning of Visual Representations From Uncurated Instructional Videos},
  author={Antoine Miech and Jean-Baptiste Alayrac and Lucas Smaira and Ivan Laptev and Josef Sivic and Andrew Zisserman},
  journal={2020 IEEE/CVF Conference on Computer Vision and Pattern Recognition (CVPR)},
  year={2020},
  pages={9876-9886},
  url={https://api.semanticscholar.org/CorpusID:209370497}
}

@article{Guo2025IdeaBenchBL,
  title={IdeaBench: Benchmarking Large Language Models for Research Idea Generation},
  author={Sikun Guo and Amir Hassan Shariatmadari and Guangzhi Xiong and Albert Huang and Eric Xie and Stefan Bekiranov and Aidong Zhang},
  journal={Proceedings of the 31st ACM SIGKDD Conference on Knowledge Discovery and Data Mining V.2},
  year={2025},
  url={https://api.semanticscholar.org/CorpusID:273821733}
}

@inproceedings{su-etal-2025-many,
    title = "Many Heads Are Better Than One: Improved Scientific Idea Generation by A {LLM}-Based Multi-Agent System",
    author = "Su, Haoyang  and
      Chen, Renqi  and
      Tang, Shixiang  and
      Yin, Zhenfei  and
      Zheng, Xinzhe  and
      Li, Jinzhe  and
      Qi, Biqing  and
      Wu, Qi  and
      Li, Hui  and
      Ouyang, Wanli  and
      Torr, Philip  and
      Zhou, Bowen  and
      Dong, Nanqing",
    editor = "Che, Wanxiang  and
      Nabende, Joyce  and
      Shutova, Ekaterina  and
      Pilehvar, Mohammad Taher",
    booktitle = "Proceedings of the 63rd Annual Meeting of the Association for Computational Linguistics (Volume 1: Long Papers)",
    month = jul,
    year = "2025",
    address = "Vienna, Austria",
    publisher = "Association for Computational Linguistics",
    url = "https://aclanthology.org/2025.acl-long.1368/",
    doi = "10.18653/v1/2025.acl-long.1368",
    pages = "28201--28240",
    ISBN = "979-8-89176-251-0"
}

@misc{olmo2025olmo3,
      title={Olmo 3}, 
      author={Team Olmo and : and Allyson Ettinger and Amanda Bertsch and Bailey Kuehl and David Graham and David Heineman and Dirk Groeneveld and Faeze Brahman and Finbarr Timbers and Hamish Ivison and Jacob Morrison and Jake Poznanski and Kyle Lo and Luca Soldaini and Matt Jordan and Mayee Chen and Michael Noukhovitch and Nathan Lambert and Pete Walsh and Pradeep Dasigi and Robert Berry and Saumya Malik and Saurabh Shah and Scott Geng and Shane Arora and Shashank Gupta and Taira Anderson and Teng Xiao and Tyler Murray and Tyler Romero and Victoria Graf and Akari Asai and Akshita Bhagia and Alexander Wettig and Alisa Liu and Aman Rangapur and Chloe Anastasiades and Costa Huang and Dustin Schwenk and Harsh Trivedi and Ian Magnusson and Jaron Lochner and Jiacheng Liu and Lester James V. Miranda and Maarten Sap and Malia Morgan and Michael Schmitz and Michal Guerquin and Michael Wilson and Regan Huff and Ronan Le Bras and Rui Xin and Rulin Shao and Sam Skjonsberg and Shannon Zejiang Shen and Shuyue Stella Li and Tucker Wilde and Valentina Pyatkin and Will Merrill and Yapei Chang and Yuling Gu and Zhiyuan Zeng and Ashish Sabharwal and Luke Zettlemoyer and Pang Wei Koh and Ali Farhadi and Noah A. Smith and Hannaneh Hajishirzi},
      year={2025},
      eprint={2512.13961},
      archivePrefix={arXiv},
      primaryClass={cs.CL},
      url={https://arxiv.org/abs/2512.13961}, 
}

@misc{qwen3technicalreport,
      title={Qwen3 Technical Report}, 
      author={Qwen Team},
      year={2025},
      eprint={2505.09388},
      archivePrefix={arXiv},
      primaryClass={cs.CL},
      url={https://arxiv.org/abs/2505.09388}, 
}

@article{Wang2013QuantifyingLS,
  title={Quantifying Long-Term Scientific Impact},
  author={Dashun Wang and Chaoming Song and Albert-Ĺaszl{\'o} Barab{\'a}si},
  journal={Science},
  year={2013},
  volume={342},
  pages={127 - 132},
  url={https://api.semanticscholar.org/CorpusID:260558492}
}

@inproceedings{arnaout2025indepthresearchimpactsummarization,
  title     = {In-depth Research Impact Summarization through Fine-Grained Temporal Citation Analysis},
  author    = {Arnaout, Hiba and Sternlicht, Noy and Hope, Tom and Gurevych, Iryna},
  booktitle = {Proceedings of the 64th Annual Meeting of the Association for Computational Linguistics (Volume 1: Long Papers)},
  year      = {2026},
  publisher = {Association for Computational Linguistics},
  note      = {arXiv:2505.14838},
  url       = {https://arxiv.org/abs/2505.14838},
}

@article{Feng2025ResponseQA,
  title={Response Quality Assessment for Retrieval-Augmented Generation via Conditional Conformal Factuality},
  author={Naihe Feng and Yi Sui and Shiyi Hou and Jesse C. Cresswell and Ga Wu},
  journal={Proceedings of the 48th International ACM SIGIR Conference on Research and Development in Information Retrieval},
  year={2025},
  url={https://api.semanticscholar.org/CorpusID:280011519}
}

@article{Li2025ConformalPF,
  title={Conformal Prediction for Signal Temporal Logic Inference},
  author={Danyang Li and Yixuan Wang and Matthew Cleaveland and Mingyu Cai and Roberto Tron},
  journal={ArXiv},
  year={2025},
  volume={abs/2509.25473},
  url={https://api.semanticscholar.org/CorpusID:281682043}
}

@inproceedings{opris2025runtimeanalysisnsgaiiimanyobjective,
  title     = {A First Runtime Analysis of NSGA-III on a Many-Objective Multimodal Problem: Provable Exponential Speedup via Stochastic Population Update},
  author    = {Opris, Andre},
  booktitle = {Proceedings of the Thirty-Fourth International Joint Conference on Artificial Intelligence, {IJCAI-25}},
  pages     = {8903--8911},
  year      = {2025},
  publisher = {International Joint Conferences on Artificial Intelligence Organization},
  doi       = {10.24963/ijcai.2025/990},
  url       = {https://doi.org/10.24963/ijcai.2025/990},
  note      = {arXiv:2505.01256}
}

@misc{fu2024diffpaddenoisingdiffusionbasedadversarial,
      title={DiffPAD: Denoising Diffusion-based Adversarial Patch Decontamination}, 
      author={Jia Fu and Xiao Zhang and Sepideh Pashami and Fatemeh Rahimian and Anders Holst},
      year={2024},
      eprint={2410.24006},
      archivePrefix={arXiv},
      primaryClass={cs.CV},
      url={https://arxiv.org/abs/2410.24006}, 
}

@inproceedings{shalyt2024unsuperviseddiscoveryformulasmathematical,
  title     = {Unsupervised Discovery of Formulas for Mathematical Constants},
  author    = {Shalyt, Michael and Seligmann, Uri and Beit Halachmi, Itay and David, Ofir and Elimelech, Rotem and Kaminer, Ido},
  booktitle = {Advances in Neural Information Processing Systems},
  volume    = {37},
  year      = {2024},
  publisher = {Curran Associates, Inc.},
  doi       = {10.52202/079017-3596},
  url       = {https://proceedings.neurips.cc/paper_files/paper/2024/hash/cd8b5de90ebfd6df2b703d2346370cba-Abstract-Conference.html},
  note      = {arXiv:2412.16818}
}

@misc{riechers2025neuralnetworksleveragenominally,
      title={Neural networks leverage nominally quantum and post-quantum representations}, 
      author={Paul M. Riechers and Thomas J. Elliott and Adam S. Shai},
      year={2025},
      eprint={2507.07432},
      archivePrefix={arXiv},
      primaryClass={cs.LG},
      url={https://arxiv.org/abs/2507.07432}, 
}

@misc{huang2025posspositionspecialistgenerates,
      title={POSS: Position Specialist Generates Better Draft for Speculative Decoding}, 
      author={Langlin Huang and Chengsong Huang and Jixuan Leng and Di Huang and Jiaxin Huang},
      year={2025},
      eprint={2506.03566},
      archivePrefix={arXiv},
      primaryClass={cs.CL},
      url={https://arxiv.org/abs/2506.03566}, 
}

@inproceedings{Alexander_2025, series={SIGIR ’25},
   title={In a Few Words: Comparing Weak Supervision and LLMs for Short Query Intent Classification},
   url={http://dx.doi.org/10.1145/3726302.3730213},
   DOI={10.1145/3726302.3730213},
   booktitle={Proceedings of the 48th International ACM SIGIR Conference on Research and Development in Information Retrieval},
   publisher={ACM},
   author={Alexander, Daria and de Vries, Arjen P.},
   year={2025},
   month=jul, pages={2977–2981},
   collection={SIGIR ’25} }

@misc{zhao2025causalinterpretabilityadversarialrobustness,
      title={Causal Interpretability for Adversarial Robustness: A Hybrid Generative Classification Approach}, 
      author={Chunheng Zhao and Pierluigi Pisu and Gurcan Comert and Negash Begashaw and Varghese Vaidyan and Nina Christine Hubig},
      year={2025},
      eprint={2412.20025},
      archivePrefix={arXiv},
      primaryClass={cs.CV},
      url={https://arxiv.org/abs/2412.20025}, 
}

@misc{lewandowski2024plasticlearningdeepfourier,
      title={Plastic Learning with Deep Fourier Features}, 
      author={Alex Lewandowski and Dale Schuurmans and Marlos C. Machado},
      year={2024},
      eprint={2410.20634},
      archivePrefix={arXiv},
      primaryClass={cs.LG},
      url={https://arxiv.org/abs/2410.20634}, 
}

@misc{zhang2025policynewtonalgorithmreproducing,
      title={Policy Newton Algorithm in Reproducing Kernel Hilbert Space}, 
      author={Yixian Zhang and Huaze Tang and Chao Wang and Wenbo Ding},
      year={2025},
      eprint={2506.01597},
      archivePrefix={arXiv},
      primaryClass={cs.LG},
      url={https://arxiv.org/abs/2506.01597}, 
}

@inproceedings{ni2024enatrethinkingspatialtemporalinteractions,
  title={{ENAT}: Rethinking Spatial-temporal Interactions in Token-based Image Synthesis},
  author={Ni, Zanlin and Wang, Yulin and Zhou, Renping and Han, Yizeng and Guo, Jiayi and Liu, Zhiyuan and Yao, Yuan and Huang, Gao},
  booktitle={Advances in Neural Information Processing Systems 37 (NeurIPS 2024)},
  year={2024},
  publisher={Curran Associates, Inc.},
  doi={10.52202/079017-2870},
  url={https://proceedings.neurips.cc/paper_files/paper/2024/hash/a488aa1a0c00d76db8a922ef7815a786-Abstract-Conference.html},
  note={arXiv:2411.06959}
}

@misc{yang2024benchmarkingllmguardrailshandling,
      title={Benchmarking LLM Guardrails in Handling Multilingual Toxicity}, 
      author={Yahan Yang and Soham Dan and Dan Roth and Insup Lee},
      year={2024},
      eprint={2410.22153},
      archivePrefix={arXiv},
      primaryClass={cs.CL},
      url={https://arxiv.org/abs/2410.22153}, 
}

@inproceedings{sharifymoghaddam2025rankllmpythonpackagereranking,
  title     = {RankLLM: A Python Package for Reranking with LLMs},
  author    = {Sharifymoghaddam, Sahel and Pradeep, Ronak and Slavescu, Andre and Nguyen, Ryan and Xu, Andrew and Chen, Zijian and Zhang, Yilin and Chen, Yidi and Xian, Jasper and Lin, Jimmy},
  booktitle = {Proceedings of the 48th International ACM SIGIR Conference on Research and Development in Information Retrieval (SIGIR '25)},
  year      = {2025},
  pages     = {3681--3690},
  publisher = {Association for Computing Machinery},
  address   = {Padua, Italy},
  doi       = {10.1145/3726302.3730331},
  url       = {https://doi.org/10.1145/3726302.3730331},
  note      = {arXiv:2505.19284}
}

@misc{hadžić2025distillationtractablemodelvqvae,
      title={Distillation of a tractable model from the VQ-VAE}, 
      author={Armin Hadžić and Milan Papez and Tomáš Pevný},
      year={2025},
      eprint={2509.01400},
      archivePrefix={arXiv},
      primaryClass={cs.LG},
      url={https://arxiv.org/abs/2509.01400}, 
}

@misc{kapoor2025accuracyecol2metricsustainable,
      title={Beyond Accuracy: EcoL2 Metric for Sustainable Neural PDE Solvers}, 
      author={Taniya Kapoor and Abhishek Chandra and Anastasios Stamou and Stephen J Roberts},
      year={2025},
      eprint={2505.12556},
      archivePrefix={arXiv},
      primaryClass={cs.LG},
      url={https://arxiv.org/abs/2505.12556}, 
}

@inproceedings{semnani2025churromakinghistoryreadable,
      title={CHURRO: Making History Readable with an Open-Weight Large Vision-Language Model for High-Accuracy, Low-Cost Historical Text Recognition},
      author={Semnani, Sina and Zhang, Han and He, Xinyan and Tekgurler, Merve and Lam, Monica},
      editor={Christodoulopoulos, Christos and Chakraborty, Tanmoy and Rose, Carolyn and Peng, Violet},
      booktitle={Proceedings of the 2025 Conference on Empirical Methods in Natural Language Processing},
      month={nov},
      year={2025},
      address={Suzhou, China},
      publisher={Association for Computational Linguistics},
      pages={34777--34824},
      url={https://aclanthology.org/2025.emnlp-main.1763/},
      doi={10.18653/v1/2025.emnlp-main.1763},
      isbn={979-8-89176-332-6},
      note={arXiv:2509.19768},
}

@misc{miao2024respdiffendtoendmultiscalernn,
      title={RespDiff: An End-to-End Multi-scale RNN Diffusion Model for Respiratory Waveform Estimation from PPG Signals}, 
      author={Yuyang Miao and Zehua Chen and Chang Li and Danilo Mandic},
      year={2024},
      eprint={2410.04366},
      archivePrefix={arXiv},
      primaryClass={eess.SP},
      url={https://arxiv.org/abs/2410.04366}, 
}

@inproceedings{jin2025veriloccendtoendcrossarchitectureregister,
    title = "{V}eri{L}occ: End-to-End Cross-Architecture Register Allocation via {LLM}",
    author = "Jin, Lesheng and Ruan, Zhenyuan and Mai, Haohui and Shang, Jingbo",
    booktitle = "Proceedings of the 2025 Conference on Empirical Methods in Natural Language Processing",
    month = nov,
    year = "2025",
    address = "Suzhou, China",
    publisher = "Association for Computational Linguistics",
    pages = "30252--30262",
    doi = "10.18653/v1/2025.emnlp-main.1538",
    url = "https://aclanthology.org/2025.emnlp-main.1538/",
    note = "arXiv:2506.17506",
}

@inbook{Lee_2025,
   title={Subcortical Masks Generation in CT Images via Ensemble-Based Cross-Domain Label Transfer},
   ISBN={9783031986949},
   ISSN={1611-3349},
   url={http://dx.doi.org/10.1007/978-3-031-98694-9_12},
   DOI={10.1007/978-3-031-98694-9_12},
   booktitle={Medical Image Understanding and Analysis},
   publisher={Springer Nature Switzerland},
   author={Lee, Augustine X. W. and Yeung, Pak-Hei and Rajapakse, Jagath C.},
   year={2025},
   month=jul, pages={160–174} }

@inproceedings{mysore-etal-2022-multi,title = "Multi-Vector Models with Textual Guidance for Fine-Grained Scientific Document Similarity",author = "Mysore, Sheshera and Cohan, Arman and Hope, Tom",editor = "Carpuat, Marine and de Marneffe, Marie-Catherine and Meza Ruiz, Ivan Vladimir",booktitle = NAACL:2022:main,month = jul,year = "2022",address = "Seattle, United States",publisher = acl,url = anth # {2022.naacl-main.331/},doi = "10.18653/v1/2022.naacl-main.331",pages = "4453--4470"
}

@inproceedings{ni-etal-2022-large,title = "Large Dual Encoders Are Generalizable Retrievers",author = "Ni, Jianmo and Qu, Chen and Lu, Jing and Dai, Zhuyun and Hernandez Abrego, Gustavo and Ma, Ji and Zhao, Vincent and Luan, Yi and Hall, Keith and Chang, Ming-Wei and Yang, Yinfei",editor = "Goldberg, Yoav and Kozareva, Zornitsa and Zhang, Yue",booktitle = EMNLP:2022:main,month = dec,year = "2022",address = "Abu Dhabi, United Arab Emirates",publisher = acl,url = anth # {2022.emnlp-main.669/},doi = "10.18653/v1/2022.emnlp-main.669",pages = "9844--9855"
}

@inproceedings{singh-etal-2023-scirepeval,title = "{S}ci{R}ep{E}val: A Multi-Format Benchmark for Scientific Document Representations",author = "Singh, Amanpreet and D{'}Arcy, Mike and Cohan, Arman and Downey, Doug and Feldman, Sergey",editor = "Bouamor, Houda and Pino, Juan and Bali, Kalika",booktitle = EMNLP:2023:main,month = dec,year = "2023",address = "Singapore",publisher = acl,url = anth # {2023.emnlp-main.338/},doi = "10.18653/v1/2023.emnlp-main.338",pages = "5548--5566"
}

@inproceedings{baek-etal-2025-researchagent,title = "{R}esearch{A}gent: Iterative Research Idea Generation over Scientific Literature with Large Language Models",author = "Baek, Jinheon and Jauhar, Sujay Kumar and Cucerzan, Silviu and Hwang, Sung Ju",editor = "Chiruzzo, Luis and Ritter, Alan and Wang, Lu",booktitle = NAACL:2025:long,month = apr,year = "2025",address = "Albuquerque, New Mexico",publisher = acl,url = anth # {2025.naacl-long.342/},pages = "6709--6738",ISBN = "979-8-89176-189-6"
}

@article{Fortunato2018ScienceOfScience,
  title={Science of Science},
  author={Santo Fortunato and Carl T. Bergstrom and Katy B{\"o}rner and James A. Evans and Dirk Helbing and Sta{\v{s}}a Milojevi{\'c} and Alexander M. Petersen and Filippo Radicchi and Roberta Sinatra and Brian Uzzi and Alessandro Vespignani and Ludo Waltman and Dashun Wang and Albert-L{\'a}szl{\'o} Barab{\'a}si},
  journal={Science},
  year={2018},
  volume={359},
  number={6379},
  pages={eaao0185},
  doi={10.1126/science.aao0185}
}

@article{Hicks2015LeidenManifesto,
  title={Bibliometrics: The {L}eiden Manifesto for research metrics},
  author={Diana Hicks and Paul Wouters and Ludo Waltman and Sarah de Rijcke and Ismael Rafols},
  journal={Nature},
  year={2015},
  volume={520},
  number={7548},
  pages={429--431},
  doi={10.1038/520429a}
}

@article{Balog2012ExpertiseRetrieval,
  title={Expertise Retrieval},
  author={Krisztian Balog and Yi Fang and Maarten de Rijke and Pavel Serdyukov and Luo Si},
  journal={Foundations and Trends in Information Retrieval},
  year={2012},
  volume={6},
  number={2--3},
  pages={127--256}
}

@article{rezaee2025fos,
  title         = {{FOS}: A Large-Scale Temporal Graph Benchmark for Scientific Interdisciplinary Link Prediction},
  author        = {Rezaee, Kiyan and Ziabakhsh, Morteza and Nikfarjam, Niloofar and others},
  journal       = {arXiv preprint arXiv:2511.18631},
  year          = {2025},
  url           = {https://arxiv.org/abs/2511.18631}
}

@article{ye2026proofoftime,
  title         = {Proof of Time: A Benchmark for Evaluating Scientific Idea Judgments},
  author        = {Ye, Bingyang and Chen, Shan and Tu, Jingxuan and others},
  journal       = {arXiv preprint arXiv:2601.07606},
  year          = {2026},
  url           = {https://arxiv.org/abs/2601.07606}
}

@article{wang2025sciencemeter,
  title         = {{ScienceMeter}: Tracking Scientific Knowledge Updates in Language Models},
  author        = {Wang, Yike and Feng, Shangbin and Tsvetkov, Yulia and Hajishirzi, Hannaneh},
  journal       = {arXiv preprint arXiv:2505.24302},
  year          = {2025},
  url           = {https://arxiv.org/abs/2505.24302}
}

@incollection{salatino2021detectionanalysisprediction,
  title         = {Detection, Analysis, and Prediction of Research Topics with Scientific Knowledge Graphs},
  author        = {Salatino, Angelo and Mannocci, Andrea and Osborne, Francesco},
  booktitle     = {Predicting the Dynamics of Research Impact},
  publisher     = {Springer},
  year          = {2021},
  doi           = {10.1007/978-3-030-86668-6_11}
}

@article{ofer2023whatsnext,
  title         = {What's next? Forecasting scientific research trends},
  author        = {Ofer, Dan and Kaufman, Hadasah and Linial, Michal},
  journal       = {Heliyon},
  year          = {2023},
  doi           = {10.1016/j.heliyon.2023.e23781}
}

@article{amancio2015comparingreal,
  title         = {Comparing the topological properties of real and artificially generated scientific manuscripts},
  author        = {Amancio, Diego Raphael},
  journal       = {Scientometrics},
  volume        = {105},
  pages         = {1763--1779},
  year          = {2015},
  doi           = {10.1007/s11192-015-1637-z}
}

@article{hu2024nova,
  title         = {{Nova}: An Iterative Planning and Search Approach to Enhance Novelty and Diversity of {LLM} Generated Ideas},
  author        = {Hu, Xiang and Fu, Hongyu and Wang, Jinge and others},
  journal       = {arXiv preprint arXiv:2410.14255},
  year          = {2024},
  url           = {https://arxiv.org/abs/2410.14255}
}
\bibliographystyle{plain}

\appendix
\newpage

\section{Dataset} \label{app:dataset}

\subsection{Summary Statistics} \label{app:dataset_statistics}

\begin{figure}[!htb]
    \centering
    \begin{subfigure}{.45\textwidth}
    \includegraphics[width=\textwidth,alt=Two highly overlapped histograms with a long right tail]{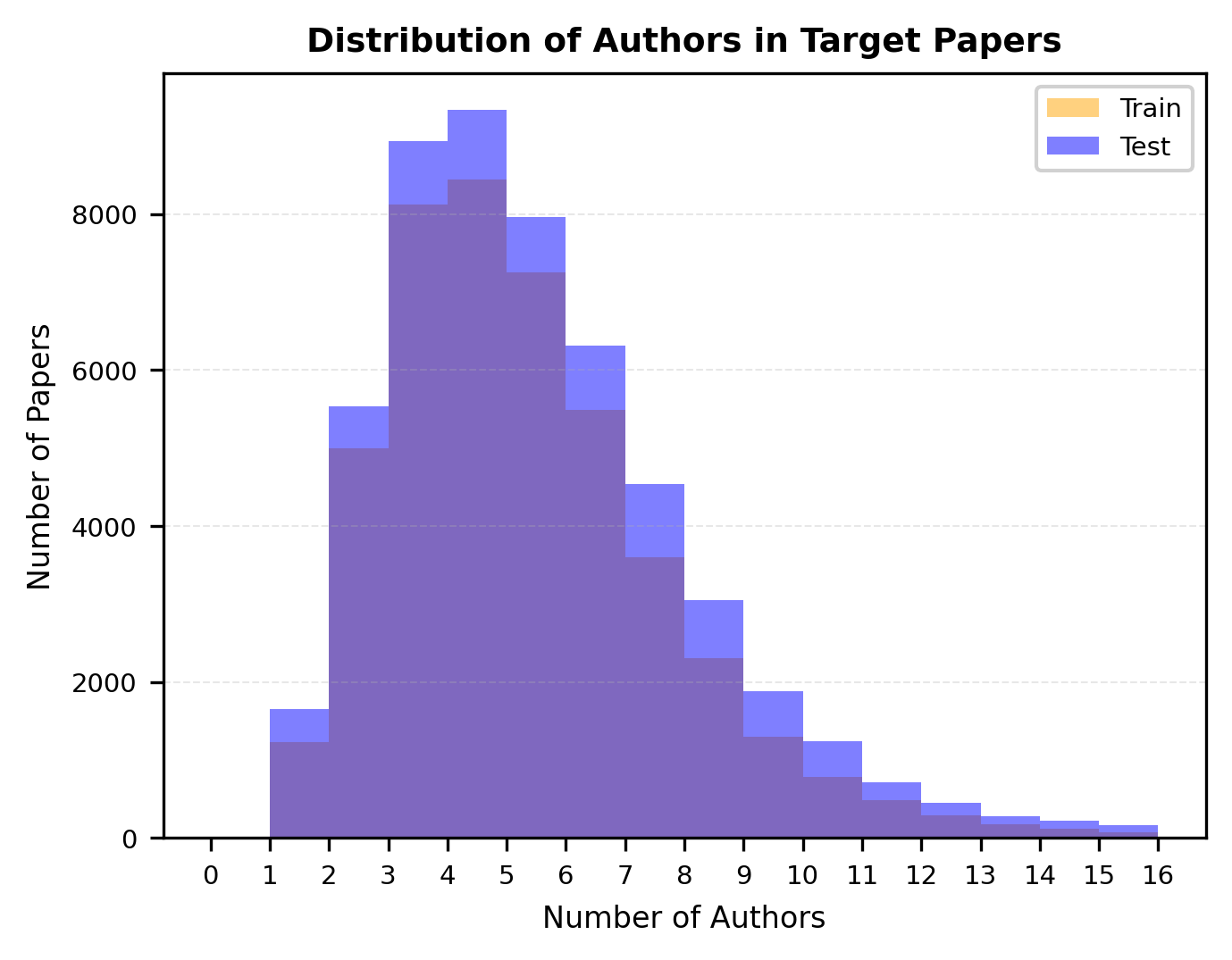}
    \caption{Number of Authors per paper}
    \label{fig:author_counts}
    \end{subfigure}\hfill
    \begin{subfigure}{.45\textwidth}
    \includegraphics[width=\textwidth,alt=Rapidly decreaseing author history lengths with most authors having little or no history]{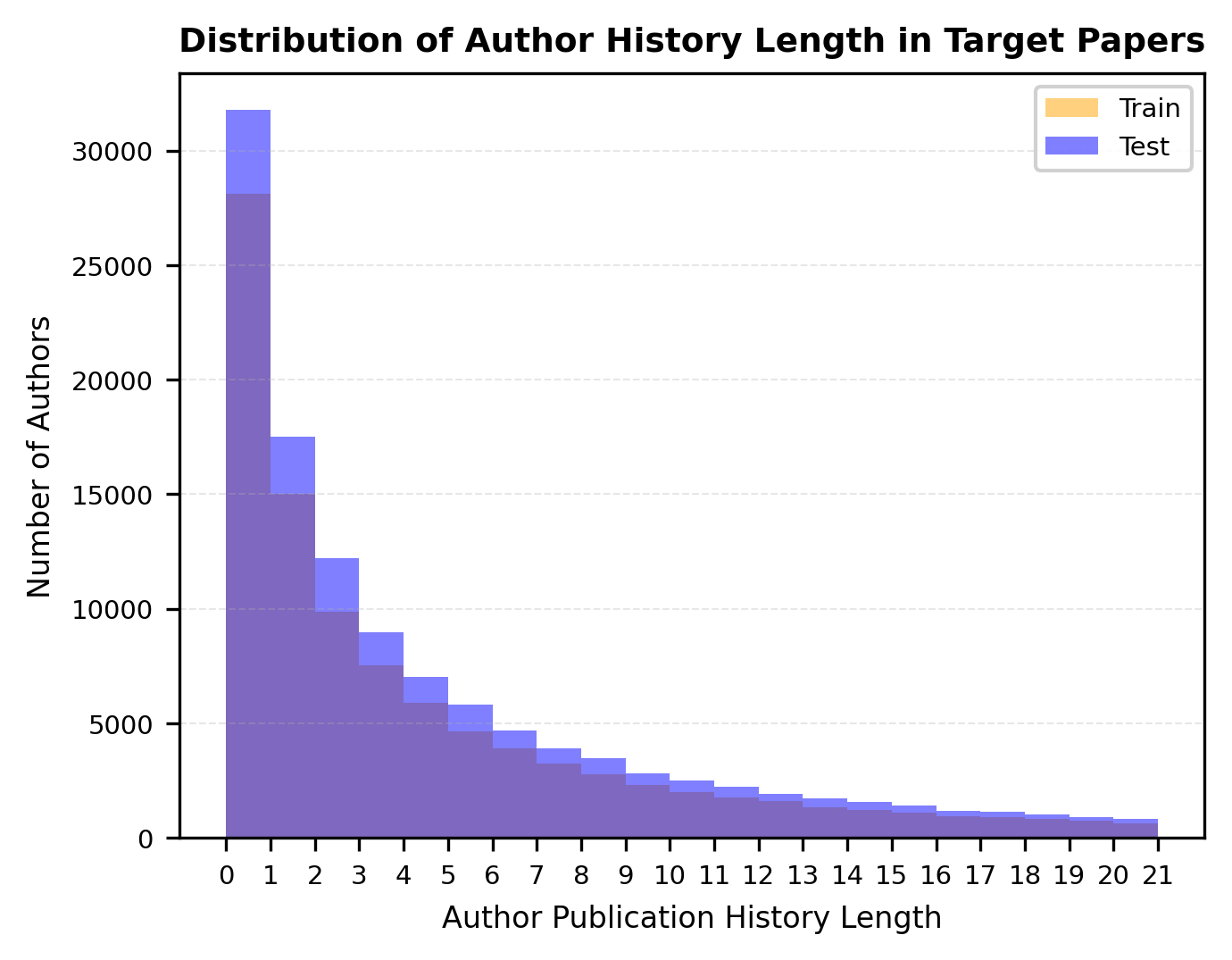}
    \caption{Author Publication History Length}
    \label{fig:author_history}
    \end{subfigure}
    \hfill
    \begin{subfigure}{0.45\textwidth}
        \centering
        \includegraphics[width=\textwidth]{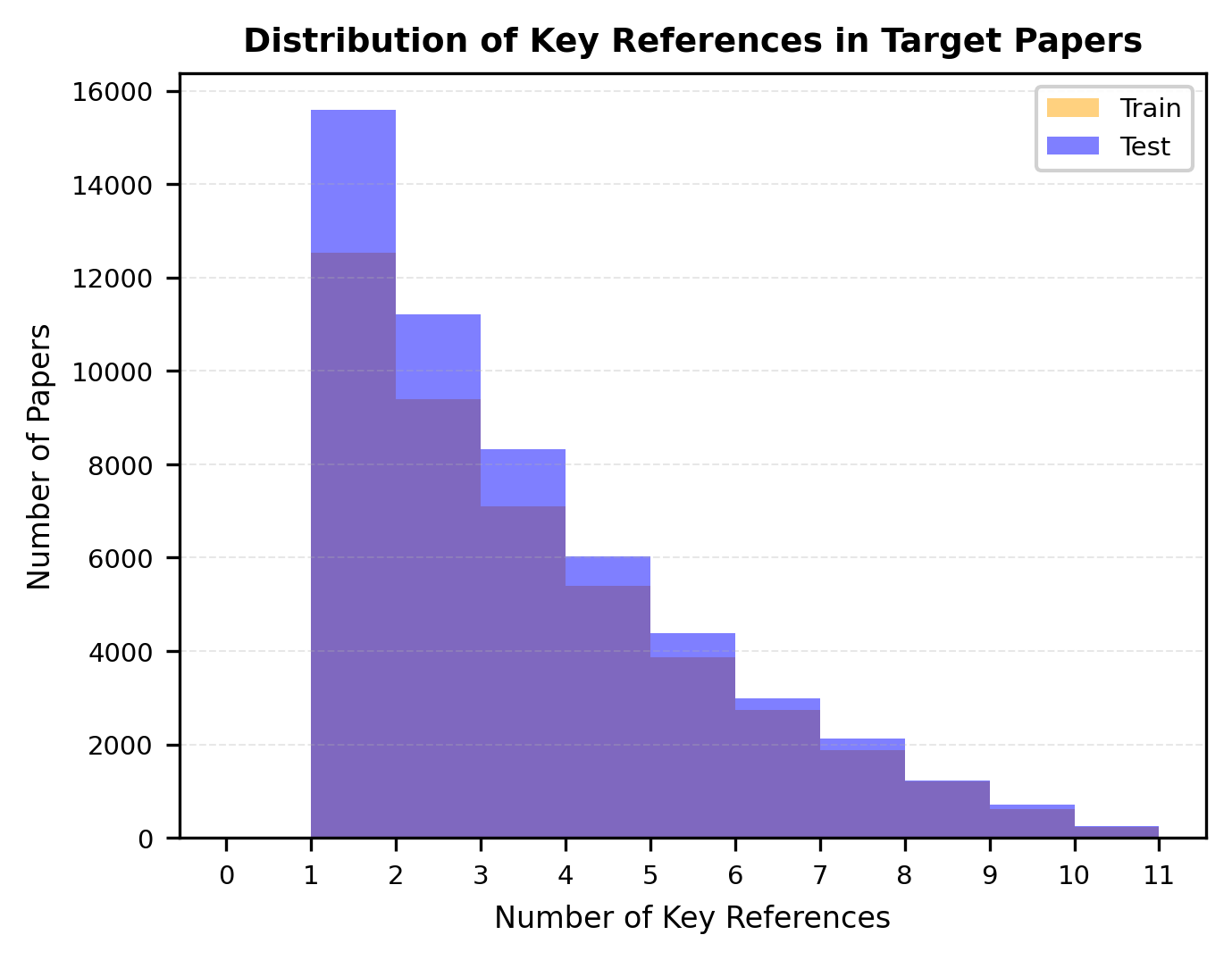}
        \caption{Number of Influential References per paper}
        \label{fig:key_papers}
    \end{subfigure}
    \hfill
    \begin{subfigure}{0.45\textwidth}
        \centering
        \includegraphics[width=\textwidth]{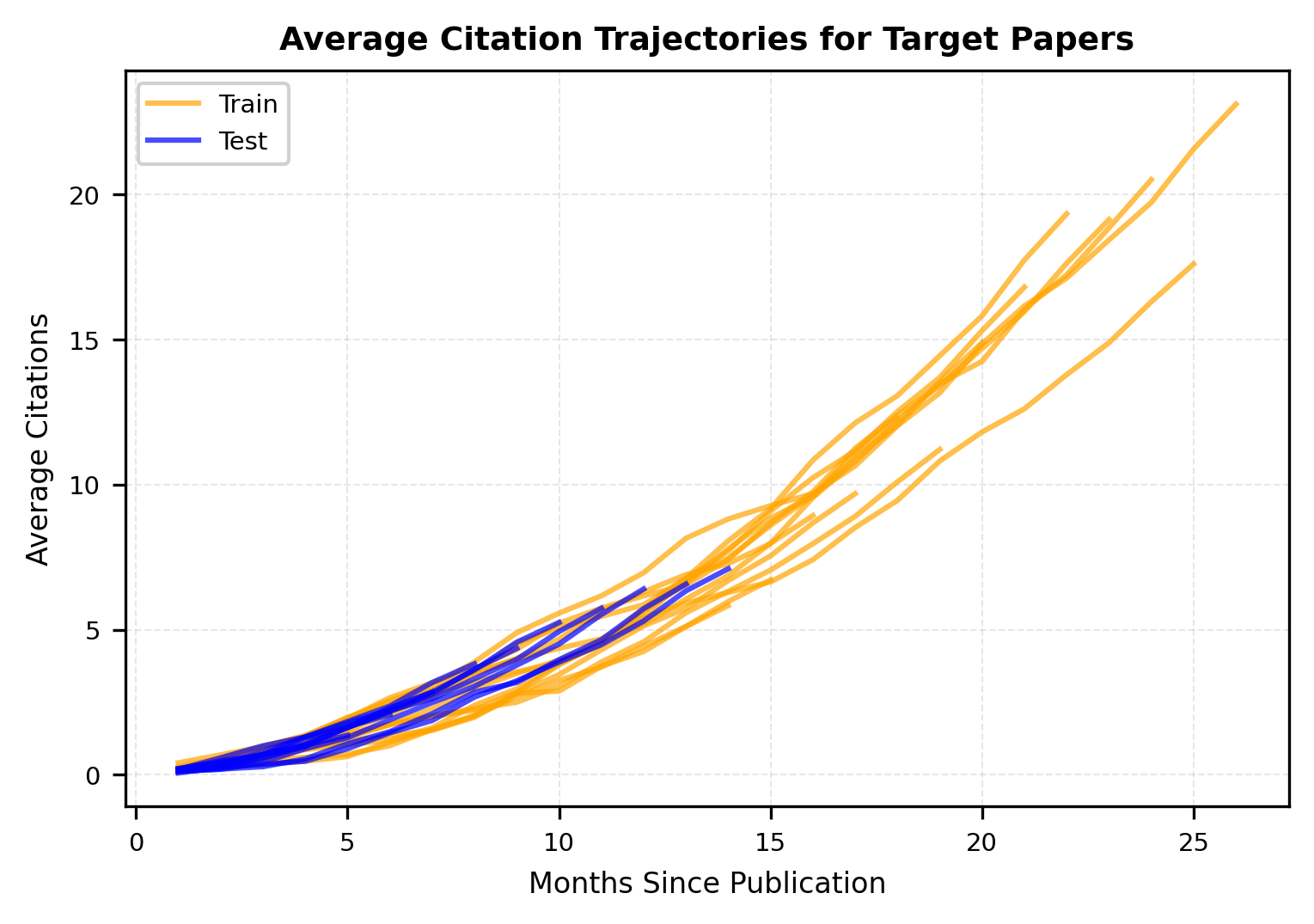}
        \caption{(Averaged) Citation Trajectories of target papers of various ages in the \benchmark~corpus}
        \label{fig:citation_counts}
    \end{subfigure}
        
        \caption{Author, Influential Reference and Citation Trajectory statistics plotted over Target papers.}
        \label{fig:dataset_statistics}
\end{figure}

Figure~\ref{fig:dataset_statistics} visualizes key properties of the \benchmark~dataset over target papers, including distributions of author counts per paper, author publication history lengths, influential reference counts, and citation trajectories. These statistics highlight the heavy-tailed and heterogeneous structure of the benchmark, which underlies the difficulty of forecasting collaboration, literature choice, and downstream impact.

\subsection{Features} \label{app:dataset_features}

We organize papers in \benchmark~into four roles: target papers, influential references of target papers, papers in a target author’s publication history, and influential references of those publication-history papers. All papers share a common set of bibliographic fields (Semantic Scholar corpus ID, arXiv ID, publication date, arXiv categories, title, and abstract).

For target papers, we additionally ensure the availability of complete, temporally aligned citation- and author-level metadata, including influential references, cumulative citation counts at the time of publication, and author statistics (IDs, names, $h$-indices, publication counts, and citation counts), as well as each author’s publication history up to the same publication time.

For certain companion papers, some feature availability is \emph{best-effort}: we include influential references and basic author identity when they can be reliably recovered from the Semantic Scholar Graph and matched to arXiv preprints, but these fields may be empty when a cited work is not indexed by Semantic Scholar or does not have an arXiv version. Table~\ref{tab:dataset_features} summarizes feature availability by role, with checkmarks indicating required fields and parentheses indicating best-effort fields.

\begin{table*}[!htb]
\centering
\small
\caption{Feature availability by paper role in the \benchmark\ dataset. A checkmark (\checkmark) indicates that the field is provided; parentheses indicate best-effort availability. All papers are restricted to arXiv preprints reachable from at least one target paper through the relations described in Section~\ref{sec:dataset}.}
\label{tab:dataset_features}
\begin{tabular}{lcccc}
\toprule
\textbf{Field} 
& \textbf{Target} 
& \textbf{Target}.\textbf{Infl. Ref} 
& \textbf{Author Pub. Hist.} 
& \textbf{Author Pub. Hist. Infl. Ref} \\
\midrule
\multicolumn{5}{l}{\textit{Paper Metadata}} \\
\cmidrule(lr){1-5}
Corpus ID                & \checkmark & \checkmark & \checkmark & \checkmark \\
arXiv ID                & \checkmark & \checkmark & \checkmark & \checkmark \\
Publication Date       & \checkmark & \checkmark & \checkmark & \checkmark \\
arXiv Categories       & \checkmark & \checkmark & \checkmark & \checkmark \\
Topics       & \checkmark & \checkmark & \checkmark & \checkmark \\
Title                  & \checkmark & \checkmark & \checkmark & \checkmark \\
Abstract               & \checkmark & \checkmark & \checkmark & \checkmark \\
\midrule
\multicolumn{5}{l}{\textit{Citation and Reference Data}} \\
\cmidrule(lr){1-5}
Influential References   & \checkmark & -- & (\checkmark) & -- \\
Citation Trajectory & \checkmark & -- & -- & -- \\
\midrule
\multicolumn{5}{l}{\textit{Author Metadata}} \\
\cmidrule(lr){1-5}
Author IDs            & \checkmark & -- & (\checkmark) & -- \\
Author Names         & \checkmark & -- & (\checkmark) & -- \\
Author $h$-index     & \checkmark & -- & -- & -- \\
Author Num. Papers   & \checkmark & -- & -- & -- \\
Author Num. Citations& \checkmark & -- & -- & -- \\
Publication History  & \checkmark & -- & -- & -- \\
\bottomrule
\end{tabular}
\end{table*}

\subsection{Topic Taxonomy and Labeling} \label{app:dataset_topics}

\paragraph{Topic taxonomy.}
We construct a flat taxonomy of 202 AI research topics for aggregate forecasting. 
We obtained an initial topic set by prompting GPT-5.4 with internet access enabled to propose approximately 200 fine-grained AI research topics active in the train-period literature associated with the arXiv categories listed in Section~\ref{sec:dataset}.
We then manually refined the initial list by adding missing topics to improve coverage over the \benchmark~dataset and merging redundant entries, yielding the final 202-topic taxonomy.

We provide the full topic list below: 

\footnotesize
\texttt{["Instruction tuning", "Preference optimization / alignment", "RLHF / RLAIF for post-training", "Prompt engineering / prompt optimization", "Prompt tuning / soft prompting", "Parameter-efficient fine-tuning", "In-context learning", "Long-context modeling", "Personalized language modeling", "Retrieval-augmented generation", "Tool / API-augmented language modeling", "Knowledge editing / model updating", "Model compression / distillation for LMs", "Chain-of-thought prompting", "Tree/search-based reasoning", "Self-critique / self-refinement", "Language-model planning", "Web/navigation agents", "Code-generation agents", "Multi-agent LLM collaboration", "Memory mechanisms for LLM agents", "Multimodal language agents", "Agent evaluation and benchmarks", "Hallucination detection and mitigation", "Calibration / uncertainty estimation for LLMs", "Jailbreak and prompt-injection robustness", "Privacy leakage / machine unlearning for LMs", "Mechanistic interpretability of transformers and LLMs", "Dialogue modeling", "Dialogue state tracking", "Dialogue evaluation", "Multimodal dialogue", "User modeling for conversational AI", "Proactive conversational AI", "Contextual dialogue modeling", "Information extraction", "Relation extraction", "Event extraction", "Question answering", "Summarization", "Long-document understanding", "Argument mining", "Stance detection", "Sentiment analysis", "Syntactic parsing / tagging / chunking", "Morphology and word segmentation", "Sentence-level semantics and textual inference", "Figurative language understanding / generation", "Machine translation", "Low-resource NLP", "Multilingual representation learning", "Cross-cultural NLP", "Multimodality and language grounding", "Knowledge-augmented NLP", "Scientific NLP", "Scholarly document processing", "Citation / evidence extraction", "Climate NLP", "Language-and-molecules modeling", "Dense retrieval", "Neural ranking / learning to rank", "Generative retrieval", "Conversational search", "RAG system design and evaluation", "Knowledge-enhanced retrieval", "Long-context retrieval", "LLM-based IR evaluation", "Collaborative filtering", "Knowledge-based recommendation", "Deep learning for recommender systems", "Natural-language / conversational recommenders", "Fair / privacy-aware recommendation", "3D reconstruction from multi-view and sensors", "3D reconstruction from single images", "Adversarial attack and defense in vision", "Biometrics", "Computational imaging", "Datasets and evaluation for vision", "Efficient and scalable vision models", "Face analysis", "Body / pose / gesture / motion understanding", "Low-level vision", "Document analysis and understanding", "Open-set / open-world recognition", "Out-of-distribution detection in vision", "Fine-grained visual categorization", "Self-supervised visual representation learning", "Semi-supervised visual learning", "Few-shot visual recognition", "Domain adaptation in vision", "Diffusion models for image synthesis", "Autoregressive image generation", "Generative-model inversion", "Image editing with generative models", "Synthetic data for visual recognition", "Analysis-by-synthesis / render-and-compare recognition", "3D foundation models", "3D content creation", "Neural radiance fields", "3D Gaussian splatting", "Implicit neural representations", "Novel-view synthesis", "Long-form video understanding", "Human action understanding / generation", "Large multimodal model evaluation", "Visual-prompt understanding in multimodal models", "Open-vocabulary / open-task vision-language models", "Vision-language reasoning", "Medical vision foundation models", "Medical image / video generation", "Biomedical image parsing", "Pathology image models", "Radiology foundation-model applications", "Medical imaging data curation", "Medical imaging data augmentation", "Fairness and safety in medical imaging AI", "Automatic speech recognition", "ASR error correction / rescoring", "Multilingual / multi-accent ASR", "Spoken language translation", "Spoken language understanding", "Text-to-speech", "Voice conversion", "Speaker verification", "Speaker diarization", "Speech processing with discrete units", "Speech emotion recognition / paralinguistics", "Pathological / health-related speech analysis", "Multi-channel speech enhancement", "Human-machine spoken interaction", "Responsible speech foundation models", "Speech deepfake / spoofing detection", "Streaming ASR", "Self-supervised learning for ASR", "Target-speaker extraction", "Deep reinforcement learning", "Hierarchical reinforcement learning", "Model-based reinforcement learning", "Policy optimization", "Bandits", "Sequential decision-making under uncertainty", "Active learning", "Adversarial learning", "Causal inference", "Causal discovery", "Causal representation learning", "Uncertainty quantification", "Probabilistic programming", "Approximate / variational inference", "Performative prediction", "Bayesian methods", "Belief propagation", "Robot foundation models", "Representation learning for robotic perception and control", "Imitation learning for robotics", "Reinforcement learning for physical robots", "Learning-and-planning hybrids in robotics", "Uncertainty-aware robotics", "Automatic robotic data generation", "Multimodal robot perception and sensor fusion", "Human-robot interaction with language / gestures", "Learning for robot task and motion planning", "Learning for hardware design and optimization", "Robot safety for learning-based systems", "Robot manipulation", "Visual navigation", "Locomotion learning", "Autonomous driving perception / prediction / planning", "Embodied question answering / embodied vision-language", "Open-world embodied AI", "Generative AI for embodied AI", "Dataset composition and curation for foundation models", "Data filtering / relabeling / augmentation", "Dataset quality / diversity / provenance analysis", "Data debugging / influence analysis", "Dataset distillation", "Synthetic-data effects / model collapse", "Weak-to-strong generalization", "Graph neural networks", "Relational / structured learning", "Spatio-temporal learning", "Time-series modeling", "Time-series foundation models", "Federated learning", "Privacy and security in data-centric ML", "Deep learning theory (training dynamics, generalization, optimization convergence)", "Scientific machine learning (PDE solvers, neural operators, PINNs)", "Continual learning and catastrophic forgetting", "Object detection, segmentation, and tracking in vision", "Video segmentation, tracking, and generation", "Remote sensing and geospatial vision", "Medical image segmentation and reconstruction (beyond foundation models)", "Biomedical signal analysis (EEG, ECG, physiological)", "Quantum machine learning", "Spiking and neuromorphic computing", "Explainable AI (non-mechanistic / non-LLM)", "AI governance, policy, and societal impact", "Systems and serving infrastructure for large models", "Audio and music modeling (non-speech)", "Tabular machine learning and AutoML", "Point cloud and 3D geometric learning", "Wireless communications and signal processing"]}

\normalsize

\paragraph{Per-paper labeling.}
We label each of the target papers in the train and test periods using  \texttt{gpt-5.4-2026-03-05} using the prompt provided in Appendix~\ref{app:prompt_topic_assignment}.
The prompt provides the paper's title and abstract along with the full taxonomy, and asks the model to return all applicable topics in JSON format, ordered by relevance.
We instruct the model to exactly match topics from the taxonomy, include every topic directly addressed by the paper rather than only its primary focus, and exclude topics mentioned only in passing.
Papers for which the model returns no applicable topic are assigned to a residual \textit{Other} category; these account for approximately 6.3\% of target papers and are excluded from topic trend forecasting evaluation. We provide associated statistics in Tables~\ref{tab:topic_labels_per_paper} and~\ref{tab:papers_per_topic_label}:

\begin{table}[t]
\centering
\caption{Summary statistics for topic labels per target paper. Papers assigned only to the residual \textit{Other} category are counted as having zero taxonomy labels.}
\label{tab:topic_labels_per_paper}
\small
\setlength{\tabcolsep}{4pt}
\begin{tabular}{lrrrrrrr}
\toprule
\textbf{Subset} & \textbf{$n$} & \textbf{Mean} & \textbf{Stdev} & \textbf{Min} & \textbf{p50} & \textbf{p95} & \textbf{Max} \\
\midrule
All papers, incl. \textit{Other} & $97{,}820$ & $3.15$ & $1.83$ & $0$ & $3$ & $6$ & $16$ \\
Papers with $\geq 1$ label & $91{,}644$ & $3.37$ & $1.69$ & $1$ & $3$ & $6$ & $16$ \\
\bottomrule
\end{tabular}
\end{table}

\begin{table}[t]
\centering
\caption{Summary statistics for the number of papers assigned to each topic label across time slices.}
\label{tab:papers_per_topic_label}
\small
\setlength{\tabcolsep}{4pt}
\begin{tabular}{lrrrrrrr}
\toprule
\textbf{Slice} & \textbf{$n$} & \textbf{Mean} & \textbf{Stdev} & \textbf{Min} & \textbf{p50} & \textbf{p95} & \textbf{Max} \\
\midrule
Train & $202$ & $665$ & $753$ & $12$ & $410$ & $2{,}046$ & $5{,}546$ \\
Test & $202$ & $862$ & $977$ & $15$ & $521$ & $2{,}758$ & $6{,}982$ \\
All-time & $202$ & $1{,}527$ & $1{,}713$ & $29$ & $973$ & $4{,}256$ & $12{,}528$ \\
\bottomrule
\end{tabular}
\end{table}

\paragraph{Cross-labeler validation.}
To assess labeling reliability, we relabel a random sample of 530 papers with Claude Opus 4.7 using the same prompt.
Agreement in assigned labels is substantial: per-paper label F1 is 0.67 over all 530 papers, and 0.71 over the 474 papers for which both labelers assign at least one topic.

\paragraph{External validation of topic labels.}
To check that the GPT-5.4 topic labels correspond to semantically meaningful paper groupings, we compare them against GRIT embeddings, an independent representation computed without access to our taxonomy.
For each of the 202 topics, we sample 2,000 pairs of papers assigned to that topic and compute their mean cosine similarity.
We compare this to a null distribution estimated from 50,000 random paper pairs.
Intra-topic similarity exceeds the null for every topic, with a median $z$-score of $1.33$.
Since GRIT is trained independently of our topic taxonomy, this consistent positive gap suggests that the labels recover semantically coherent groups of papers rather than arbitrary partitions.

Together, the cross-labeler agreement and embedding-based validation suggest that the GPT-5.4 labels provide a reasonable topic assignment for our topic trend forecasting task; however, \benchmark~also provides various kinds of paper metadata (Appendix~\ref{app:dataset_features}), allowing users to construct alternative topic taxonomies or labels for their own analyses, including with more expensive or specialized methods.

\newpage
\section{A Metric for Contribution Generation}
\label{app:followup_work_metric}

\subsection{FacetScore}
\label{app:followup_work_metric_facetscore}
Unsatisfied with existing measures of textual similarity (ROUGE-L, BERTScore~\citep{Zhang2019BERTScoreET}, ASPIRE-OT~\citep{mysore-etal-2022-multi}), we developed a similarity metric called FacetScore based on the Scideator project \citep{Radensky2024ScideatorHS}. Intended to assist in scientific ideation, Scideator \citep{Radensky2024ScideatorHS} introduced the representation of a scientific advance as a combination of several \textit{facets}: purpose, mechanism, and evaluation. We further added a notion of the scientific contribution type \citep{Pramanick2024TheNO} (an artifact, knowledge, or better understanding) into this collection of facets. Once FacetScore extracts each of these fields from the provided pair of title-abstracts, it prompts an LLM to score the similarity between corresponding pairs of facets on a five-point scale. Finally, FacetScore returns the average of these facet-level similarity scores as the overall similarity score between the two provided papers. 

However, we opted to omit FacetScore computations from out later experiments since we found that \followupmetric~judgements correlate significantly better with human judgements (Figure~\ref{fig:lacerscore_matrix}).

\subsection{\followupmetric}
\label{app:followup_work_metric_lacerscore}

In our experiments, existing automatic textual similarity metrics (ROUGE-L, BERTScore~\citep{Zhang2019BERTScoreET}, ASPIRE-OT~\citep{mysore-etal-2022-multi}, retrieval-based mean reciprocal rank (MRR), etc.) exhibited limited dynamic range: substantially different generated title-abstract pairs received similar scores, and it was often unclear or arbitrary, the degree of dissimilarity that yields a score at the lower extreme of the scale. To address this, we define \textit{\followupmetric}, an LLM-as-judge metric explicitly calibrated to a 1-10 semantic alignment scale using automatically constructed demonstrations.

Instead of relying on human-annotated similarity judgments, we generate calibration examples by interpolating between two intuitive endpoints. For each demonstration sequence, an influential reference whose abstract lies at the 50th percentile\footnote{We also experimented with using 25th percentile, 75th percentile, and randomly chosen influential references to define the minimum \followupmetric. We found that this choice does not meaningfully change its judgements.} of $n$-gram overlap with the target defines score 1, representing related but clearly distinct prior work. A paraphrase of the target abstract defines score 10, representing near-semantic equivalence.

We then prompt (Appendix~\ref{app:prompt_lacerscore_interpolations}) GPT-5 to generate intermediate title–abstract pairs for scores 2 through 9 by gradually modifying one semantic aspect at a time (e.g., contribution type, model architecture, or task domain), forming a smooth transition between these endpoints. Five such interpolation sequences are included as few-shot examples in the scoring prompt (Appendix~\ref{app:prompt_lacerscore_judge}), which the judge model uses to assign calibrated 1–10 scores to new generation–reference pairs.

This approach anchors the \followupmetric~scale in concrete, task-specific contrasts while avoiding the cost of manual annotation, while also providing a well-defined and sensitive measure of conceptual alignment in scientific contributions.

\subsubsection{Effect of LLM Judge Choice on \followupmetric}
\label{app:followup_work_metric_lacerscore_judges}
Table~\ref{tab:lacer_judges} compares LACER scores for Contribution Generation under two independent LLM-based judges (GPT-5 and Claude Opus 4.5). We observe shifts in absolute values but strong agreement in relative ordering (Pearson 0.97, Spearman 0.81), suggesting that our main comparisons are not sensitive to the specific choice of judge.

\begin{table*}[!htb]
\centering
\caption{LACER scores (n=1000) for Contribution Generation as judged by GPT-5 and Claude Opus 4.5. The scores have a Pearson correlation of 0.97 and a Spearman correlation of 0.81.}
\label{tab:lacer_judges}
\begin{tabular}{lccc}
\toprule
Model & LACER (GPT-5) & LACER (Opus) & LACER (Avg) \\
\midrule
Influential Reference & 4.30 & 4.05 & 4.18 \\
GPT-4o & 4.74 & 4.53 & 4.63 \\
GPT-4.1 & 5.11 & 4.60 & 4.86 \\
o3 & 5.50 & 4.67 & 5.08 \\
GPT-5 & 5.66 & 4.65 & 5.16 \\
GPT-5.1 & 5.39 & 4.74 & 5.07 \\
GPT-5.2 & 5.63 & 4.70 & 5.16 \\
Claude Sonnet 4.5 & 5.04 & 4.44 & 4.74 \\
Claude Opus 4.5 & 5.04 & 4.55 & 4.79 \\
Gold Paraphrase & 10.00 & 9.79 & 9.89 \\
\bottomrule
\end{tabular}
\end{table*}

\subsection{Validating Contribution Generation Metrics}
\label{app:followup_work_metric_validation}

To develop and select among candidate metrics for the Contribution Generation task, we constructed both a validation set and a test set of human preference judgments.

For the test set, we sampled ten (influential reference, target) pairs from \benchmark’s training set. For each pair, we generated ten possible candidate scientific advances, effectively attempting to reproduce each target ten times from its influential references: Concretely, we prompted each of \texttt{claude-3-sonnet-20240229}, \texttt{gpt-4o-2024-11-20}, \texttt{meta-llama/Meta-Llama-3.1-8B-Instruct}, and \texttt{o3\_mini} to produce three follow-up contributions, yielding 30 candidates per target. From the union of these 30 generations and the target paper’s ground-truth influential references, we randomly sampled ten title–abstract pairs to use for annotation and evaluation.

Human annotators (drawn from \benchmark’s authors) ranked the sampled generated contributions by conceptual similarity to the target paper’s ground-truth abstract, allowing ties. In the test set, each annotator ranked all candidates. For the validation set, we generated ten additional targets with corresponding candidate generations; these were singly annotated. We used the validation set internally to experiment with variations of FacetScore and LACER, and report results in this paper computed over the test set and the final versions of these metrics.

\begin{figure}[!htb]
    \centering
    \includegraphics[width=0.95\textwidth]{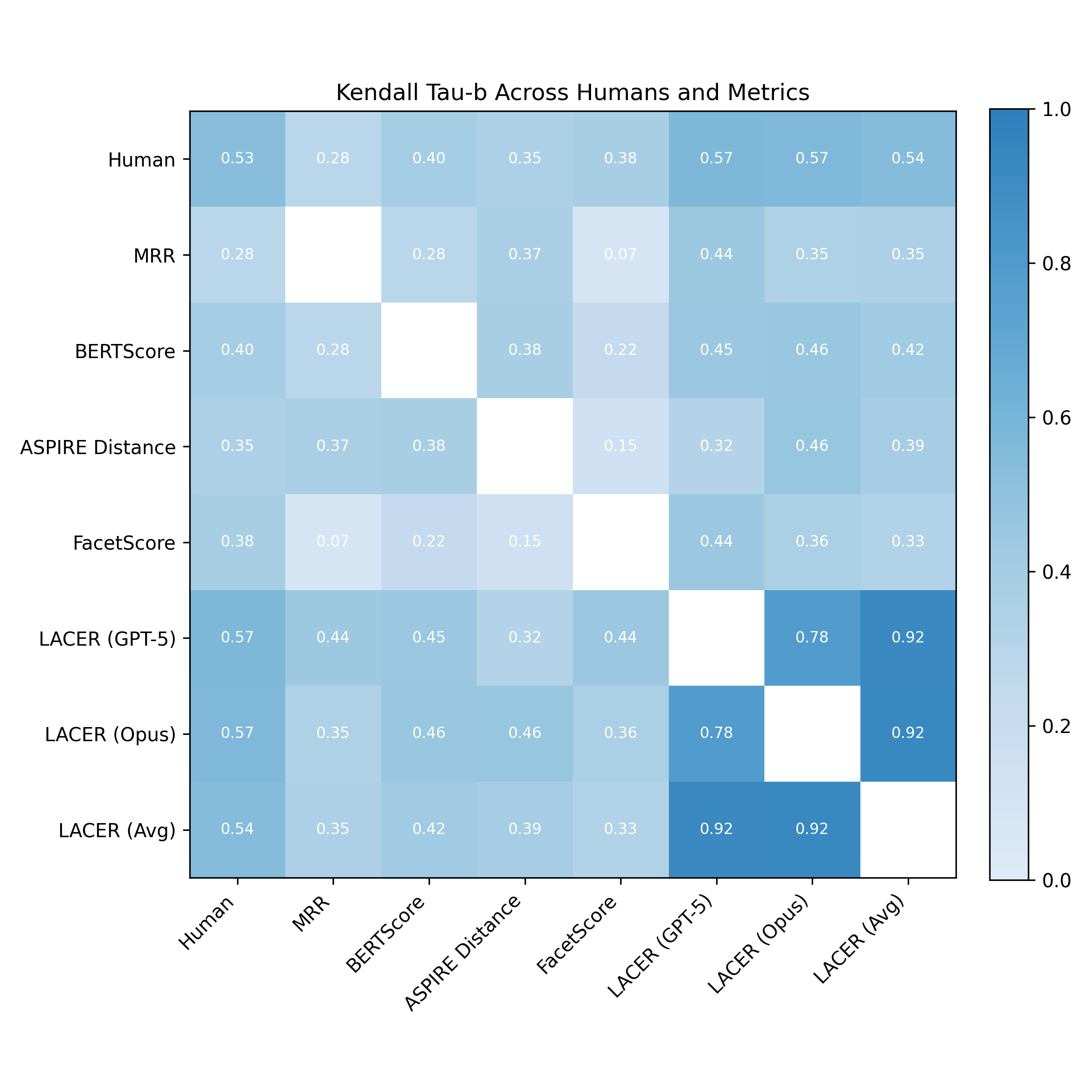}
    \caption{Agreement between humans, models, and aggregates. `Human' refers to an \textit{average} agreement among all five annotators -- agreement computed against all the human annotators (excluding self-agreement), and then averaged. LACER is the only automated metric that reaches agreement with humans that is comparable to their IAA.}
    \label{fig:lacerscore_matrix}
\end{figure}

We measured inter-annotator agreement (IAA) and automated metrics' agreement with human judgments using Kendall’s $\tau_b$ \cite{Kendall1938ANM} evaluated over the above test set. Annotators exhibit substantial agreement on this task, though some variability remains. We find that \followupmetric~is the only metric that approaches agreement comparable with human IAA. Further, \followupmetric~judgments appear robust to the choice of underlying LLM judge (Figure~\ref{fig:lacerscore_matrix}).

\newpage
\section{Experiment Implementation Details} \label{app:exp_details}

\subsection{Contribution Generation}
\label{app:exp_details_contribution_generation}

\paragraph{Finetuned open models.}
We fully finetune OLMo~3~7B and Qwen3~8B on the train split using the same input-output format as the prompted models.
The input is the serialized influential-reference context and the target output is the gold title and abstract. We finetune each model for one epoch using the target papers from the train period. We find that training for more than one epoch leads to overfitting.

\paragraph{Reference baselines.}
The \textit{random influential reference} baseline returns the title and abstract of a randomly selected influential reference of $p$. 
The \textit{primary category} baseline returns the title and abstract of a randomly selected paper from the same primary arXiv category as $p$. 
The \textit{gold paraphrase} baseline uses a paraphrase of the target title and abstract.

\paragraph{Context ablations.}
All GPT-5 ablations use the same $n=500$ target-paper subset and include the influential references $R_p$.
For the related-paper settings, we retrieve up to 10 nearest-neighbor prior papers from $\mathbf{H}^{<t_p}$ under GRIT cosine similarity, querying with the target's title-abstract embedding.
For the author-history settings, we walk the target's authors in listed order and include up to 3 most-recent prior papers per author, stopping once 15 papers have been collected.
For the oracle-citation-count settings, we provide the eventual 12-month citation count as an additional scalar input.

\paragraph{Tool-using GPT-5 agent.}
This baseline replaces the static prompted setup with a \texttt{gpt-5-2025-08-07} agent running an OpenAI tool-calling loop for at most 10 rounds per target. Given the target's authors and influential references, the agent may invoke four corpus tools --- \texttt{search\_papers} (GRIT semantic search over $\mathbf{H}^{<t_p}$), \texttt{get\_paper}, \texttt{get\_author\_recent\_papers}, and \texttt{get\_references\_of} --- and is required to emit \texttt{Reasoning:}, \texttt{Title:}, and \texttt{Abstract:} lines, which are scored as in the prompted baselines. Every tool result passes through a per-target leakage filter that, given the target's publication date $t_p$, drops the target paper itself, any paper with publication date $\geq t_p$, any paper sharing the target's title, and any paper whose abstract overlaps the target's by 50 or more contiguous characters. The system prompt is provided in Appendix~\ref{app:prompt_agent_contribution_generation}.

\subsection{Collaborator Prediction} \label{app:exp_details_collaborator_prediction}

\paragraph{Evaluation.}
We retain target papers with at least two authors whose publication histories are non-empty as of $t_p$, since baselines that rank a candidate set rather than generate it require some signal to anchor on. Candidate authors are similarly restricted to those with at least one prior publication. Unless otherwise noted, we use the first author as the seed; Appendix~\ref{app:sub_seed_author} shows that this choice does not affect baseline ordering.

\paragraph{Baselines.}
Embedding-based methods represent each author by the mean of their $10$ most recent prior papers, and use cosine similarity for GTR and GRIT and Euclidean distance for Specter2. \textit{Hierarchical Clustering} represents each author with $\lceil\sqrt{n}\rceil$ centroids (where $n$ is the author's number of prior papers) over PCA-reduced paper embeddings, scoring candidates by the maximum similarity over (query centroid, candidate centroid) pairs. \textit{Embedding Fusion Proj.} fits a 2-layer MLP on top of frozen GRIT embeddings using a multi-positive InfoNCE objective: positives are the target paper's coauthors; the $5$ hard negatives are sampled from authors who have coauthored with the positives' prior coauthors but not directly with any positive; the $10$ easy negatives are random pre-$t_p$ authors outside both neighborhoods.

\paragraph{Autoresearch.}
We adapt Karpathy's autoresearch framework~\cite{autoresearch}: a Claude Opus~4.7-driven outer loop iteratively designs, implements, and benchmarks candidate ranking pipelines on a 1000-instance train-period dev set under a 12-hour wall-clock budget, returning the highest-scoring configuration. Within each evaluation instance, a per-target temporal filter is applied: the target paper is hidden and all tool calls return $\varnothing$ for any paper, author publication, or citation with publication date $\geq t_p$; candidate sets are restricted to pre-$t_p$ IDs. The same per-target filter applies when the frozen pipeline is evaluated on the test set. The best-found pipeline is a frequency prior over the seed's $45$ most-recent papers, expanded via 2-hop and 3-hop coauthor traversal (top-$150$ and top-$100$ contributors per hop, with multiplicative damping factors of $0.6$ and $0.3$), boosted by a citation-history pull that scores the authors of the seed's $10$ most-recently cited references, and rescaled by an exponential activity decay with a $2.5$-year half-life on each candidate's most-recent publication date.

\paragraph{GPT-5 Agent.}
Uses the same agent loop as in Appendix~\ref{app:exp_details_contribution_generation}, with a fifth tool \texttt{get\_papers\_citing} exposed in addition to the four above and a system prompt instructing the agent to predict the ranked \texttt{author\_id}s of likely future coauthors of the seed (full prompt in Appendix~\ref{app:prompt_agent_collaborator}). The user message provides only the seed \texttt{author\_id} and target date --- no target paper text. All tool results are filtered to papers with publication date strictly before the target's $t_p$. The agent emits \texttt{Reasoning:} and \texttt{Predictions: [author\_id, ...]}; the harness pads short lists to $k=1000$ with random pre-cutoff candidate authors. Evaluated on a random $n=500$ subset.

\subsection{Prior Work Selection} \label{app:exp_details_prior_work_selection}

\paragraph{Evaluation.}
We restrict to target papers with at least one author whose prior publications include a paper with a non-empty influential-reference list, so that frequency- and rank-fusion baselines have a meaningful citation history to draw from. The candidate pool is the full $\mathbf{H}^{<t_p}$ with no topic or neighborhood pre-filter, which makes the task substantially harder than open-domain retrieval --- the model must distinguish a small (1--10) ground-truth set from hundreds of thousands of plausibly relevant papers.

\paragraph{Baselines.}
All retrieval baselines use up to the $10$ most recent prior papers per author. \textit{Embedding Fusion} mean-pools the embeddings of those papers per author, averages the per-author embeddings into a single team query, and ranks candidates by similarity to that one query. \textit{Embedding Fusion Refs} differs in two ways: per author it instead mean-pools the embeddings of the influential references that author has cited in their recent work, and at the team level it queries the index once per author and combines results by summed rank rather than averaging into a single query. The former thus treats an author's own writing as the query, while the latter treats their citation history as the query. \textit{Rank Fusion} queries the index once per cited reference and aggregates rankings weighted by citation multiplicity. \textit{Embedding Fusion Proj.} reuses the projection architecture and training recipe from Appendix~\ref{app:exp_details_collaborator_prediction}, supervised so that mean-pooled author queries are pulled towards the embeddings of their target papers' influential references.

\paragraph{Autoresearch.}
Best-found pipeline is a three-tier ranker that fuses (i) cited-reference frequency across the team's recent papers, (ii) 2-hop co-cited references, and (iii) embedding similarity reranked with team-centroid, per-author-centroid, and cited-reference-centroid boosts. Outer loop, 12-hour budget, and per-target temporal filter as in Appendix~\ref{app:exp_details_collaborator_prediction}.

\paragraph{GPT-5 Agent.}
Five-tool agent loop as in Appendix~\ref{app:exp_details_collaborator_prediction}, with a system prompt instructing the agent to predict the ranked \texttt{corpus\_id}s of the target paper's likely influential references (full prompt in Appendix~\ref{app:prompt_agent_priorwork}). The user message provides the target authors and date but not the target's title or abstract. All tool results are filtered to papers with publication date strictly before the target's $t_p$. Evaluated over a random $n=500$ subset.

\subsection{Future Combination Prediction} \label{app:exp_details_future_influential_co-citation}

\paragraph{Evaluation.}
Targets are drawn from the first four months of the test period, and ground-truth co-citation counts are tallied over the eight months that follow each target's publication date. Targets with no observed co-citations in this window are dropped. The candidate pool is the full $\mathbf{H}^{<t_p}$, with the paper index rolled forward chronologically so that, for each target, only papers published before $t_p$ are visible.

\paragraph{Baselines.}
\textit{Target Similarity} and \textit{Reference Similarity} both retrieve from this chronologically updated index, differing only in the query: the target's title-abstract embedding for the former, and the mean embedding of the target's influential references for the latter. \textit{Citation Frequency} ranks each prior paper $q$ by the number of pre-$t_p$ papers that cite $q$ as an influential reference, and \textit{Co-citation Frequency} ranks $q$ by how often it co-occurs in pre-$t_p$ influential-reference lists with any of $p$'s influential references. We sweep all three embedding types in Appendix~\ref{app:cocitation_embedding_choice}.

\paragraph{Autoresearch.}
Best-found pipeline is reciprocal-rank fusion ($k{=}6$) over three rankings: a reference-centroid query in which each of the target's influential references contributes to the centroid weighted by an IDF computed over pre-$t_p$ papers (so rarely-cited references count more), a target title--abstract query, and a co-citation-graph score that weights citers by similarity to the target with a 365-day recency decay and a topic/category-overlap boost. Outer loop, 12-hour budget, and per-target temporal filter as in Appendix~\ref{app:exp_details_collaborator_prediction}.

\paragraph{GPT-5 Agent.}
Five-tool agent loop as in Appendix~\ref{app:exp_details_collaborator_prediction} (full prompt in Appendix~\ref{app:prompt_agent_cocitation}). Unlike the other ranking tasks, the target paper's title, abstract, topic labels, and influential references \emph{are} exposed to the agent in the user message --- these are causally available, since co-citation labels are derived from papers published after $t_p$, not from the target itself. Other tool results remain filtered to papers with publication date strictly before $t_p$. The system prompt instructs the agent to traverse \texttt{get\_papers\_citing} from each of the target's references and aggregate the other references those citers share. Evaluated over a random $n=500$ subset.

\subsection{Citation Count Prediction} \label{app:exp_details_citation_count}

\paragraph{Evaluation.}
We restrict targets to papers with at least $12$ months of observed citations and have models regress against the log-transformed 12-month count, inverting back to raw counts and clipping at zero before computing raw-space metrics.

\paragraph{Baselines.}
\textit{Bibliometrics} combine per-author $h$-indices, publication and citation counts, team size, and the citation counts of influential references --- all measured at $t_p$. \textit{Target Text} uses the target paper's title-abstract embedding. \textit{Context Text} concatenates two embedding components: the mean over the target's influential-reference embeddings and a per-author embedding equal to the mean over each author's prior publications. Per-paper feature vectors --- whether bibliometric or embedding-based --- expose up to four author slots (positions $1$, $2$, $-2$, and $-1$ in author order), giving the regressor a fixed-width view of lead and anchor authors regardless of team size. XGBoost hyperparameters are tuned via Bayesian optimization (Hyperopt, $100$ evaluations) on a $30\%$ validation slice and the final model is refit on the full training set with the best configuration.

\paragraph{Autoresearch.}
Best-found pipeline is an XGBoost regressor over $\log(1+c)$ trained on train-period papers disjoint from the 1000-instance dev set; the autoresearch outer loop selects the configuration that minimizes dev-set loss, and the frozen pipeline is then evaluated on the test set as in the other tasks. Per-paper features combine the target's GRIT embedding, the mean GRIT embedding of its influential references, and 28 bibliometric scalars (author $h$-index, paper, and citation statistics; reference counts; title/abstract length; topic and category counts; target--reference cosine similarity; publication year and month). Outer loop, 12-hour budget, and per-target temporal filter as in Appendix~\ref{app:exp_details_collaborator_prediction}.

\subsection{Topic Trend Forecasting} \label{app:exp_details_topic_trends}

\paragraph{Baselines.}
All three baselines calibrate against an inner split of the train period: its first six months serve as an inner train period and the remaining six as an inner test period. \textit{Mean} averages the per-topic inner-test paper count across topics and rescales it by the ratio of forecast length to inner-test length, predicting that single constant for every topic. \textit{Citation Momentum} and \textit{Author Momentum} instead fit a topic-agnostic linear scale that converts a per-topic score into a paper-count forecast on the inner split, then apply that scale to scores computed over the full train period. \textit{Citation Momentum} scores each topic by the cumulative observable citations of its papers as of the relevant cutoff date (inner-history end when fitting the scale; history end when applying it); \textit{Author Momentum} substitutes a per-author metric (number of prior papers, $h$-index, or total citations at $t_p$) summed across each paper's authors. \textit{Linear Extrapolation} side-steps the inner split entirely, fitting a per-topic OLS line to monthly counts in the train period and integrating the predicted rate over the test period. Topic pairs with fewer than ten papers in both periods are excluded.

\subsection{Corpus Generation} \label{app:exp_details_corpus_generation}

\paragraph{Setup.}
The simulation is initialized from the final date of the train period and proceeds in daily steps for $365$ days. The number of papers per day is drawn from an empirical distribution of daily target-paper counts in the train period; per-paper author counts and influential-reference counts are drawn from analogous train-period distributions. New authors are introduced at the empirical first-time-author rate among train-period target authors and assigned fresh synthetic identifiers.

\paragraph{Composition.}
Coauthor and prior-work prediction both use one-shot \textit{Embedding Fusion} with GRIT, with author embeddings computed as the mean of each author's $10$ most recent papers. Title and abstract are then generated by GPT-5 (\texttt{gpt-5-2025-08-07}) conditioned on the predicted influential references. After each simulated day, generated papers are embedded with GRIT and added back to the literature state. Because they are added back, synthetic papers become eligible as influential references and as members of authors' publication histories for subsequent days, a compounding feedback structure that helps explain the novelty decay reported in Section~\ref{sec:corpus_generation}. Diversity and novelty estimates are aggregated over six independent rollouts to bound Monte Carlo variance in the reported confidence intervals.

\paragraph{Algorithm.} \label{app:corpus_generation_algorithm}
We provide pseudocode (Algorithm~\ref{alg:pseudocode_compact}) for the corpus generation algorithm we employ and describe in Section~\ref{sec:corpus_generation}.

\begin{algorithm}[!htb]
\caption{Corpus Generation}
\label{alg:pseudocode_compact}
\footnotesize
\begin{algorithmic}[1]
\REQUIRE $\mathbf{H}^{<t_0}$, rollout horizon $[t_0,t_f)$
\ENSURE $\mathbf{H}^{<t_f}$

\STATE $P_N, P_{|C|}, P_{|R|}, p_{\text{new}} \leftarrow$ \textsc{EstimateDist}($\mathbf{H}^{<t_0}$)

\FOR{$t = t_0$ \TO $t_f - 1$}
  \STATE $N \sim P_N$, $\mathcal{S}_t \leftarrow \emptyset$
  \FOR{$i = 1$ \TO $N$}
    \STATE $|C| \sim P_{|C|}$, $|R| \sim P_{|R|}$
    \STATE $C \leftarrow \textsc{SampleResearchTeam}(|C|, p_{\text{new}}, \mathbf{H}^{<t})$
    \STATE $R \leftarrow \textsc{SelectPriorWork}(C, |R|, \mathbf{H}^{<t})$
    \STATE $(\tau,\alpha) \leftarrow \textsc{GenerateTitleAbstract}(C, R, \mathbf{H}^{<t})$
    \STATE $\mathcal{S}_t \leftarrow \mathcal{S}_t \cup \{\textsc{Paper}(\tau,\alpha,C,R,t)\}$
  \ENDFOR
  \STATE $\mathbf{H}^{<t+1} \leftarrow \mathbf{H}^{<t} \cup \mathcal{S}_t$
\ENDFOR

\RETURN $\mathbf{H}^{<t_f}$
\end{algorithmic}
\end{algorithm}

\subsection{Computational Resources} \label{app:exp_details_compute}

\paragraph{Dataset construction.}
The pipeline runs on a single multi-core CPU machine in a few hours, except for S2AND author disambiguation (Stage 4), which takes up to a few days on a similar machine.

\paragraph{Topic labeling.}
Per-paper topic labels (GPT-5.4 over the $\sim$96k target papers, Appendix~\ref{app:dataset_topics}) complete in roughly an hour at 128-worker parallelism.

\paragraph{Embedding precomputation.}
GRIT-7B embeddings for all $\sim$465k corpus papers take a few hours on $2\times$H100 GPUs (one-time cost). GTR-T5-large and Specter2 are smaller and complete on a single H100 in comparable time.

\paragraph{Contribution generation.}
Finetuning OLMo~3~7B and Qwen3~8B on the train-period targets each takes roughly half a day on $4\times$H100 GPUs. Frontier-model evaluation (GPT-4o, GPT-o3, GPT-5, GPT-5.2, Claude Sonnet/Opus~4.5) consumes $\sim$5k chat-completion API calls per model on the test split. \followupmetric~scoring adds one GPT-5 judge call per generated paper; across all evaluated models this amounts to a few tens of thousands of judge calls.

\paragraph{Paper-anchored ranking tasks (collaborator, prior work, future combination).}
Static baselines (frequency heuristics, embedding ranking, hierarchical clustering) are CPU-bound and complete within minutes to a few hours per task on a multi-core machine. \textit{Embedding Fusion Proj.} adds a few hours of single-GPU training on an H100.

\paragraph{Citation count prediction.}
XGBoost training plus Bayesian hyperparameter search (Hyperopt, 100 evaluations) completes in roughly one CPU-hour.

\paragraph{Topic trend forecasting.}
Pure-CPU baselines that complete within minutes.

\paragraph{GPT-5 Agent.}
Each agent run completes within a few hours wall-clock at 128-worker parallelism. The contribution-generation agent runs over the full $n=5000$ test split (one $\leq$10-round agent loop per target); the three ranking-task agents (collaborator, prior work, future combination) each run on $n=500$. A 100-target calibration on contribution generation found that an agent run averages $3.9$ tool-call rounds and consumes roughly $9.7\times$ the prompt tokens, $1.7\times$ the completion tokens, and $\sim\!3\times$ the per-target API cost of the corresponding static GPT-5 baseline.

\paragraph{Autoresearch.}
Each of the four task-specific outer loops consumes a fixed 12-hour wall-clock budget within which a Claude Opus~4.7 agent iteratively writes and benchmarks candidate Python pipelines against the 1000-instance train-period dev set. Total: $\sim$48 hours of LLM-driven experimentation.

\paragraph{Corpus generation.}
The heaviest single experiment in the paper. We perform six independent 12-month rollouts; each rollout simulates roughly $250$ papers per day for 365 days, and each simulated paper triggers a collaborator-prediction call (CPU), a prior-work-selection call (CPU), and a GPT-5 contribution-generation API call. Across the six rollouts this amounts to roughly $5\times10^5$ GPT-5 contribution-generation calls.

\section{Additional Analyses} \label{app:analyses}

\subsection{Contribution Generation}
\subsubsection{Effect of Pretraining Corpus Contamination}
\label{app:llm_cutoff_ablation}
Table~\ref{tab:lacer_cutoff} compares mean LACER scores in the month immediately before and after each model’s reported knowledge cutoff date. We observe modest changes in absolute scores and no changes in relative model ordering, suggesting that any cutoff-related effects are small relative to the performance differences reported in the main results.

\begin{table*}[!htb]                                                                                                                
  \centering                                                                                                                      
  \caption{Mean LACER scores (over 1 month) before and after model knowledge cutoff dates.}                                                      
  \label{tab:lacer_cutoff}                                                                                                        
  \begin{tabular}{lccc}                                                                                                           
  \toprule                                                                                                                        
  Model & Cutoff Date & Pre-cutoff & Post-cutoff \\                                                                               
  \midrule                                                                                                                        
  Claude Sonnet 4.5 & Jan 31, 2025 & 4.900 & 5.062 \\                                                                             
  Claude Opus 4.5 & May 31, 2025 & 5.054 & 5.008 \\                                                                               
  GPT-5.2 & Aug 31, 2025 & 5.706 & 5.595 \\                                                                                       
  \bottomrule                                                                                                                     
  \end{tabular}                                                                                                                   
\end{table*}

\subsubsection{Further Task Analyses}
\label{app:followup_prediction_analysis}
\begin{figure*}[!htb]
    \centering
    \begin{subfigure}[t]{0.32\linewidth}
        \centering
        \includegraphics[width=\linewidth]{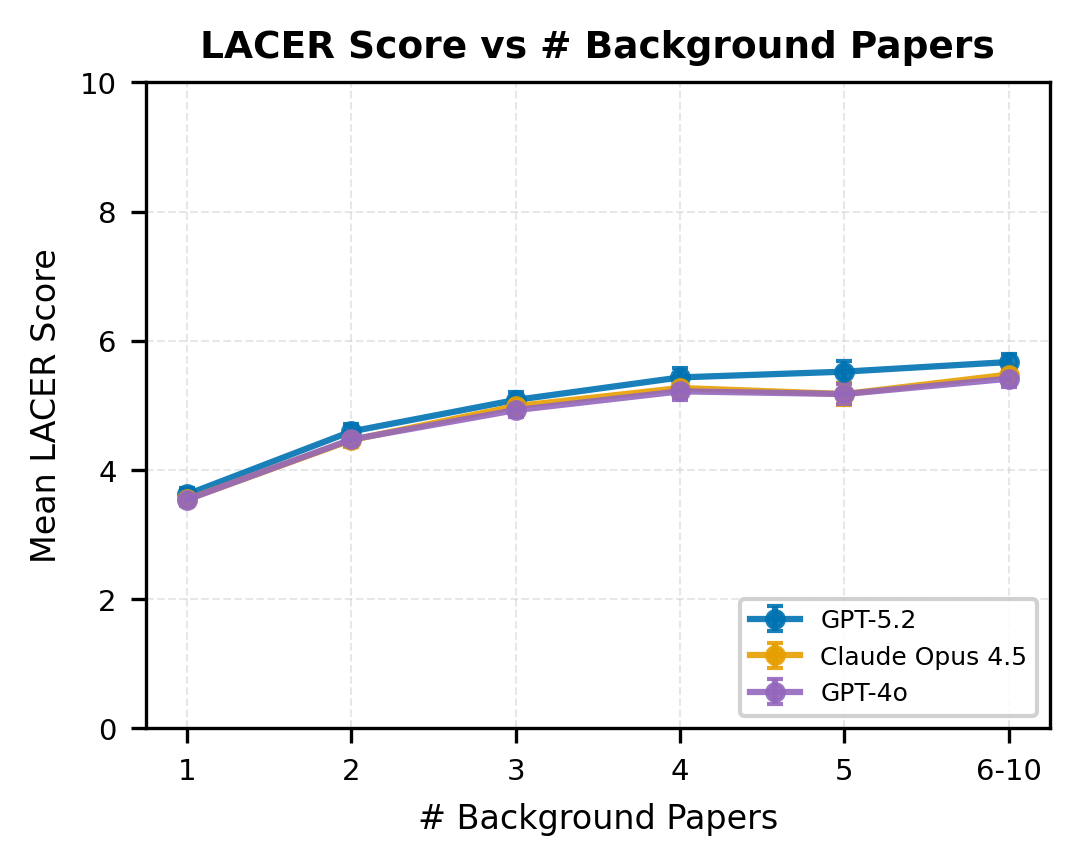}
        \caption{Prediction is easier in instances with more influential references.}

    \end{subfigure}
    \hfill
    \begin{subfigure}[t]{0.32\linewidth}
        \centering
        \includegraphics[width=\linewidth]{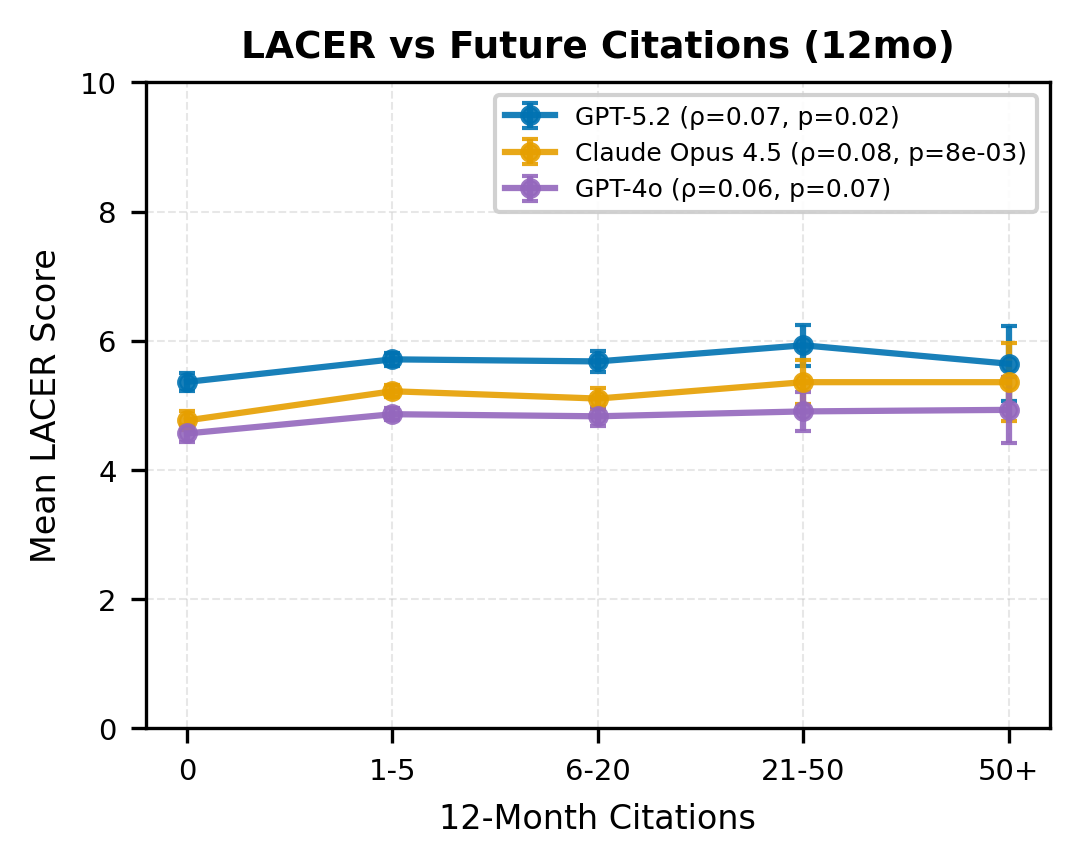}
        \caption{Prediction difficulty appears agnostic to future impact of paper.}
    \end{subfigure}
    \hfill
    \begin{subfigure}[t]{0.32\linewidth}
        \centering
        \includegraphics[width=\linewidth]{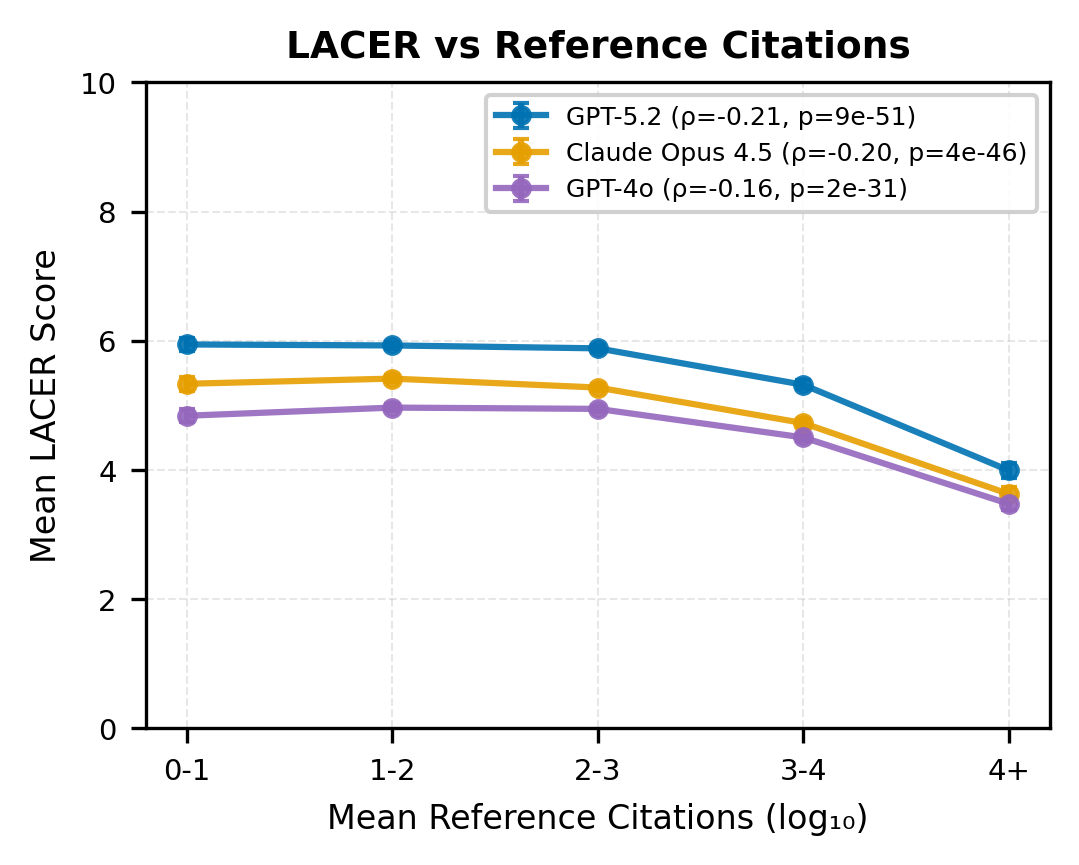}
        \caption{Papers whose influential references have fewer citations are easier to predict.}
    \end{subfigure}
    \hfill
    \begin{subfigure}[t]{0.32\linewidth}
        \centering
        \includegraphics[width=\linewidth]{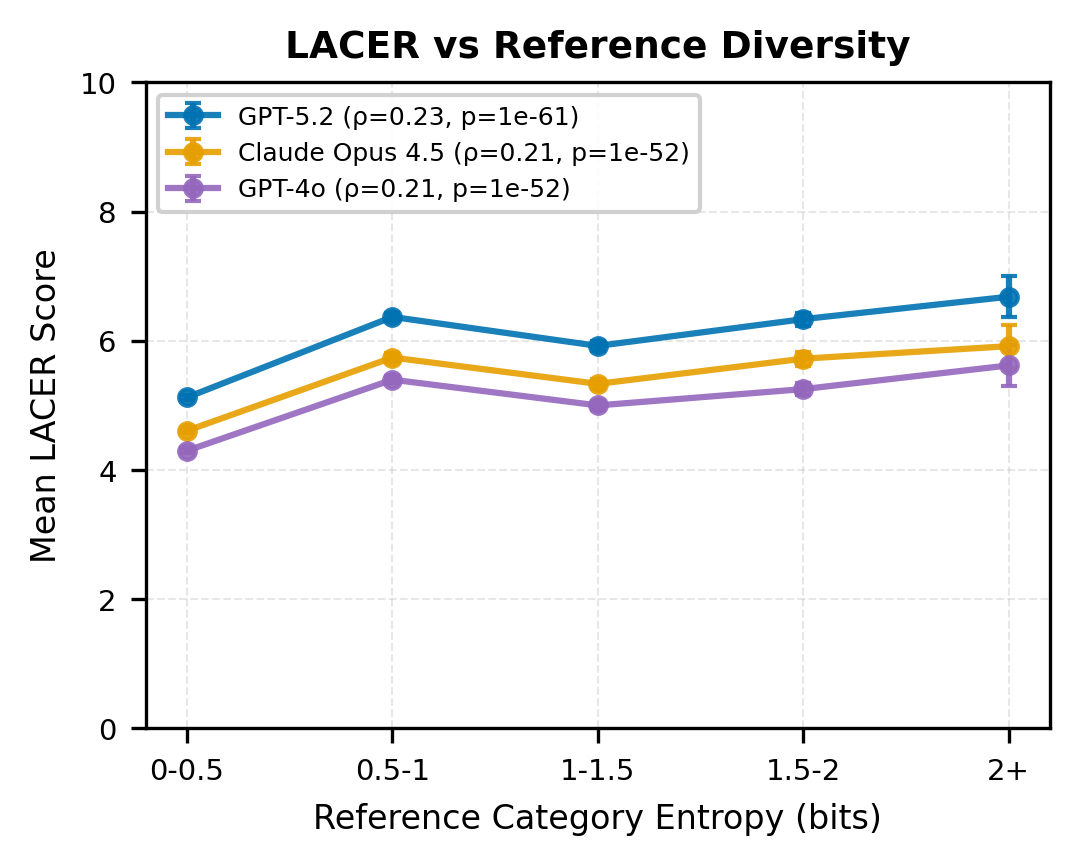}
        \caption{Papers whose influential references are more diverse in topics appear easier to predict.}
    \end{subfigure}
    \hfill
    \begin{subfigure}[t]{0.32\linewidth}
        \centering
        \includegraphics[width=\linewidth]{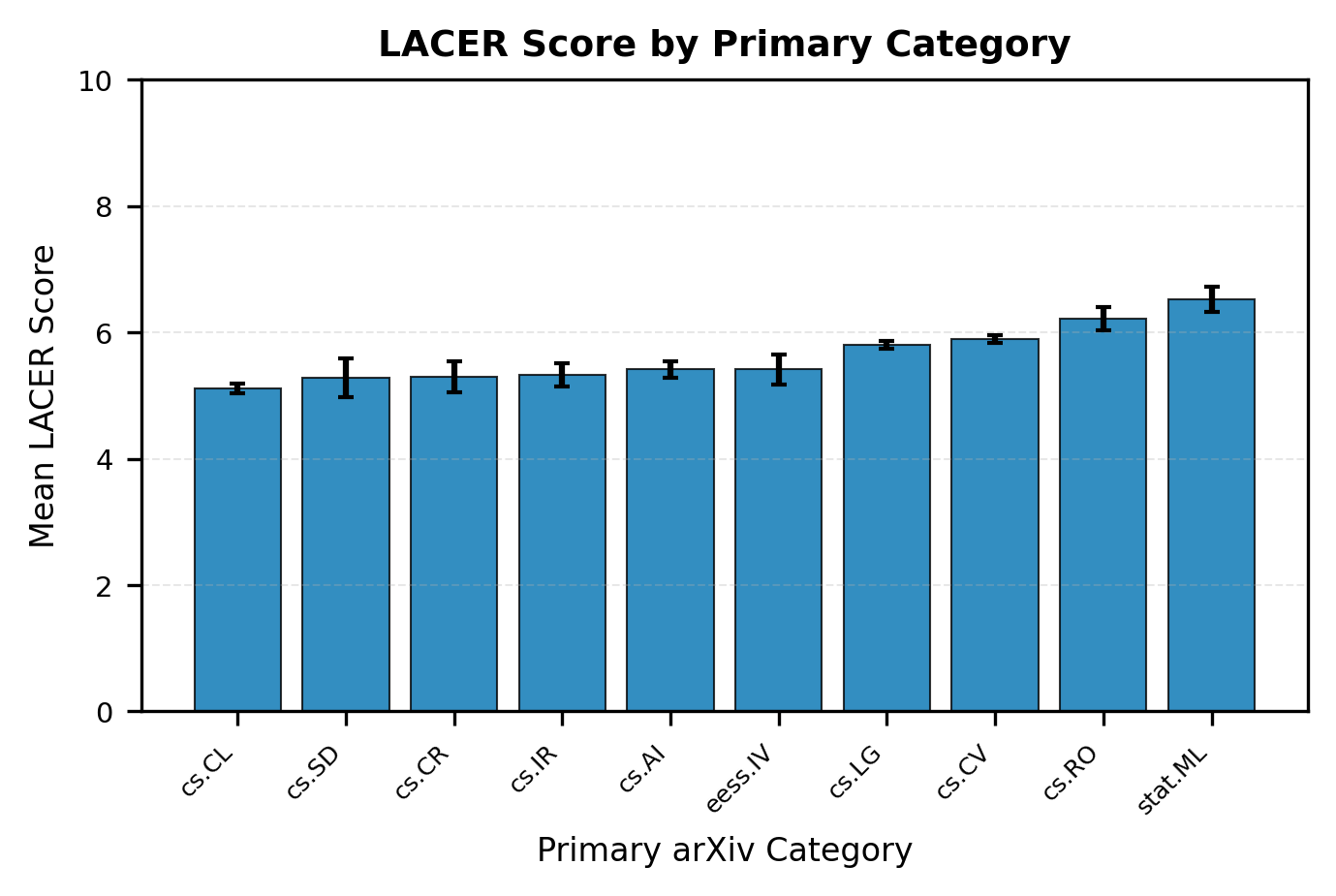}
        \caption{There is statistically significant variation in LLMs' to predict work corresponding to different ArXiv topics with ``Computation and Language (cs.CL)" papers being the hardest to predict among the shown topics and ``Machine Learning (stat.ML)" papers being the easiest.}
    \end{subfigure}
    \hfill
    \begin{subfigure}[t]{0.32\linewidth}
        \centering
        \includegraphics[width=\linewidth]{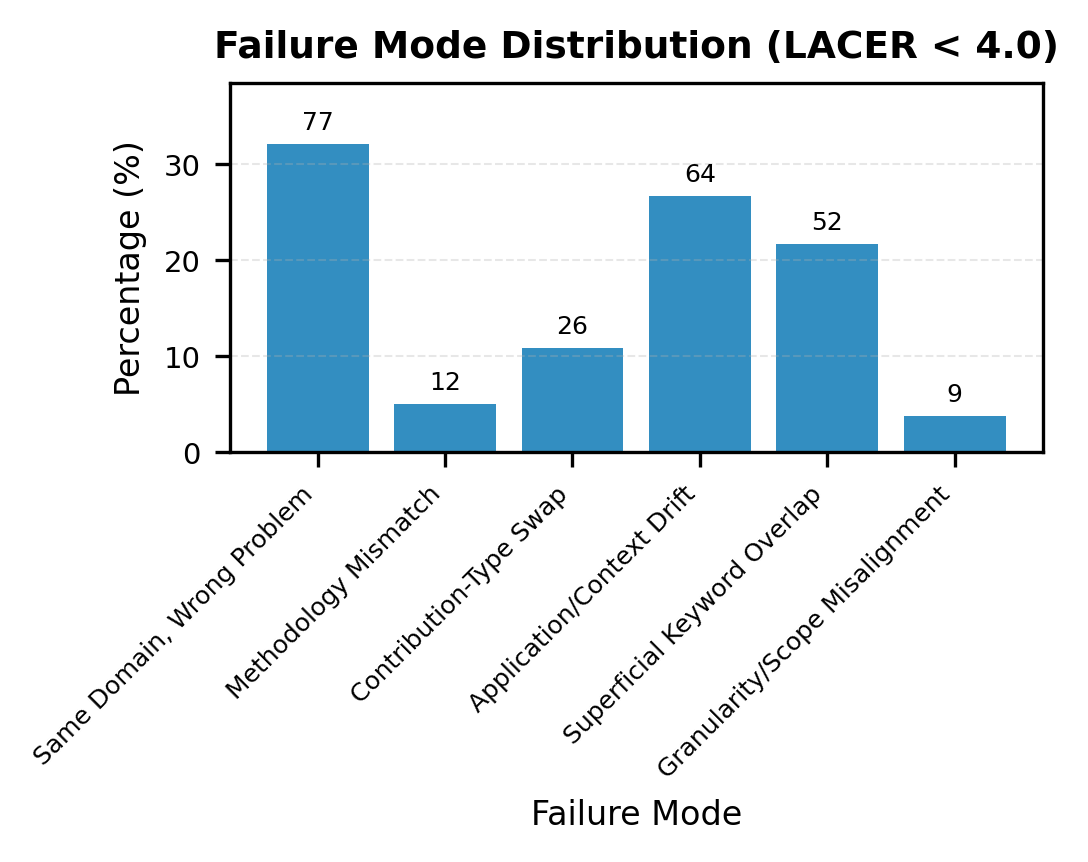}
        \caption{Prediction Failure Modes}
    \end{subfigure}

\caption{Contribution Generation}
\label{fig:followup_work_analyses}
\end{figure*}

Figure~\ref{fig:followup_work_analyses}(a) shows that contribution generation becomes easier as more influential references are available, consistent with additional contextual signal improving conceptual alignment. Figure~\ref{fig:followup_work_analyses}(b) indicates that LACER scores are largely insensitive to a paper’s future citation impact, suggesting that predictive difficulty is decoupled from downstream popularity. Figure~\ref{fig:followup_work_analyses}(c) shows that papers whose influential references have lower average citation counts are easier to predict. This is consistent with highly cited prior work being useful to a wide application space. Figure~\ref{fig:followup_work_analyses}(d) shows that higher topical diversity among influential references is associated with improved prediction performance, perhaps indicating fewer ``valid" ways in which diverse work can be combined (given that the subset can in fact be combined). Figure~\ref{fig:followup_work_analyses}(e) reveals systematic variation across arXiv categories, with computation-and-language papers exhibiting lower scores and machine-learning papers higher scores. Figure~\ref{fig:followup_work_analyses}(f) summarizes common failure modes, dominated by domain mismatch and application-context drift rather than surface-level keyword errors.\footnote{We categorize these failure modes by employing prompting GPT-5.2 with a sample of 240 low-scoring generated abstracts along with their corresponding ground truths and instructing it to study and categorize them into common failure modes.}

\subsubsection{Effect of Influential Reference Choice} \label{app:contribution_generation_reference_choice}
We evaluate contribution generation across three definitions of a target paper's prior work --- Semantic Scholar's highly influential references~\citep{ValenzuelaEscarcega2015IdentifyingMC}, all of its references, and \emph{impact-revealing references}~\citep{arnaout2025indepthresearchimpactsummarization} --- using GPT-5, GPT-5.2, and Claude Opus 4.5 (Table~\ref{tab:contribution_generation_reference_choice}). In absolute terms, larger reference sets yield higher \followupmetric, but this trend is largely a context-size effect: when each paper's reference set is randomly subsampled to match the size of the highly-influential set, both alternatives perform substantially worse than the default. The same pattern holds when all references are subsampled to the size of the impact-revealing set. Thus, at matched context budget, Semantic Scholar's classifier selects a stronger subset of references for contribution generation than random subsamples of the larger reference sets. Relative model rankings are stable across all six conditions.

\begin{table}[!htb]
\centering
\caption{Contribution generation \followupmetric~under alternative definitions of a target paper's prior work ($n=984$). Subsampled rows random-subsample each paper's references down to the count of the reference type given in parentheses, isolating reference-set selection from reference-set size.}
\footnotesize
\begin{tabular}{lcccc}
\toprule
\textbf{Reference Set} & \textbf{Ref. Count} & \textbf{GPT-5} & \textbf{GPT-5.2} & \textbf{Claude Opus 4.5} \\
\midrule
Highly influential references                                  & $3.02$  & $5.62$ & $5.50$ & $4.78$ \\
\midrule
Impact-revealing references                                    & $9.89$  & $6.45$ & $6.35$ & $5.46$ \\
All references                                                 & $35.16$ & $6.89$ & $6.63$ & $6.22$ \\
\midrule
Impact-revealing references (sub. to highly-influential count) & $3.02$  & $5.11$ & $5.08$ & $3.49$ \\
All references (sub. to highly-influential count)              & $3.02$  & $4.74$ & $4.68$ & $4.14$ \\
All references (sub. to impact-revealing count)                & $9.89$  & $6.19$ & $6.00$ & --- \\
\bottomrule
\end{tabular}
\label{tab:contribution_generation_reference_choice}
\end{table}

\subsection{Collaborator Prediction} \label{app:analyses_collaborator_prediction}
\subsubsection{Effect of Seed Author Choice}
\label{app:sub_seed_author}
We find that the choice of seed author in the collaborator prediction task does not affect the relative ordering of baseline performance. This is a non-trivial result, as research team formation and collaborator discovery may be governed by different mechanisms for authors at different career stages or seniority levels. However, the baselines we evaluate primarily operate on order-invariant features of the observed co-authorship graph (e.g., local neighborhoods and aggregated publication histories), so it appears changing the seed largely only shifts the strength of the underlying collaboration signal without favoring any of the baselines over others. 

\begin{table}[!htb]
\centering
\caption{Effect of seed author choice on collaborator prediction performance (nDCG) (n=1000). The relative performance order among baselines remains unchanged.}
\label{tab:author_seed_ablation}
\setlength{\tabcolsep}{6pt}
\begin{tabular}{lcccc}
\toprule
\textbf{Baseline} & \textbf{First} & \textbf{Last} & \textbf{Random} & \textbf{Argmax h-index} \\
\midrule
Frequency              & 0.38 & 0.34 & 0.37 & 0.26 \\
Rank Fusion (GRIT)    & 0.15 & 0.12 & 0.14 & 0.10 \\
Embedding Fusion (GRIT) & 0.29 & 0.23 & 0.26 & 0.18 \\
Embedding Fusion + Projection (GRIT) & 0.24 & 0.22 & 0.23 & 0.18 \\
\bottomrule
\end{tabular}
\end{table}

\subsubsection{Further Task Analyses}

\begin{figure*}[!htb]
    \centering
    \begin{subfigure}[t]{0.47\linewidth}
        \centering
        \includegraphics[width=\linewidth]{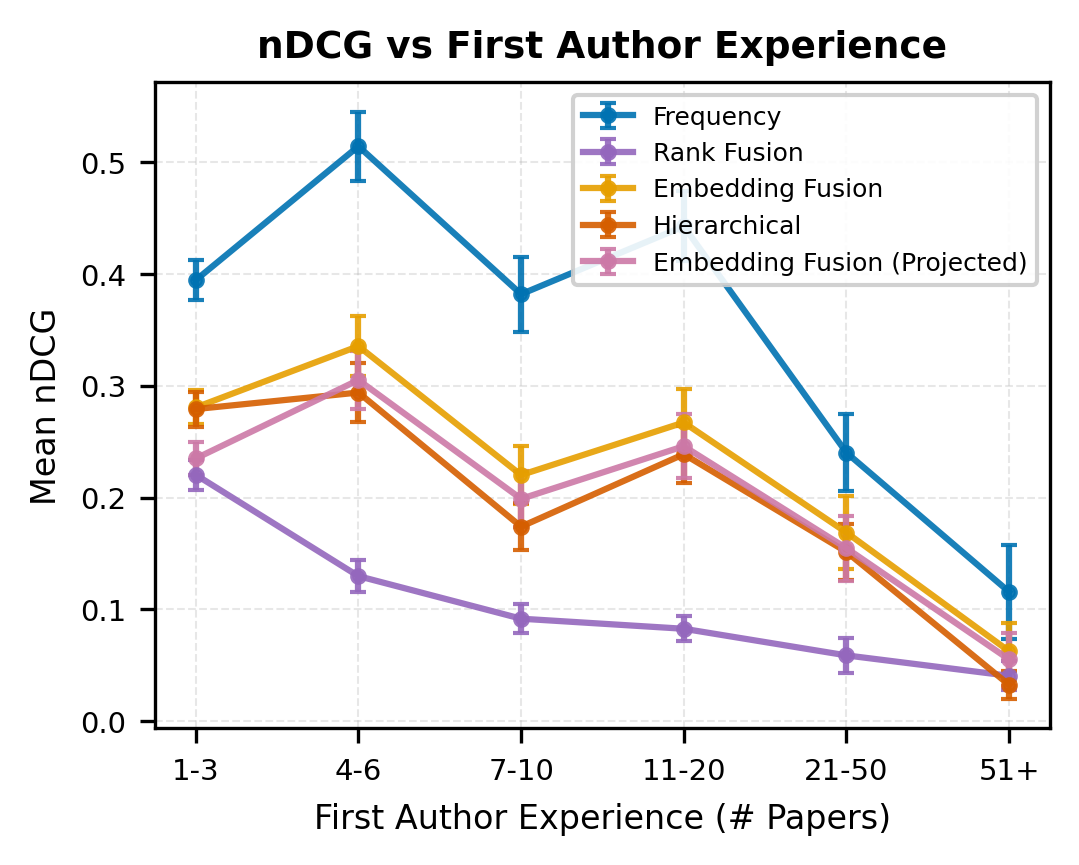}
        \caption{nDCG scores as author profiles become crowded with more publications.}
    \end{subfigure}
    \hfill
    \begin{subfigure}[t]{0.47\linewidth}
        \centering
        \includegraphics[width=\linewidth]{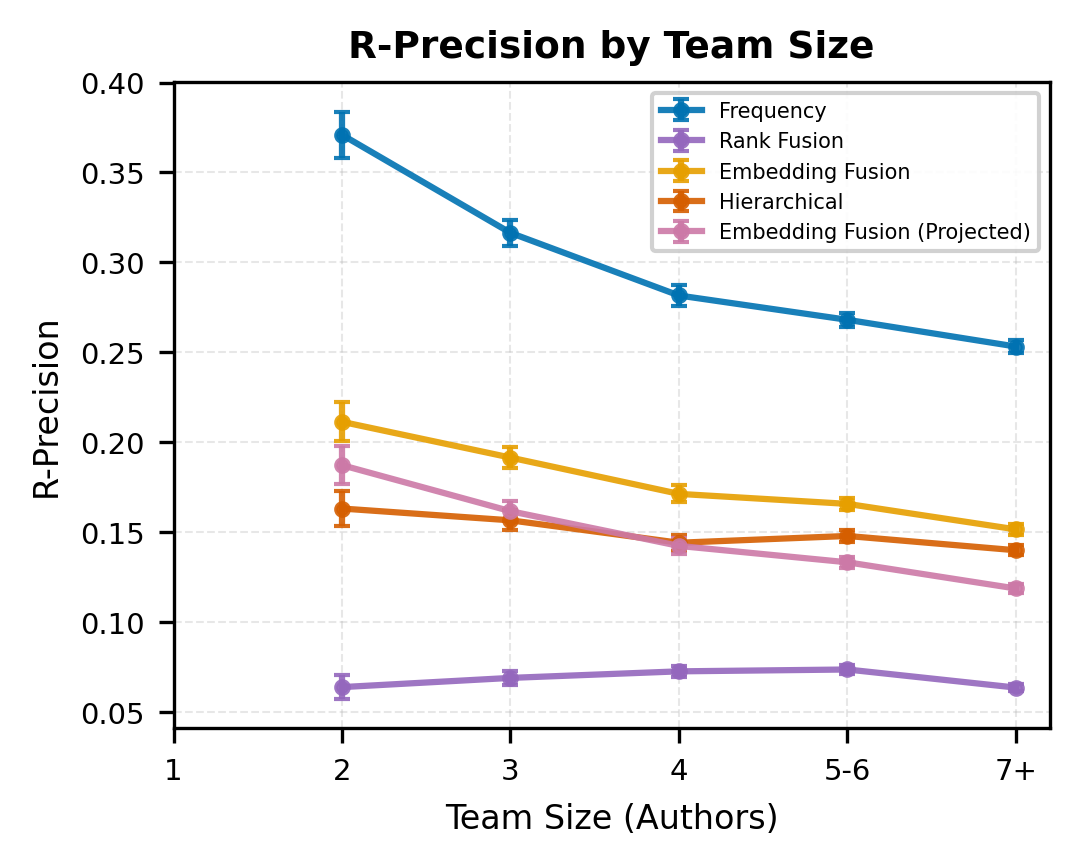}
        \caption{Larger teams are harder to correctly predict.}
    \end{subfigure}
    \caption{Collaborator Prediction}
    \label{fig:collaborator_prediction_analyses}
\end{figure*}

Figure~\ref{fig:collaborator_prediction_analyses} analyzes collaborator prediction performance across two sources of variation. Panel (a) shows that nDCG decreases as the first author’s publication history grows, indicating that larger and more crowded collaboration neighborhoods dilute the signal available to frequency- and embedding-based baselines. Panel (b) shows that R-precision declines monotonically with team size, reflecting the increasing combinatorial difficulty of recovering all collaborators as the target set grows.

\subsubsection{Hit Rate by Familiarity} \label{app:collaborator_familiarity}
\begin{figure}[!htb]
    \centering
    \includegraphics[width=\linewidth,height=0.25\textheight,keepaspectratio]{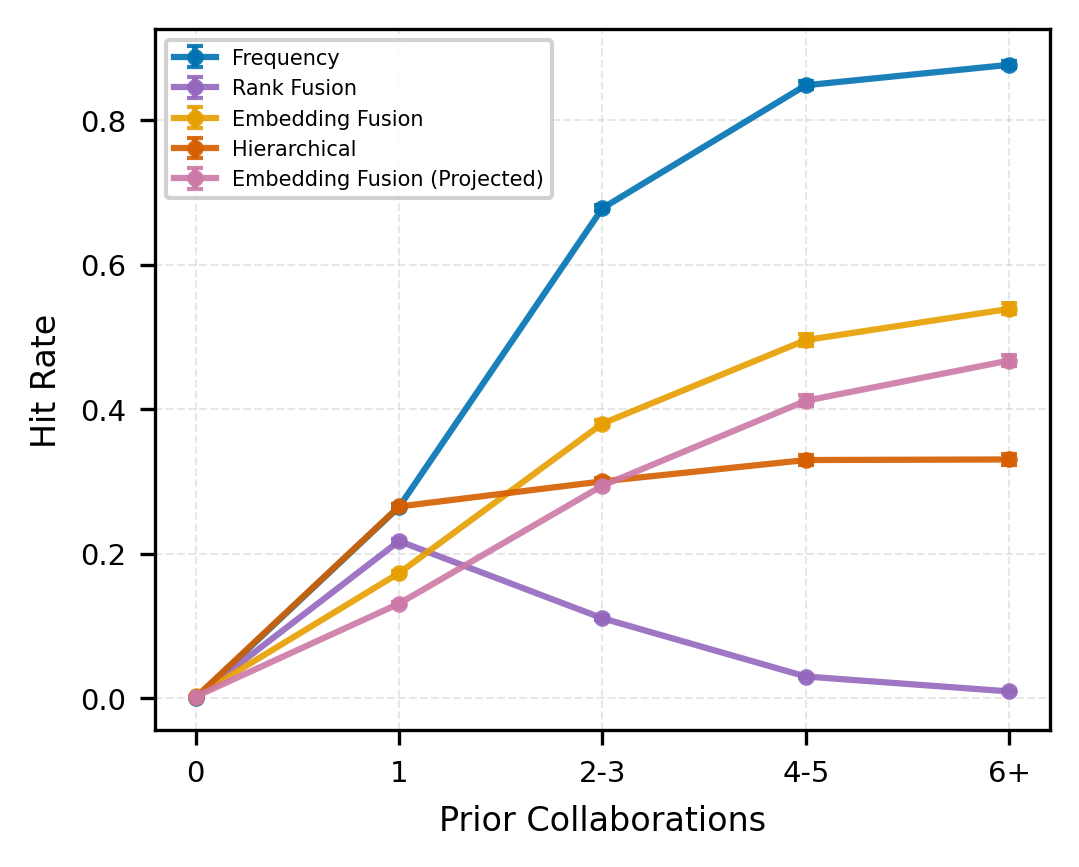}
    \caption{Collaborator prediction hit rate --- the fraction of top-$R$ predicted authors that are among the $R$ ground-truth authors --- as the number of prior co-authorships between candidate and seed varies. All baselines exhibit near-zero performance when predicting first-time collaborators.}
    \label{fig:task_coauthor_prediction_analysis_rprecision_by_familiarity}
\end{figure}

Figure~\ref{fig:task_coauthor_prediction_analysis_rprecision_by_familiarity} reports collaborator prediction hit rate as a function of how often the candidate has previously co-authored with the seed. All baselines exhibit near-zero hit rates when predicting first-time collaborators, indicating that current methods recover repeat-collaboration structure but cannot anticipate genuinely new relationships.

\subsubsection{Effect of Embedding Choice} \label{app:collaborator_embedding_choice}
Table~\ref{tab:collaborator_embedding_choice} reports collaborator prediction baselines across GTR, Specter2, and GRIT embeddings. GRIT yields the strongest results for Embedding Fusion, while Hierarchical Clustering is largely insensitive to the choice of embedding.

\begin{table}[t]
\centering
\caption{Effect of embedding choice on collaborator prediction baselines.}
\label{tab:collaborator_embedding_choice}
\small
\setlength{\tabcolsep}{4pt}
\begin{tabular}{lcc}
\toprule
\textbf{Method (Embed)} & \textbf{nDCG} & \textbf{R-Prec} \\
\midrule
Rank Fusion (GTR) & $0.15$ & $0.06$ \\
Rank Fusion (Specter2) & $0.11$ & $0.05$ \\
Rank Fusion (GRIT) & $0.17$ & $0.08$ \\
\midrule
Emb. Fusion (GTR) & $0.24$ & $0.16$ \\
Emb. Fusion (Specter2) & $0.19$ & $0.11$ \\
Emb. Fusion (GRIT) & $\mathbf{0.28}$ & $\mathbf{0.18}$ \\
\midrule
Hier. Clustering (GTR) & $0.25$ & $0.15$ \\
Hier. Clustering (Specter2) & $0.25$ & $0.14$ \\
Hier. Clustering (GRIT) & $0.25$ & $0.15$ \\
\bottomrule
\end{tabular}
\end{table}

\subsection{Prior Work Selection}
\label{app:priorwork_prediction_analysis}
\subsubsection{Effect of Influential Reference Choice} \label{app:priorwork_reference_choice}

In addition to Semantic Scholar’s \emph{highly influential} references~\citep{ValenzuelaEscarcega2015IdentifyingMC}, we evaluate two alternative definitions of influential prior work: (i) using the full set of references cited by each paper, and (ii) using \emph{impact-revealing references}~\citep{arnaout2025indepthresearchimpactsummarization}. As shown in Table~\ref{tab:key_reference_ablations}, neither alternative yields a dramatic improvement in prediction performance over the default influential-reference definition\footnote{These results are reported on an earlier snapshot of the corpus; we expect the updated release to preserve the relative trends across reference definitions, even if absolute values shift.
}. However, both incur substantially higher computational and data costs: including all references dramatically expands the set of companion papers and causes the historical corpus $H_{<t}$ to balloon, while computing impact-revealing references requires repeated calls to commercial LLM APIs. These results motivate our use of Semantic Scholar influential references as a practical trade-off between predictive signal and scalability.

\begin{table}[!htb]
\centering
\caption{Prior work selection performance (n=1000) across reference types. Standard deviations are shown in subscript parentheses.}
\footnotesize
\begin{tabular}{lcccc}
\toprule
\textbf{Reference Type} & \textbf{Reference Count} & \textbf{nDCG} & \textbf{R-Prec} \\
\midrule
S2 Highly Influential & $5.43_{(0.12)}$ & $4.2_{(0.4)}$ & $3.0_{(0.3)}$ \\
All References & $34.04_{(0.63)}$ & $7.6_{(0.3)}$ & $4.6_{(0.2)}$ \\
Impact-Revealing & $10.65_{(0.22)}$ & $5.7_{(0.3)}$ & $3.6_{(0.3)}$ \\
\bottomrule
\end{tabular}
\label{tab:key_reference_ablations}
\end{table}

\subsubsection{Further Task Analyses}
\begin{figure*}[!htb]
    \centering
    \begin{subfigure}[t]{0.31\linewidth}
        \centering
        \includegraphics[width=\linewidth]{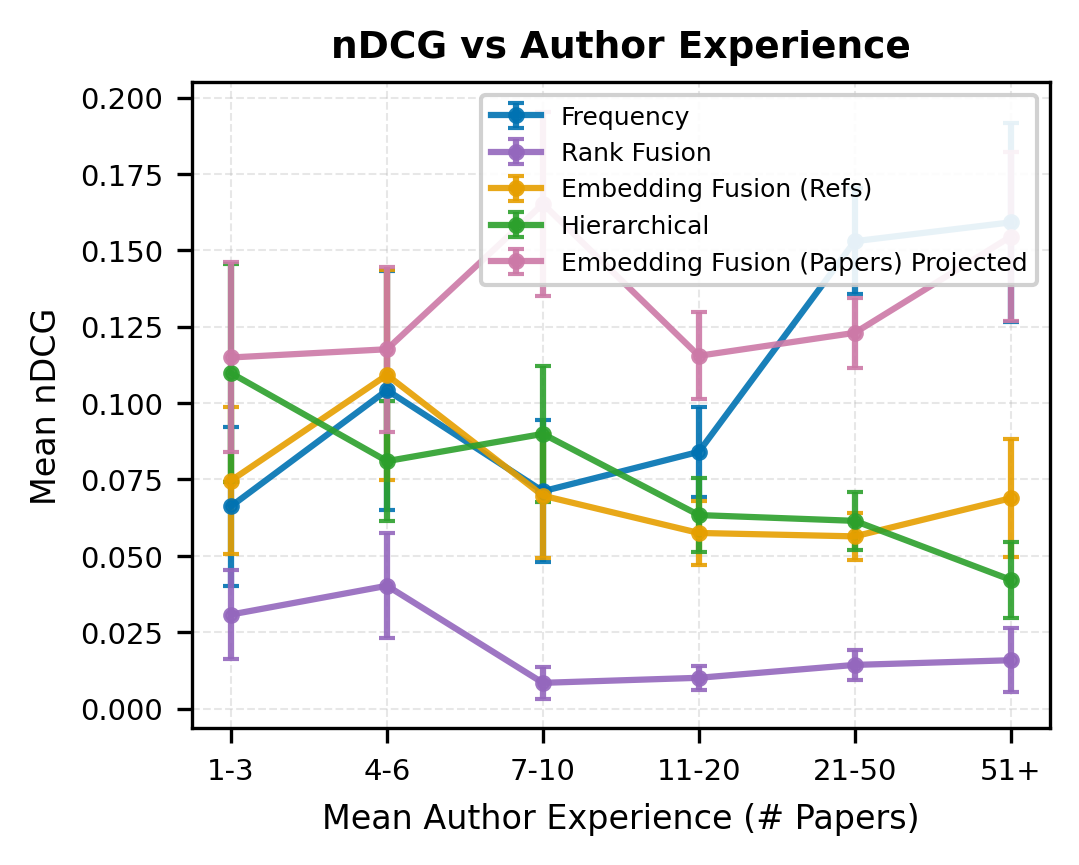}

    \end{subfigure}
    \hfill
    \begin{subfigure}[t]{0.31\linewidth}
        \centering
        \includegraphics[width=\linewidth]{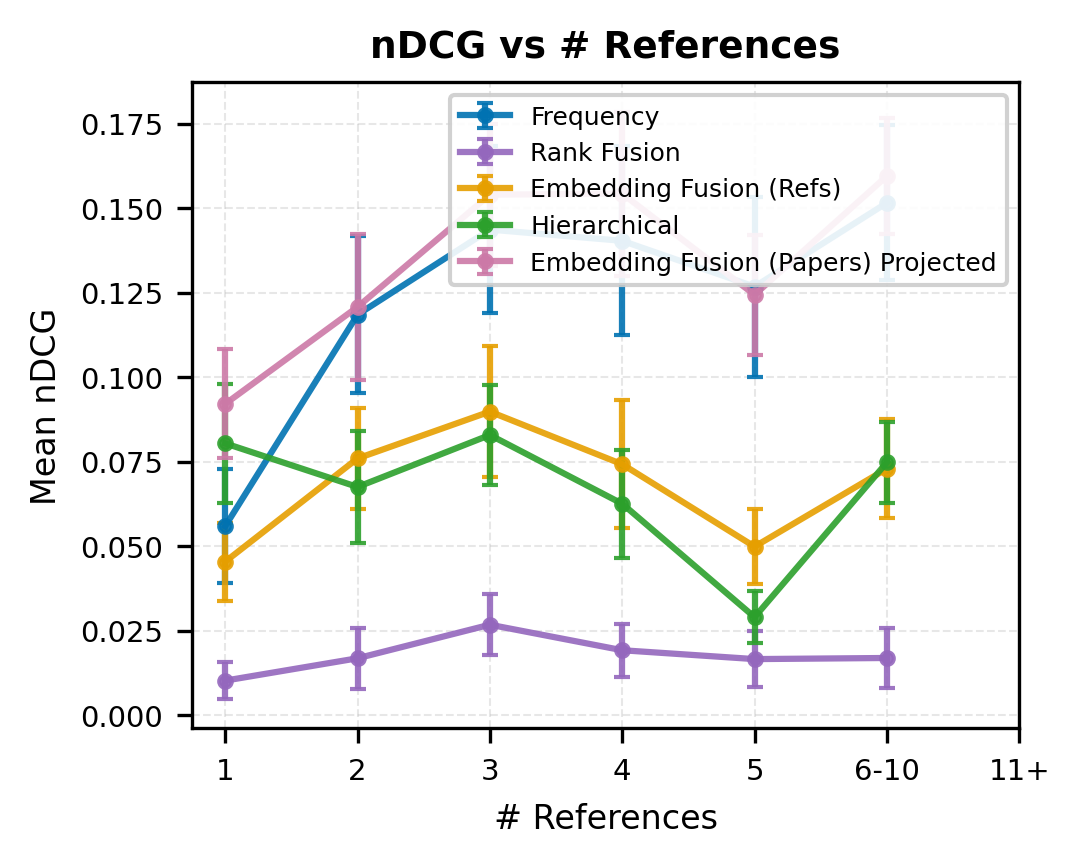}

    \end{subfigure}
    \hfill
    \begin{subfigure}[t]{0.31\linewidth}
        \centering
        \includegraphics[width=\linewidth]{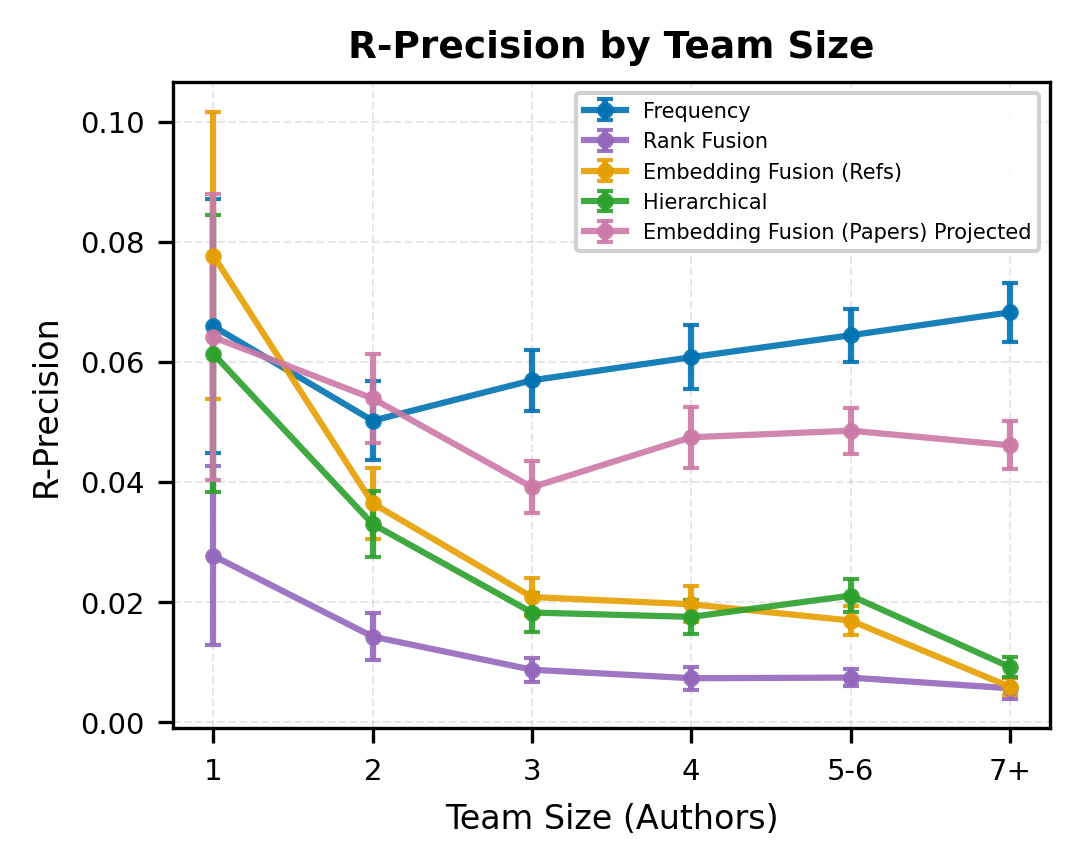}

    \end{subfigure}

    \caption{Prior Work Prediction}
    \label{fig:task_priorwork_prediction_analysis}
\end{figure*}

Figure~\ref{fig:task_priorwork_prediction_analysis} presents diagnostic analyses of prior work selection performance across author experience, number of references, and team size. Across all three views, we observe limited and non-monotonic variation in nDCG and R-precision across baselines, suggesting that no single factor strongly governs performance in isolation.

\subsubsection{Hit Rate by Familiarity} \label{app:priorwork_familiarity}
\begin{figure}[!htb]
    \centering
    \includegraphics[width=\linewidth,height=0.25\textheight,keepaspectratio]{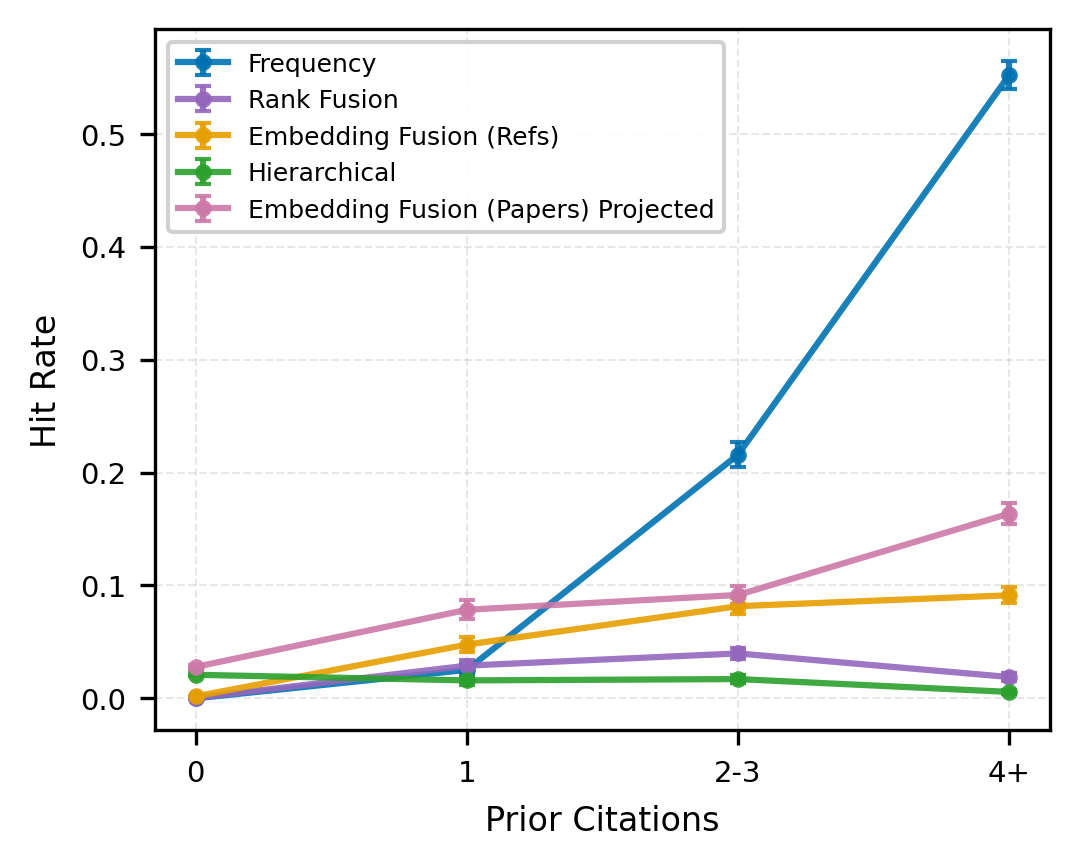}
    \caption{Prior work prediction hit rate as a function of how often a target paper's authors have previously cited the candidate reference. Methods struggle to predict novel references; \textit{Frequency} dominates for previously-cited works.}
    \label{fig:task_priorwork_prediction_analysis_rprecision_by_familiarity}
\end{figure}

Figure~\ref{fig:task_priorwork_prediction_analysis_rprecision_by_familiarity} reports prior work selection hit rate as a function of how often the target paper's authors have previously cited the candidate reference. Aside from \textit{Frequency}, which dominates in high-familiarity regimes, embedding-based methods achieve low hit rates across all buckets and degrade further for less-precedented references.

\subsubsection{Effect of Embedding Choice} \label{app:priorwork_embedding_choice}
Table~\ref{tab:priorwork_embedding_choice} reports prior work selection baselines across GTR, Specter2, and GRIT embeddings. GRIT consistently outperforms the other two for Embedding Fusion, while Rank Fusion and Hierarchical Clustering are largely insensitive to the choice of embedding.

\begin{table}[t]
\centering
\caption{Effect of embedding choice on prior work selection baselines.}
\label{tab:priorwork_embedding_choice}
\small
\setlength{\tabcolsep}{4pt}
\begin{tabular}{lcc}
\toprule
\textbf{Method (Embed)} & \textbf{nDCG} & \textbf{R-Prec} \\
\midrule
Rank Fusion (GTR) & $0.03$ & $0.01$ \\
Rank Fusion (Specter2) & $0.02$ & $0.01$ \\
Rank Fusion (GRIT) & $0.02$ & $0.01$ \\
\midrule
Emb. Fusion (GTR) & $0.05$ & $0.02$ \\
Emb. Fusion (Specter2) & $0.07$ & $0.03$ \\
Emb. Fusion (GRIT) & $0.11$ & $0.05$ \\
\midrule
Hier. Clustering (GTR) & $0.06$ & $0.02$ \\
Hier. Clustering (Specter2) & $0.07$ & $0.02$ \\
Hier. Clustering (GRIT) & $0.06$ & $0.02$ \\
\bottomrule
\end{tabular}
\end{table}

\subsection{Future Combination Prediction}
\subsubsection{Effect of Embedding Choice} \label{app:cocitation_embedding_choice}
Table~\ref{tab:cocitation_embedding_choice} reports future combination prediction baselines across GTR, Specter2, and GRIT embeddings. GRIT yields the strongest results for both Target Similarity and Reference Similarity, with Reference Similarity outperforming Target Similarity at every embedding choice.

\begin{table}[t]
\centering
\caption{Effect of embedding choice on future combination prediction.}
\label{tab:cocitation_embedding_choice}
\small
\setlength{\tabcolsep}{4pt}
\begin{tabular}{lcc}
\toprule
\textbf{Method (Embed)} & \textbf{nDCG} & \textbf{R-Prec} \\
\midrule
Target Similarity (GTR) & $0.14$ & $0.05$ \\
Target Similarity (Specter2) & $0.21$ & $0.08$ \\
Target Similarity (GRIT) & $0.26$ & $0.11$ \\
\midrule
Reference Similarity (GTR) & $0.22$ & $0.12$ \\
Reference Similarity (Specter2) & $0.26$ & $0.12$ \\
Reference Similarity (GRIT) & $0.31$ & $0.16$ \\
\bottomrule
\end{tabular}
\end{table}

\subsection{Citation Count Prediction}
\subsubsection{Further Task Analyses}
\label{app:impact_prediction_analysis}

\begin{figure}[!htb]
    \centering
    \begin{subfigure}[t]{0.48\linewidth}
        \centering
        \includegraphics[width=\linewidth]{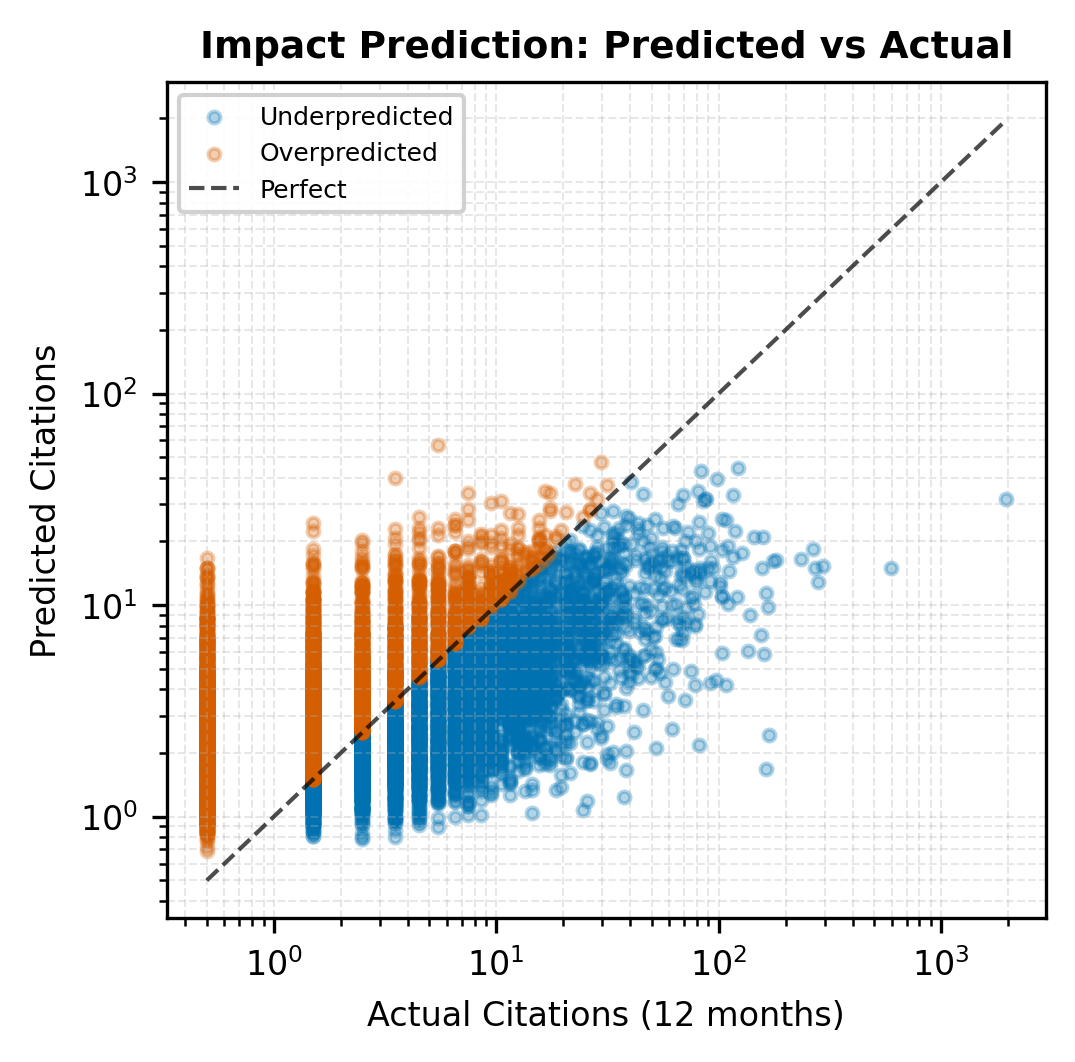}
        \caption{Scatter plot of citation count predictions (XGBoostRegressor; all features).}
        \label{fig:task_impact_prediction_analysis_prediction_scatter}
    \end{subfigure}
    \hfill
    \begin{subfigure}[t]{0.48\linewidth}
        \centering
        \includegraphics[width=\linewidth]{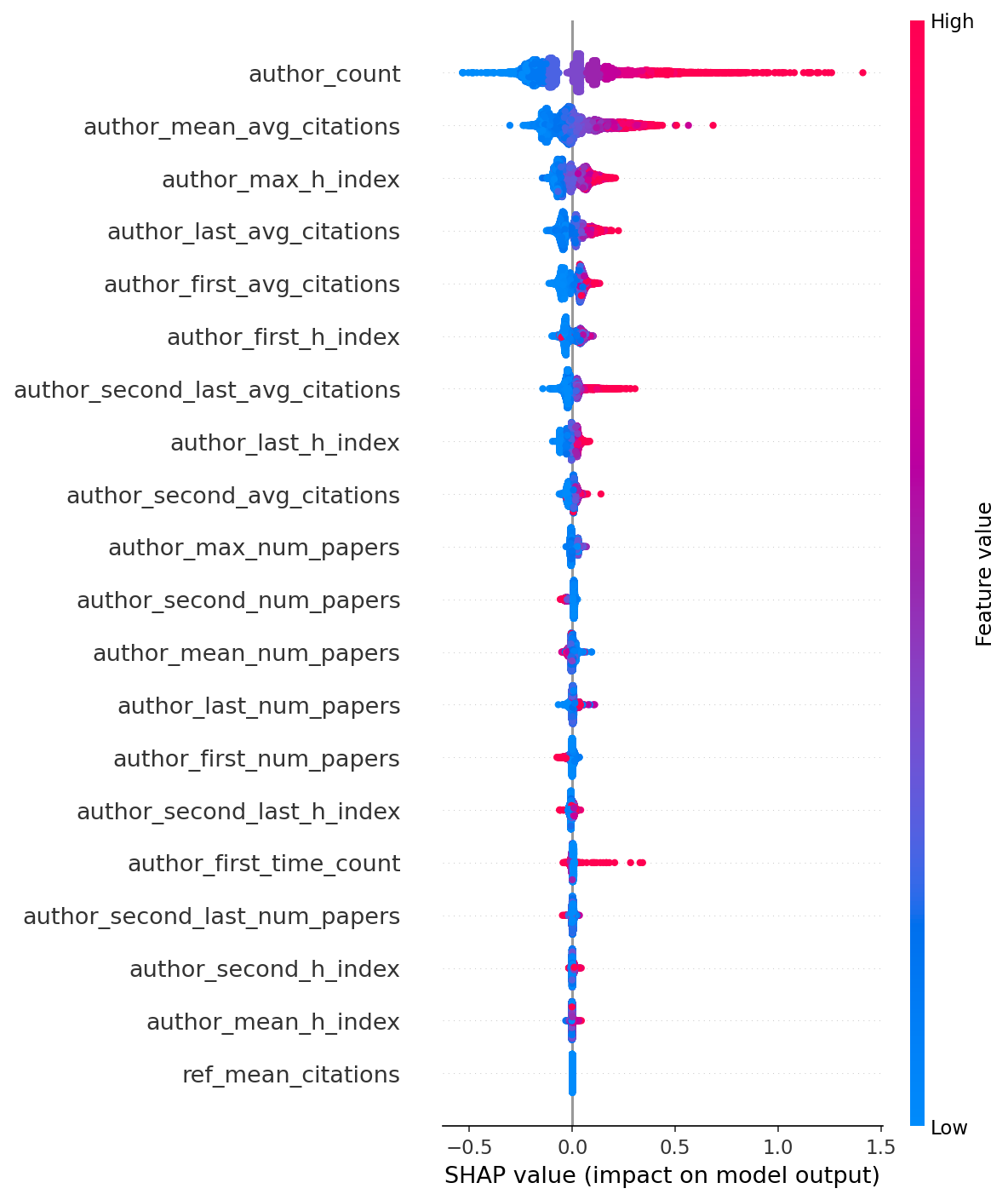}
        \caption{SHAP values for the XGBoost regressor trained over author- and influential-reference-related numerical metadata features.}
        \label{fig:task_impact_prediction_analysis_shap_summary_plot}
    \end{subfigure}
    \caption{Citation count prediction analysis: (a) scatter plot of predicted vs.\ true citation counts using all features, and (b) SHAP summary of feature attributions for XGBoostRegressor trained over bibliometric features.}
    \label{fig:task_impact_prediction_analysis_combined}
\end{figure}

Figure~\ref{fig:task_impact_prediction_analysis_combined} analyzes citation count prediction using an XGBoost regressor trained over the full feature set described in Section~\ref{sec:experiments_impact}. Panel (a) plots predicted versus true 12-month citation counts and exhibits clear heteroscedasticity: variance in prediction error increases with citation magnitude, indicating that highly cited papers are systematically harder to predict than low-impact papers. Panel (b) summarizes feature attributions via SHAP for the bibliometric features we use in Section~\ref{sec:experiments_impact}.

\subsubsection{Effect of Embedding Choice} \label{app:impact_embedding_choice}
Table~\ref{tab:impact_embedding_choice} reports Target Text citation count prediction baselines across GTR, Specter2, and GRIT embeddings. GRIT yields the strongest performance across all metrics.

\begin{table*}[t]
\centering
\footnotesize
\caption{Effect of embedding choice on citation count prediction baselines.}
\label{tab:impact_embedding_choice}
\setlength{\tabcolsep}{6pt}
\begin{tabular}{lccccc}
\toprule
\textbf{Baseline}
& \textbf{MAE}
& \textbf{MAE (log)}
& \textbf{Pearson}
& \textbf{Pearson (log)}
& \textbf{Spearman} \\
\midrule
Target Text (GTR)                 & $4.83$ & $0.74$ & $0.18$ & $0.40$ & $0.38$ \\
Target Text (Specter2)            & $4.78$ & $0.73$ & $0.20$ & $0.45$ & $0.42$ \\
Target Text (GRIT)                & $4.67$ & $0.71$ & $0.29$ & $0.49$ & $0.46$ \\
\bottomrule
\end{tabular}
\end{table*}

\subsection{Corpus Generation}
\subsubsection{Further Task Analyses}
\label{app:sub_corpus_gen_analysis}
\begin{figure*}[!htb]
    \centering
    \begin{subfigure}[t]{0.48\linewidth}
        \centering
        \includegraphics[width=\linewidth]{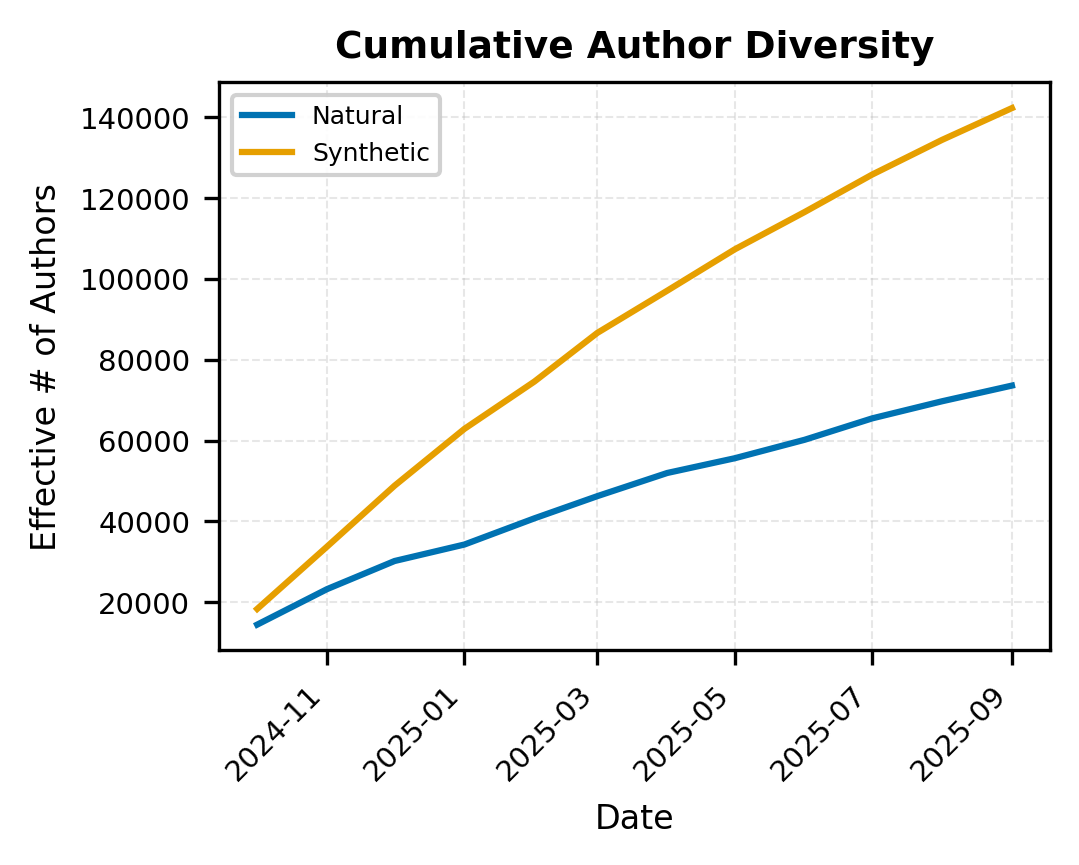}
        \caption{Authors surfaced by the synthetic rollouts are more diverse than the corresponding natural authors.}
        \label{fig:multiturn_figures_author_diversity_monthly_matched}
    \end{subfigure}
    \hfill
    \begin{subfigure}[t]{0.48\linewidth}
        \centering
        \includegraphics[width=\linewidth]{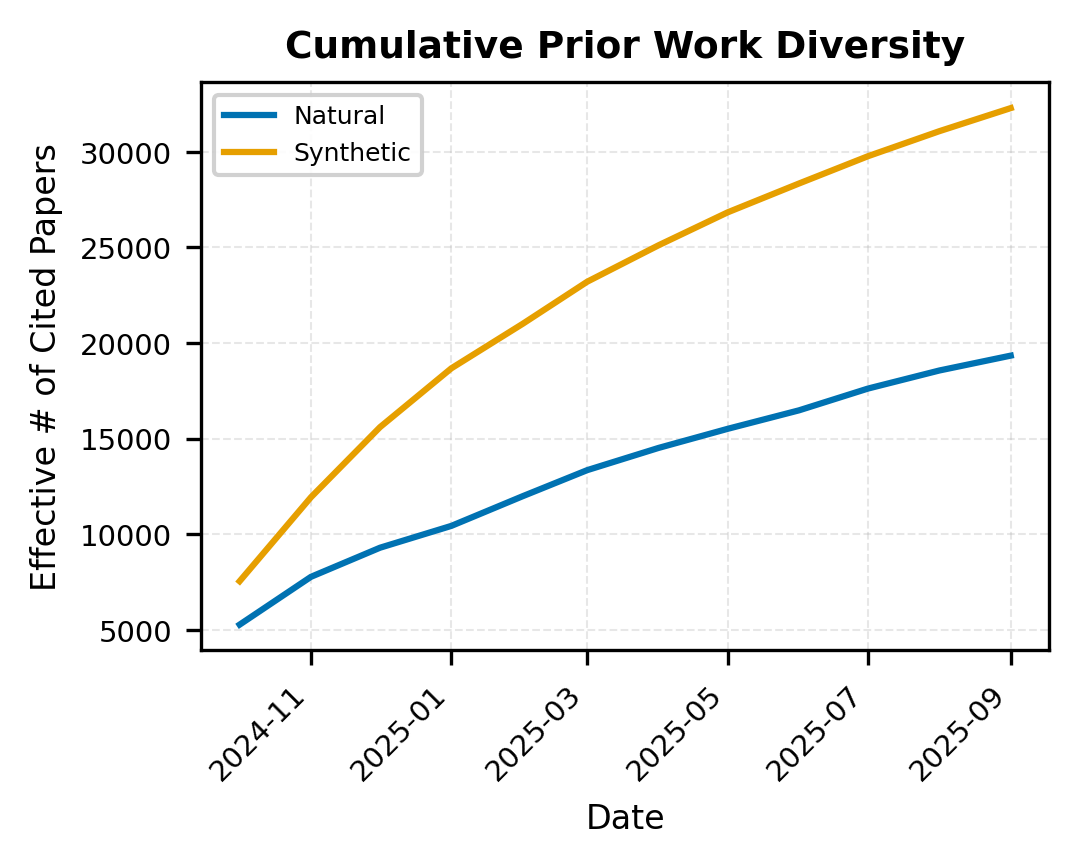}
        \caption{Prior work surfaced during synthetic rollouts is more diverse than by natural papers from the same time period.}
        \label{fig:multiturn_figures_key_references_diversity_monthly_matched}
    \end{subfigure}

    \caption{Diversity of Authors and Prior work surfaced during corpus generation. We employ retrieval pool subsampling to remove systemic biases due to discrepancies between natural and synthetic corpus sizes.}
    \label{fig:multiturn_figures_author_and_key_reference_diversity_monthly_matched}
\end{figure*}

\subsubsection{Discussion}
\begin{figure}[!htb]
    \centering
    \includegraphics[width=\linewidth]{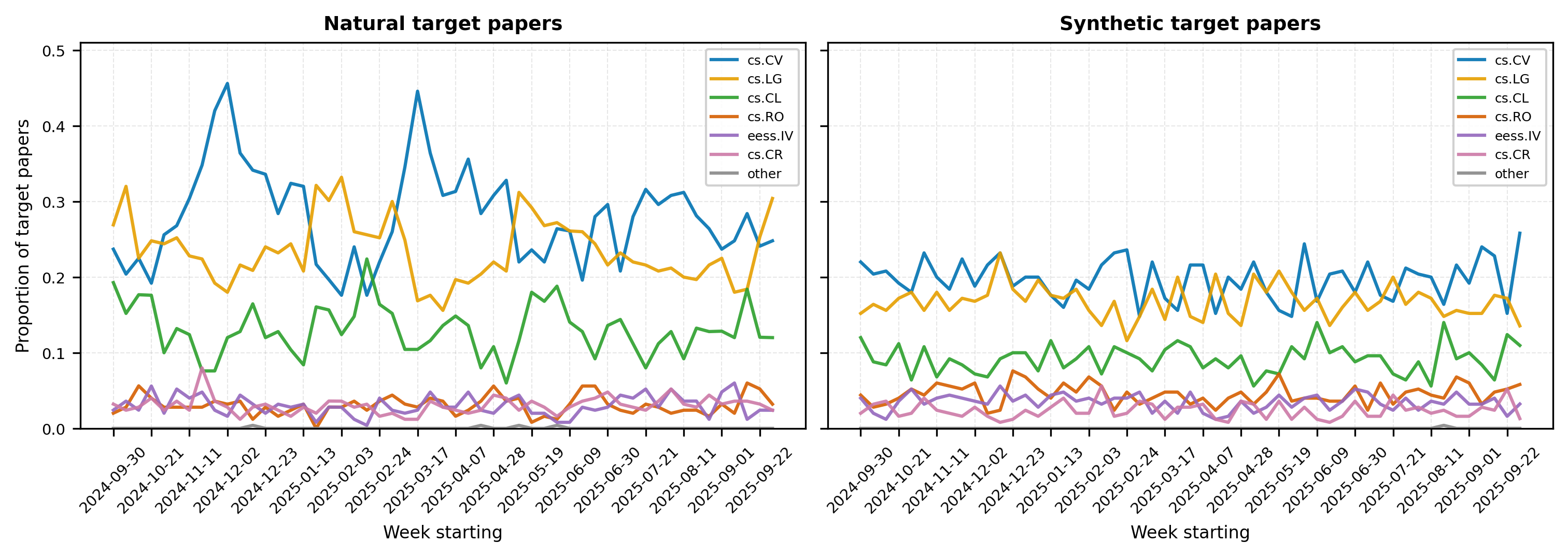}
    \caption{Primary arXiv topics of real-world (Natural) and Synthetic papers. Real-world papers show significant seasonal variation while synthetic ones do not.}
    \label{fig:topics}
\end{figure}

\paragraph{Realistically simulating corpus rollouts can be difficult.} Even assuming access to models that can perform individual tasks well, it can be challenging to use them to generate realistic corpora. Choices made while designing the procedure that utilizes these models for multi-turn corpus roll-outs can have unintended consequences that bias corpus statistics. For example, Figure~\ref{fig:topics} shows the distribution of primary arXiv topics of Natural and Synthetic\footnote{We use a classifier with $\sim70\%$ accuracy on a held-out set from the train period to predict the topics of synthesized papers.} papers corresponding to this period. Real-world research shows much more seasonal variation in the distribution over papers published year-round than the synthetic corpus. This arises from the fact that our simulation uniformly and independently randomly selects seed authors from the \benchmark~dataset for each synthesized paper whereas certain seed authors may be more likely to publish their work at certain times of the year than others due to external circumstances like venue deadlines or academic schedules. 

\paragraph{Individual statistics computed over a generated corpus can be misleading.}

\begin{figure}[!htb]
    \centering
    \includegraphics[width=0.5\linewidth]{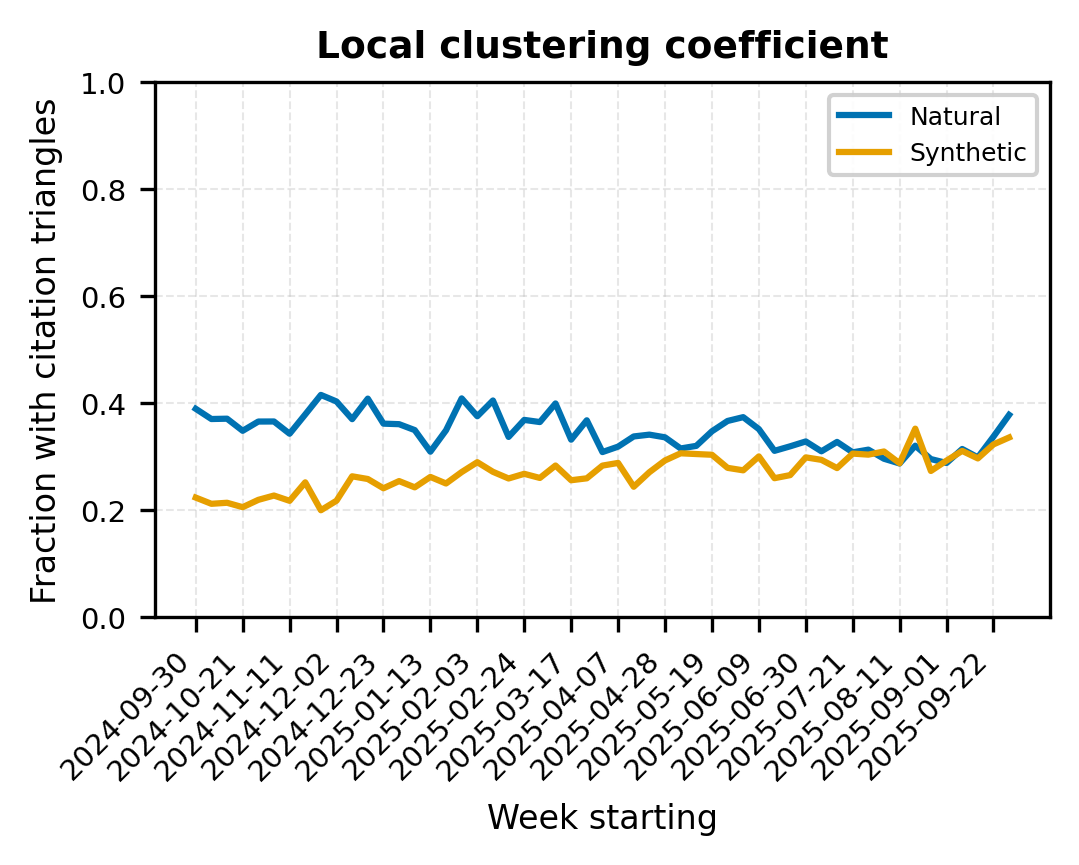}
    \caption{Local clustering coefficient (i.e. the fraction of pairs of influential references of the target paper that cite another of its influential references.}
    \label{fig:lcc}
\end{figure}

It can be difficult to accurately measure the quality of a synthetic corpus. For instance, in Figure~\ref{fig:lcc}, we measure the fraction of target papers from the natural and simulated corpora that contain at least one influential reference that cites another of the target paper's influential references. Although it appears that this coefficient approaches that from the natural corpus as the simulation proceeds, an observation that can be mistaken for implying that the simulated papers citation patterns become more natural as simulation proceeds--we find that its upward slope is due to another factor: Synthetic papers that enter the corpus can connect two previously disparate papers, and a future synthetic paper that cites the same set would now raise the local clustering coefficient as the simulation proceeds. A phenomenon that skews the synthetic corpus away from the natural corpus thus has the effect of causing this statistic to trend in the ``correct" direction.

\newpage \newpage

\section{Prompts}
\subsection{Topic Assignment Prompt} \label{app:prompt_topic_assignment}
\begin{tcolorbox}[
  title={Topic Assignment Prompt},
  breakable,
  colback=white,
  colframe=black,
  boxrule=0.4pt,
  arc=0pt,
  left=6pt,right=6pt,top=6pt,bottom=6pt
]
\begin{Verbatim}[fontsize=\footnotesize, breaklines=true, breakanywhere=true]

You are a scientific paper topic classifier. Given a paper title and abstract, assign all topics from the list below that clearly apply.

Topics:
- Instruction tuning
- Preference optimization / alignment
- RLHF / RLAIF for post-training
- Prompt engineering / prompt optimization
- Prompt tuning / soft prompting
- Parameter-efficient fine-tuning
- In-context learning
- Long-context modeling
- Personalized language modeling
- Retrieval-augmented generation
- Tool / API-augmented language modeling
- Knowledge editing / model updating
- Model compression / distillation for LMs
- Chain-of-thought prompting
- Tree/search-based reasoning
- Self-critique / self-refinement
- Language-model planning
- Web/navigation agents
- Code-generation agents
- Multi-agent LLM collaboration
- Memory mechanisms for LLM agents
- Multimodal language agents
- Agent evaluation and benchmarks
- Hallucination detection and mitigation
- Calibration / uncertainty estimation for LLMs
- Jailbreak and prompt-injection robustness
- Privacy leakage / machine unlearning for LMs
- Mechanistic interpretability of transformers and LLMs
- Dialogue modeling
- Dialogue state tracking
- Dialogue evaluation
- Multimodal dialogue
- User modeling for conversational AI
- Proactive conversational AI
- Contextual dialogue modeling
- Information extraction
- Relation extraction
- Event extraction
- Question answering
- Summarization
- Long-document understanding
- Argument mining
- Stance detection
- Sentiment analysis
- Syntactic parsing / tagging / chunking
- Morphology and word segmentation
- Sentence-level semantics and textual inference
- Figurative language understanding / generation
- Machine translation
- Low-resource NLP
- Multilingual representation learning
- Cross-cultural NLP
- Multimodality and language grounding
- Knowledge-augmented NLP
- Scientific NLP
- Scholarly document processing
- Citation / evidence extraction
- Climate NLP
- Language-and-molecules modeling
- Dense retrieval
- Neural ranking / learning to rank
- Generative retrieval
- Conversational search
- RAG system design and evaluation
- Knowledge-enhanced retrieval
- Long-context retrieval
- LLM-based IR evaluation
- Collaborative filtering
- Knowledge-based recommendation
- Deep learning for recommender systems
- Natural-language / conversational recommenders
- Fair / privacy-aware recommendation
- 3D reconstruction from multi-view and sensors
- 3D reconstruction from single images
- Adversarial attack and defense in vision
- Biometrics
- Computational imaging
- Datasets and evaluation for vision
- Efficient and scalable vision models
- Face analysis
- Body / pose / gesture / motion understanding
- Low-level vision
- Document analysis and understanding
- Open-set / open-world recognition
- Out-of-distribution detection in vision
- Fine-grained visual categorization
- Self-supervised visual representation learning
- Semi-supervised visual learning
- Few-shot visual recognition
- Domain adaptation in vision
- Diffusion models for image synthesis
- Autoregressive image generation
- Generative-model inversion
- Image editing with generative models
- Synthetic data for visual recognition
- Analysis-by-synthesis / render-and-compare recognition
- 3D foundation models
- 3D content creation
- Neural radiance fields
- 3D Gaussian splatting
- Implicit neural representations
- Novel-view synthesis
- Long-form video understanding
- Human action understanding / generation
- Large multimodal model evaluation
- Visual-prompt understanding in multimodal models
- Open-vocabulary / open-task vision-language models
- Vision-language reasoning
- Medical vision foundation models
- Medical image / video generation
- Biomedical image parsing
- Pathology image models
- Radiology foundation-model applications
- Medical imaging data curation
- Medical imaging data augmentation
- Fairness and safety in medical imaging AI
- Automatic speech recognition
- ASR error correction / rescoring
- Multilingual / multi-accent ASR
- Spoken language translation
- Spoken language understanding
- Text-to-speech
- Voice conversion
- Speaker verification
- Speaker diarization
- Speech processing with discrete units
- Speech emotion recognition / paralinguistics
- Pathological / health-related speech analysis
- Multi-channel speech enhancement
- Human-machine spoken interaction
- Responsible speech foundation models
- Speech deepfake / spoofing detection
- Streaming ASR
- Self-supervised learning for ASR
- Target-speaker extraction
- Deep reinforcement learning
- Hierarchical reinforcement learning
- Model-based reinforcement learning
- Policy optimization
- Bandits
- Sequential decision-making under uncertainty
- Active learning
- Adversarial learning
- Causal inference
- Causal discovery
- Causal representation learning
- Uncertainty quantification
- Probabilistic programming
- Approximate / variational inference
- Performative prediction
- Bayesian methods
- Belief propagation
- Robot foundation models
- Representation learning for robotic perception and control
- Imitation learning for robotics
- Reinforcement learning for physical robots
- Learning-and-planning hybrids in robotics
- Uncertainty-aware robotics
- Automatic robotic data generation
- Multimodal robot perception and sensor fusion
- Human-robot interaction with language / gestures
- Learning for robot task and motion planning
- Learning for hardware design and optimization
- Robot safety for learning-based systems
- Robot manipulation
- Visual navigation
- Locomotion learning
- Autonomous driving perception / prediction / planning
- Embodied question answering / embodied vision-language
- Open-world embodied AI
- Generative AI for embodied AI
- Dataset composition and curation for foundation models
- Data filtering / relabeling / augmentation
- Dataset quality / diversity / provenance analysis
- Data debugging / influence analysis
- Dataset distillation
- Synthetic-data effects / model collapse
- Weak-to-strong generalization
- Graph neural networks
- Relational / structured learning
- Spatio-temporal learning
- Time-series modeling
- Time-series foundation models
- Federated learning
- Privacy and security in data-centric ML
- Deep learning theory (training dynamics, generalization, optimization convergence)
- Scientific machine learning (PDE solvers, neural operators, PINNs)
- Continual learning and catastrophic forgetting
- Object detection, segmentation, and tracking in vision
- Video segmentation, tracking, and generation
- Remote sensing and geospatial vision
- Medical image segmentation and reconstruction (beyond foundation models)
- Biomedical signal analysis (EEG, ECG, physiological)
- Quantum machine learning
- Spiking and neuromorphic computing
- Explainable AI (non-mechanistic / non-LLM)
- AI governance, policy, and societal impact
- Systems and serving infrastructure for large models
- Audio and music modeling (non-speech)
- Tabular machine learning and AutoML
- Point cloud and 3D geometric learning
- Wireless communications and signal processing

Rules:
1) Output ONLY compact JSON of the form {"topics": ["<topic>", "<topic>", ...]}.
2) Each string in the list MUST exactly match one topic from the list above (same spelling and punctuation).
3) If no topic clearly applies, output {"topics": []}.
4) Never invent topics outside the list.
5) List all applicable topics in descending order of relevance. Papers commonly span multiple topics — a method, a task, a domain, and a data concern can each map to different topics. Include every topic the paper directly addresses, not just the primary focus. Exclude topics that are only mentioned in passing or tangentially adjacent.

Title: {title}
Abstract: {abstract}

List all applicable topics in descending order of relevance.

\end{Verbatim}
\end{tcolorbox}

\subsection{Contribution Generation Prompt} \label{app:prompt_contribution_generation}
\begin{tcolorbox}[
  title={Contribution Generation Prompt},
  breakable,
  colback=white,
  colframe=black,
  boxrule=0.4pt,
  arc=0pt,
  left=6pt,right=6pt,top=6pt,bottom=6pt
]
\begin{Verbatim}[fontsize=\footnotesize, breaklines=true, breakanywhere=true]
You are a seasoned computer science researcher who has done extensive work in machine learning, deep learning, computer vision, natural language processing, reinforcement learning, artificial intelligence, human computer interaction, and many related fields. 
You have spent many years on the organizing and peer-review committees of many relevant conferences and publications like NeurIPS, ICLR, ICML, ICCV, ACL, EMNLP, NAACL, AAAI, CHI, TMLR, TACL, etc.

You need to use your expertise to accurately and realistically predict a followup paper that builds on (cites) the set of background papers given to you. For the paper you predict, you must output its title and abstract.

Below are a few solved examples for this prediction problem where we provide only one possible followup.

<example 1>
Background Paper 1:
Title: Adam: A Method for Stochastic Optimization
Abstract: We introduce Adam, an algorithm for first-order gradient-based optimization of stochastic objective functions. The method is straightforward to implement and is based on adaptive estimates of lower-order moments of the gradients. The method is computationally efficient, has little memory requirements and is well suited for problems that are large in terms of data and/or parameters. The method is also appropriate for non-stationary objectives and problems with very noisy and/or sparse gradients. The method exhibits invariance to diagonal rescaling of the gradients by adapting to the geometry of the objective function. The hyper-parameters have intuitive interpretations and typically require little tuning. Some connections to related algorithms, on which Adam was inspired, are discussed. We also analyze the theoretical convergence properties of the algorithm and provide a regret bound on the convergence rate that is comparable to the best known results under the online convex optimization framework. We demonstrate that Adam works well in practice and compares favorably to other stochastic optimization methods.

Background Paper 2:
Title: IP-Adapter: Text Compatible Image Prompt Adapter for Text-to-Image Diffusion Models
Abstract: Recent years have witnessed the strong power of large text-to-image diffusion models for the impressive generative capability to create high-fidelity images. However, it is very tricky to generate desired images using only text prompt as it often involves complex prompt engineering. An alternative to text prompt is image prompt, as the saying goes:"an image is worth a thousand words". Although existing methods of direct fine-tuning from pretrained models are effective, they require large computing resources and are not compatible with other base models, text prompt, and structural controls. In this paper, we present IP-Adapter, an effective and lightweight adapter to achieve image prompt capability for the pretrained text-to-image diffusion models. The key design of our IP-Adapter is decoupled cross-attention mechanism that separates cross-attention layers for text features and image features. Despite the simplicity of our method, an IP-Adapter with only 22M parameters can achieve comparable or even better performance to a fully fine-tuned image prompt model. As we freeze the pretrained diffusion model, the proposed IP-Adapter can be generalized not only to other custom models fine-tuned from the same base model, but also to controllable generation using existing controllable tools. With the benefit of the decoupled cross-attention strategy, the image prompt can also work well with the text prompt to achieve multimodal image generation. The project page is available at \url{https://ip-adapter.github.io}.

Background Paper 3:
Title: High-Resolution Image Synthesis with Latent Diffusion Models
Abstract: By decomposing the image formation process into a sequential application of denoising autoencoders, diffusion models (DMs) achieve state-of-the-art synthesis results on image data and beyond. Additionally, their formulation allows for a guiding mechanism to control the image generation process without retraining. However, since these models typically operate directly in pixel space, optimization of powerful DMs often consumes hundreds of GPU days and inference is expensive due to sequential evaluations. To enable DM training on limited computational resources while retaining their quality and flexibility, we apply them in the latent space of powerful pretrained autoencoders. In contrast to previous work, training diffusion models on such a representation allows for the first time to reach a near-optimal point between complexity reduction and detail preservation, greatly boosting visual fidelity. By introducing cross-attention layers into the model architecture, we turn diffusion models into powerful and flexible generators for general conditioning inputs such as text or bounding boxes and high-resolution synthesis becomes possible in a convolutional manner. Our latent diffusion models (LDMs) achieve a new state of the art for image inpainting and highly competitive performance on various tasks, including unconditional image generation, semantic scene synthesis, and super-resolution, while significantly reducing computational requirements compared to pixel-based DMs. Code is available at https://github.com/CompVis/latent-diffusion.

Background Paper 4:
Title: BrainVis: Exploring the Bridge between Brain and Visual Signals via Image Reconstruction
Abstract: Analyzing and reconstructing visual stimuli from brain signals effectively advances the understanding of human visual system. However, the EEG signals are complex and contain significant noise. This leads to substantial limitations in existing works of visual stimuli reconstruction from EEG, such as difficulties in aligning EEG embeddings with the fine-grained semantic information and a heavy reliance on additional large self-collected dataset for training. To address these challenges, we propose a novel approach called BrainVis. Firstly, we divide the EEG signals into various units and apply a self-supervised approach on them to obtain EEG time-domain features, in an attempt to ease the training difficulty. Additionally, we also propose to utilize the frequency-domain features to enhance the EEG representations. Then, we simultaneously align EEG time-frequency embeddings with the interpolation of the coarse and fine-grained semantics in the CLIP space, to highlight the primary visual components and reduce the cross-modal alignment difficulty. Finally, we adopt the cascaded diffusion models to reconstruct images. Using only 10\% training data of the previous work, our proposed BrainVis outperforms state of the arts in both semantic fidelity reconstruction and generation quality. The code is available at https://github.com/RomGai/BrainVis.

Predicted Followup Paper:
Title: BrainDecoder: Style-Based Visual Decoding of EEG Signals
Abstract: Decoding neural representations of visual stimuli from electroencephalography (EEG) offers valuable insights into brain activity and cognition. Recent advancements in deep learning have significantly enhanced the field of visual decoding of EEG, primarily focusing on reconstructing the semantic content of visual stimuli. In this paper, we present a novel visual decoding pipeline that, in addition to recovering the content, emphasizes the reconstruction of the style, such as color and texture, of images viewed by the subject. Unlike previous methods, this ``style-based'' approach learns in the CLIP spaces of image and text separately, facilitating a more nuanced extraction of information from EEG signals. We also use captions for text alignment simpler than previously employed, which we find work better. Both quantitative and qualitative evaluations show that our method better preserves the style of visual stimuli and extracts more fine-grained semantic information from neural signals. Notably, it achieves significant improvements in quantitative results and sets a new state-of-the-art on the popular Brain2Image dataset.
</example 1>



<example 2>
Background Paper 1:
Title: CrypTen: Secure Multi-Party Computation Meets Machine Learning
Abstract: Secure multi-party computation (MPC) allows parties to perform computations on data while keeping that data private. This capability has great potential for machine-learning applications: it facilitates training of machine-learning models on private data sets owned by different parties, evaluation of one party's private model using another party's private data, etc. Although a range of studies implement machine-learning models via secure MPC, such implementations are not yet mainstream. Adoption of secure MPC is hampered by the absence of flexible software frameworks that"speak the language"of machine-learning researchers and engineers. To foster adoption of secure MPC in machine learning, we present CrypTen: a software framework that exposes popular secure MPC primitives via abstractions that are common in modern machine-learning frameworks, such as tensor computations, automatic differentiation, and modular neural networks. This paper describes the design of CrypTen and measure its performance on state-of-the-art models for text classification, speech recognition, and image classification. Our benchmarks show that CrypTen's GPU support and high-performance communication between (an arbitrary number of) parties allows it to perform efficient private evaluation of modern machine-learning models under a semi-honest threat model. For example, two parties using CrypTen can securely predict phonemes in speech recordings using Wav2Letter faster than real-time. We hope that CrypTen will spur adoption of secure MPC in the machine-learning community.

Predicted Followup Paper:
Title: Low-Latency Privacy-Preserving Deep Learning Design via Secure MPC
Abstract: Secure multi-party computation (MPC) facilitates privacy-preserving computation between multiple parties without leaking private information. While most secure deep learning techniques utilize MPC operations to achieve feasible privacy-preserving machine learning on downstream tasks, the overhead of the computation and communication still hampers their practical application. This work proposes a low-latency secret-sharing-based MPC design that reduces unnecessary communication rounds during the execution of MPC protocols. We also present a method for improving the computation of commonly used nonlinear functions in deep learning by integrating multivariate multiplication and coalescing different packets into one to maximize network utilization. Our experimental results indicate that our method is effective in a variety of settings, with a speedup in communication latency of $10\sim20\%$.
</example 2>



<example 3>
Background Paper 1:
Title: Retrieval-Augmented Generation for Large Language Models: A Survey
Abstract: Large Language Models (LLMs) demonstrate significant capabilities but face challenges such as hallucination, outdated knowledge, and non-transparent, untraceable reasoning processes. Augmented Generation (RAG) has emerged as a promising solution to these issues by incorporating real-time data from external databases into LLM responses. This enhances the accuracy and credibility of the models, particularly for knowledge-intensive tasks, and allows for continuous knowledge updates and integration of domain-specific information. RAG synergistically merges LLMs' intrinsic knowledge with the vast, dynamic repositories of external databases. This survey paper provides an in-depth analysis of the evolution of RAG, focusing on three key paradigms: Naive RAG, Advanced RAG, and Modular RAG. It methodically examines the three fundamental components of RAG systems: the retriever, the generator, and the augmentation methods, underscoring the cutting-edge technologies within each componenet. Additionally, the paper introduces novel metrics and capabilities for evaluating RAG models, as well as the most recent evaluation framework. Finally, the paper outlines future research directions from three perspectives: future challenges,modality extension,and the development of the RAG technical stack and ecosystem.

Background Paper 2:
Title: From Local to Global: A Graph RAG Approach to Query-Focused Summarization
Abstract: The use of retrieval-augmented generation (RAG) to retrieve relevant information from an external knowledge source enables large language models (LLMs) to answer questions over private and/or previously unseen document collections. However, RAG fails on global questions directed at an entire text corpus, such as"What are the main themes in the dataset?", since this is inherently a query-focused summarization (QFS) task, rather than an explicit retrieval task. Prior QFS methods, meanwhile, fail to scale to the quantities of text indexed by typical RAG systems. To combine the strengths of these contrasting methods, we propose a Graph RAG approach to question answering over private text corpora that scales with both the generality of user questions and the quantity of source text to be indexed. Our approach uses an LLM to build a graph-based text index in two stages: first to derive an entity knowledge graph from the source documents, then to pregenerate community summaries for all groups of closely-related entities. Given a question, each community summary is used to generate a partial response, before all partial responses are again summarized in a final response to the user. For a class of global sensemaking questions over datasets in the 1 million token range, we show that Graph RAG leads to substantial improvements over a naive RAG baseline for both the comprehensiveness and diversity of generated answers. An open-source, Python-based implementation of both global and local Graph RAG approaches is forthcoming at https://aka.ms/graphrag.

Predicted Followup Paper:
Title: LightRAG: Simple and Fast Retrieval-Augmented Generation
Abstract: Retrieval-Augmented Generation (RAG) systems enhance large language models (LLMs) by integrating external knowledge sources, enabling more accurate and contextually relevant responses tailored to user needs. However, existing RAG systems have significant limitations, including reliance on flat data representations and inadequate contextual awareness, which can lead to fragmented answers that fail to capture complex inter-dependencies. To address these challenges, we propose LightRAG, which incorporates graph structures into text indexing and retrieval processes. This innovative framework employs a dual-level retrieval system that enhances comprehensive information retrieval from both low-level and high-level knowledge discovery. Additionally, the integration of graph structures with vector representations facilitates efficient retrieval of related entities and their relationships, significantly improving response times while maintaining contextual relevance. This capability is further enhanced by an incremental update algorithm that ensures the timely integration of new data, allowing the system to remain effective and responsive in rapidly changing data environments. Extensive experimental validation demonstrates considerable improvements in retrieval accuracy and efficiency compared to existing approaches. We have made our LightRAG open-source and available at the link: https://github.com/HKUDS/LightRAG.
</example 3>



You need to first think through how exactly you want to combine the background papers (i.e. which aspects from these papers will be used in the followup work) before making each prediction. This will constitute the 'reasoning' part of your response. Only then will you make your prediction of the title and abstract of the followup work.

When making your prediction, please use the output format shown below. Please don't use any newlines or whitespace that cause deviation from this format.

Reasoning: ...
Title: ...
Abstract: ...

EXTREMELY IMPORTANT: Please make sure to output all 3 fields: Reasoning, Title and Abstract in that order before ending your response. RESPONSES WITHOUT THE Title AND Abstract FIELDS WILL BE CONSIDERED INVALID. YOU MUST OUTPUT ALL THREE FIELDS IN THE SAME RESPONSE SEPERATED BY NEWLINES. DO NOT SPLIT UP FIELDS BETWEEN RESPONSES.
\end{Verbatim}
\end{tcolorbox}

\subsection{Paraphrase Generation Prompt} \label{app:prompt_paraphrase}
\begin{tcolorbox}[
  title={Paraphrase Generation Prompt},
  breakable,
  colback=white,
  colframe=black,
  boxrule=0.4pt,
  arc=0pt,
  left=6pt,right=6pt,top=6pt,bottom=6pt
]
\begin{Verbatim}[fontsize=\footnotesize, breaklines=true, breakanywhere=true]
You are a seasoned computer science researcher who has done extensive work in machine learning, deep learning, computer vision, natural language processing, reinforcement learning, artificial intelligence, human computer interaction, and many related fields.
You have spent many years on the organizing and peer-review committees of many relevant conferences and publications like NeurIPS, ICLR, ICML, ICCV, ACL, EMNLP, NAACL, AAAI, CHI, TMLR, TACL, etc.

You need to use your expertise to paraphrase the title and abstract of an existing research paper that will be provided to you. Make sure you paraphrase them well. I am not looking for a one-one sentence-level paraphrasal, but another title and abstract that another author could have written about the same ideas and findings. Feel free to change the narrative structure of the abstract as you please.

When making your predictions, please use the following output format.

Title: ...
Abstract: ...

Do not generate any other text or explanations.
\end{Verbatim}
\end{tcolorbox}

\subsection{\followupmetric~Interpolations Prompt}
\label{app:prompt_lacerscore_interpolations}
\begin{tcolorbox}[
  title={LACER Interpolations Prompt},
  breakable,
  colback=white,
  colframe=black,
  boxrule=0.4pt,
  arc=0pt,
  left=6pt,right=6pt,top=6pt,bottom=6pt
]
\begin{Verbatim}[fontsize=\footnotesize, breaklines=true, breakanywhere=true]

You are an expert computer scientist.

You will be given a pair of title-abstracts (A and J) corresponding to real research papers. Your task is to synthesize a realistic sequence of 8 intermediate title-abstracts (B, C, D, E, F, G, H, I) that sequentially interpolate smoothly between the two provided ones that each change one specific facet of the corresponding contribution. Some examples of facets you could change are 1. the type of contribution (eg. analysis, benchmark, system demo, theoretical advance, etc.) 2. Type of architecture/technique used (eg. CNNs, transformers, RL, RAG, diffusion models, RNNs, etc. 3. The type of task / motivation (eg. image recognition, semantic parsing, language modeling, code generation, image generation, etc.)

Feel free to change other facets too as you see appropriate. It is important that the synthesized title-abstracts B-I represent a smooth interpolation between A and J such that no two consecutive title-abstracts differ too dramatically from each other. However, it is also important that every pair of consecutive title-abstracts from A-J do differ from each other in at least one important way.

Please return you responses in the following format:

Title B: ...
Abstract B: ...

Title C: ...
Abstract C: ...

Title D: ...
Abstract D: ...

Title E: ...
Abstract E: ...

Title F: ...
Abstract F: ...

Title G: ...
Abstract G: ...

Title H: ...
Abstract H: ...

Title I: ...
Abstract I: ...

Here are your inputs:

Title A: {{title_a}}
Abstract A: {{abstract_a}}

Title J: {{title_j}}
Abstract J: {{abstract_j}}

\end{Verbatim}
\end{tcolorbox}

\subsection{\followupmetric~Judge Prompt}
\label{app:prompt_lacerscore_judge}
\begin{tcolorbox}[
  title={\followupmetric~Judge Prompt},
  breakable,
  colback=white,
  colframe=black,
  boxrule=0.4pt,
  arc=0pt,
  left=6pt,right=6pt,top=6pt,bottom=6pt
]
\begin{Verbatim}[fontsize=\footnotesize, breaklines=true, breakanywhere=true]

You will be given a pair of title-abstracts Reference and Generated. Your task is to assign a score from 1-10 measuring how similar the Generated is to the Reference title-abstract where 1 represents the lowest similarity score and 10 represents the highest.

Here are some examples of References along with 10 Generations each corresponding to scores 1-10.

Reference A:
Intelligent Urban Surveillance in India: Real-Time Recognition of Individual Attributes
This study introduces a smart surveillance system for Indian cities that performs real-time identification and analysis of people's attributes. Leveraging advanced artificial intelligence and machine learning, the system recognizes features such as upper-body color, clothing, accessories, headgear, and so on, and analyzes behavior using camera feeds installed throughout the city.

Generations:
PyTorch: A High-Performance Deep Learning Library with an Imperative, Pythonic Design
Deep learning frameworks have often emphasized either ease of use or performance, but seldom both. PyTorch demonstrates that these aims can be jointly achieved: it provides an imperative, Python-native programming model that treats code as the model, simplifies debugging, and harmonizes with popular scientific computing libraries, while remaining efficient and supporting hardware accelerators such as GPUs. In this work, we articulate the principles that guided PyTorch's implementation and show how they manifest in its architecture. We underscore that every part of PyTorch is an ordinary Python program under the user's full control. We also describe how a careful, pragmatic realization of the runtime's key components enables them to interoperate to deliver compelling performance. Finally, we present evidence of the efficiency of individual subsystems and the overall speed of PyTorch on several common benchmarks.
Score: 1

PyTorch-Stream: Eager, High-Throughput Extensions for Real-Time Vision and Video
We present PyTorch-Stream, a set of runtime and library extensions that bring low-latency, multi-stream video analytics to PyTorch while preserving its imperative programming model. PyTorch-Stream introduces asynchronous camera ingest, zero-copy decoding, fused preprocessing ops, and CUDA graph capture to minimize per-frame overhead. We describe design decisions that retain Pythonic control flow yet enable ahead-of-time export of hot paths for steady-state performance. Across common video tasks (object detection, tracking-by-detection, and attribute classification), PyTorch-Stream reduces end-to-end latency by 30-55%
Score: 2

PyTorch-PersonAttr: A Reproducible Toolkit for Person Attribute Recognition in Surveillance Video
We release PyTorch-PersonAttr, a modular toolkit for training, evaluating, and deploying person attribute recognition models using the PyTorch-Stream runtime. The toolkit standardizes data adapters (e.g., RAP, PA-100k, PETA), implements strong CNN and hybrid CNN-Transformer baselines, and provides consistent evaluation protocols for multi-label classification (e.g., apparel type, color, accessories). We detail reference training recipes, mixed-precision/compiled execution for real-time inference, and uncertainty calibration methods for attribute outputs. Experiments show state-of-the-art or competitive results on public benchmarks and >60 FPS inference for 720p streams on a single GPU.
Score: 3

CityAttr: A Benchmark for Fine-Grained Person Attribute Recognition in Urban Camera Feeds
We introduce CityAttr, a benchmark targeting fine-grained person attributes in urban environments, with 250k annotated person crops spanning diverse weather, lighting, and camera viewpoints. CityAttr adds difficult classes pertinent to city operations (upper-body color under motion blur, reflective accessories, safety gear) and defines standardized train/val/test splits and occlusion strata. We provide strong PyTorch-PersonAttr baselines, error taxonomies, and fairness analyses across scene contexts. Results highlight failure modes in crowded scenes and under low light, motivating temporal modeling and domain adaptation for real-time deployments.
Score: 4

TorchStream City: A Low-Latency, Multi-Camera Video Analytics System for Urban Attribute Tagging
Building on PyTorch-Stream and CityAttr, we present TorchStream City, a reference system for city-scale, real-time attribute tagging. The system integrates multi-camera ingest, on-GPU batching/scheduling, online multi-object tracking, and temporal ensembling for stable attribute estimates. A hybrid eager-compiled execution path allows imperative control for orchestration while compiling per-operator graphs for steady-state throughput. Deployed on traffic and pedestrian cameras, TorchStream City sustains 25-30 FPS per stream across 32 feeds on two GPUs, with <200 ms median latency for detection, tracking, and attribute tagging, and provides APIs for downstream behavior analytics.
Score: 5

EdgeTorch-Attr: Efficient Person Attribute Recognition on Embedded and On-Camera Accelerators
We propose EdgeTorch-Attr, an optimization stack enabling person attribute recognition on edge devices. Techniques include compound scaling, quantization-aware training, structured pruning, knowledge distillation from server-grade models, and deployment via PyTorch 2 AOT compilation and TensorRT backends. We co-design the streaming pipeline to overlap decode, preproc, and inference on limited memory budgets. Field trials on NVIDIA Jetson- and VPU-class devices show 3-5x energy efficiency gains while maintaining >90%
Score: 6

IndAttr: Culturally-Aware Attribute Recognition for Indian Urban Scenes via Domain-Adaptive Temporal Transformers
We present IndAttr, a domain-adaptive model suite tailored to Indian urban video feeds. IndAttr augments attribute taxonomies to include culturally salient apparel and accessories (e.g., kurta, sari, dupatta, school uniforms, turbans, safety vests), and leverages temporal Vision Transformers with cross-frame attention for stability under occlusion and crowding. We introduce unsupervised domain adaptation from CityAttr using camera-style augmentation and self-training, plus limited expert-labeled Indian samples collected under ethical protocols. IndAttr improves mAP by 7.8 points over generic baselines on Indian pilot datasets while sustaining edge-ready throughput via distillation to compact temporal backbones.
Score: 7

PriCity-Attr: Privacy-Preserving, Federated Multi-Camera Attribute Analytics for Indian Smart Cities
We develop PriCity-Attr, a privacy-preserving training and deployment framework for multi-camera attribute analytics in Indian cities. The system performs on-device anonymization (face/body blurring for raw frames), stores only attribute and trajectory metadata, and trains models via cross-ward federated learning with secure aggregation and differential privacy accounting. We include governance hooks (policy-driven retention, audit logs) and human-in-the-loop tools for bias auditing. In a city pilot across 120 cameras, PriCity-Attr maintains IndAttr-level accuracy, reduces cross-site generalization error by 25%
Score: 8

Real-Time Indian Urban Monitoring: City-Scale Deployment of Attribute and Behavior Analytics
We present a production-grade, city-scale deployment that integrates IndAttr and PriCity-Attr into a unified smart monitoring platform for Indian metros. The system performs real-time identification of salient person attributes (upper-body color, apparel type, accessories, headgear) and stabilizes outputs via temporal modeling to feed behavior analytics (zone occupancy, queueing, safety-gear compliance). An edge-cloud architecture supports heterogeneous cameras, with resilient scheduling, adaptive bitrate, and failover. We report month-long operations over 500+ feeds, covering throughput, latency, availability, and model drift monitoring, demonstrating actionable, privacy-aware analytics for urban management at scale.
Score: 9

Intelligent Urban Surveillance in India: Real-Time Recognition of Individual Attributes
This study introduces a smart surveillance system for Indian cities that performs real-time identification and analysis of people's attributes. Leveraging advanced artificial intelligence and machine learning, the system recognizes features such as upper-body color, clothing, accessories, headgear, and so on, and analyzes behavior using camera feeds installed throughout the city.
Score: 10

Reference B:
Ada-Instruct: Retooling Instruction Generation for Complex Reasoning
Augmenting instructions is critical for unlocking the full potential of large language models (LLMs) on downstream tasks. Current Self-Instruct approaches largely fabricate new instructions from a small seed set via in-context learning. Our analysis reveals a key limitation of this paradigm: even with GPT4o, Self-Instruct fails to produce complex instructions of length 100 or more, which are necessary for demanding tasks such as code completion. To overcome this, we observe that fine-tuning open source LLMs with only ten examples can yield complex instructions that preserve distributional consistency for complex reasoning tasks. We introduce Ada-Instruct, an adaptive instruction generator obtained through fine-tuning. We validate Ada-Instruct empirically across different applications, and the results demonstrate its ability to generate long, intricate, and distributionally consistent instructions.

Generations:
Evaluating Mathematical Problem Solving Using the MATH Dataset
Many forms of intellectual work depend on mathematical problem solving, yet computers still struggle with this capability. To assess this skill in machine learning systems, we present MATH, a dataset of 12,500 difficult competition mathematics problems. Each problem includes a complete step-by-step solution, enabling models to be trained to produce answer derivations and explanations. To support future research and improve accuracy on MATH, we additionally release a large auxiliary pretraining dataset aimed at teaching models the fundamentals of mathematics. While we are able to raise accuracy on MATH, our results show that performance remains relatively low, even with extremely large Transformer models. Moreover, if current scaling trends persist, simply expanding compute budgets and model parameter counts appears impractical for achieving strong mathematical reasoning. Although scaling Transformers is largely solving most other text-based tasks, it is not currently solving MATH. Substantive progress on mathematical problem solving will likely require new algorithmic advances from the broader research community.
Score: 1

Fine-Grained Evaluation of Mathematical Reasoning with Stepwise Derivations
We revisit evaluation on competition mathematics by exploiting the full step-by-step solutions accompanying problems. Beyond final-answer accuracy, we introduce derivation-aware metrics that align a model's generated reasoning with reference solutions via step matching, dependency agreement, and algebraic edit distance. We release standardized evaluation scripts and human-verified annotations for a subset of the MATH dataset, enabling assessment of intermediate reasoning quality and error localization. Our analysis across model scales shows that improvements in final accuracy often mask brittle derivations and shortcut patterns. We find that models benefiting from chain-of-thought prompting still diverge significantly from canonical derivations, suggesting the need for training signals that target the structure and granularity of mathematical reasoning rather than only end answers.
Score: 2

AutoHint-MATH: Augmenting Competition Problems with Model-Generated Hints and Subgoals
To bridge the gap between final-answer evaluation and robust derivations, we introduce AutoHint-MATH, a data augmentation pipeline that generates targeted hints, subgoals, and intermediate checks for competition-level math problems. Using large language models guided by our stepwise metrics, we produce pedagogically structured scaffolds that highlight crucial transformations and decision points. We curate 60k high-quality hint-problem-solution triplets and show that training with these augmentations yields consistent gains in both final answers and derivation alignment on MATH. Compared to vanilla chain-of-thought prompting, AutoHint-MATH improves sample efficiency and reduces off-track reasoning, indicating that explicit scaffolding can better shape model search through complex solution spaces.
Score: 3

InstructMATH: Instruction-Tuning for Step-by-Step Mathematical Problem Solving
We move from passive scaffolds to active task framing by introducing InstructMATH, a collection of instruction-problem-solution triples designed to instruction-tune models for mathematical reasoning. Instructions specify goals, permissible operations, and verification strategies, teaching models how to structure derivations before generating them. We compile 25k diverse instruction templates spanning algebra, geometry, and number theory, and show that instruction-tuned models outperform hint-augmented baselines on MATH under both few-shot and zero-shot settings. Analyses indicate that explicit procedural instructions reduce hallucinated steps and improve adherence to algebraic invariants. Our results suggest instruction-tuning is a promising path to controllable, reliable mathematical reasoning.
Score: 4

ReasonBench: A Multi-Domain Corpus of Instruction-Driven Problems with Worked Solutions
To test whether instruction-tuned reasoning transfers beyond mathematics, we create ReasonBench, a multi-domain corpus covering competition math, physics word problems, logic puzzles, and algorithmic thinking tasks, each paired with domain-specific instructions and worked solutions. We design a shared schema for instructions (goal, constraints, permitted tools, validation) and provide cross-domain transfer splits. Instruction-tuned models trained on ReasonBench improve on out-of-domain reasoning tasks, especially when instructions emphasize decomposition and verification. However, performance plateaus on tasks requiring long, intricate directives (e.g., multi-function code synthesis, multi-lemma proofs), motivating a closer study of instruction complexity and its role in reasoning.
Score: 5

On the Complexity of Instructions for Complex Reasoning Tasks
We systematically measure how instruction complexity-length, structural depth, and constraint density-affects performance on challenging tasks such as proof sketching and code completion. We find that instructions longer than ~100 tokens with nested constraints substantially improve reliability in multi-step reasoning, but current self-instruction pipelines rarely produce such long, structured directives and often drift from the target task's distribution. Using ReasonBench extensions for coding and formal reasoning, we quantify this mismatch and show that in-context self-instruct deteriorates instruction quality as length increases. These results highlight a bottleneck: generating long, distributionally faithful instructions is essential for unlocking complex reasoning, yet existing automatic pipelines fall short.
Score: 6

Scaffold-Instruct: Hierarchical Self-Instruct for Long-Form, Structured Prompts
We propose Scaffold-Instruct, a hierarchical self-instruction framework that builds long-form instructions via iterative decomposition: a planner drafts high-level goals, a refiner injects constraints and verification steps, and a validator enforces structural templates. Scaffold-Instruct increases instruction length and structural depth while improving internal consistency over standard self-instruct. On math, logic, and code tasks, models trained with Scaffold-Instruct show better adherence to multi-step procedures and fewer dead ends. Nevertheless, the approach is compute-intensive, relies on strong planners, and still exhibits distributional drift on specialized domains like code completion. These limitations motivate moving from purely in-context pipelines toward trainable instruction generators.
Score: 7

Few-Shot Fine-Tuned Instruction Generators for Mathematical and Program Reasoning
We introduce a trainable alternative to hierarchical self-instruct: fine-tuning a compact open-source LLM to act as an instruction generator using a small curated set (20-50) of long, structured seed instructions per domain. The resulting generator produces instructions that are longer, more constraint-aware, and more distributionally aligned with math and coding datasets than in-context self-instruct. When used to create training corpora, these instructions yield larger gains on multi-step reasoning and code completion benchmarks than Scaffold-Instruct, at a fraction of the cost. Ablations show that even minimal fine-tuning stabilizes long-instruction formatting and reduces semantic drift, suggesting few-shot fine-tuned generators are a scalable path forward.
Score: 8

AdaGen: Adaptive Few-Shot Instruction Generators for Complex Reasoning
Building on few-shot fine-tuned generators, we present AdaGen, an adaptive instruction generation framework that (1) fine-tunes open-source LLMs with as few as ten carefully selected seed instructions per domain, and (2) adaptively controls length, constraint density, and style to match target-task distributions. AdaGen automatically diagnoses gaps in coverage using lightweight discriminators and iteratively steers generation to maintain distributional fidelity. Across mathematical reasoning and code completion, AdaGen produces long (>=100 tokens), structurally rich instructions that lead to consistent downstream gains over self-instruct and hierarchical baselines. These findings set the stage for general-purpose, low-budget instruction generators tailored to complex reasoning tasks.
Score: 9

Ada-Instruct: Retooling Instruction Generation for Complex Reasoning
Augmenting instructions is critical for unlocking the full potential of large language models (LLMs) on downstream tasks. Current Self-Instruct approaches largely fabricate new instructions from a small seed set via in-context learning. Our analysis reveals a key limitation of this paradigm: even with GPT4o, Self-Instruct fails to produce complex instructions of length 100 or more, which are necessary for demanding tasks such as code completion. To overcome this, we observe that fine-tuning open source LLMs with only ten examples can yield complex instructions that preserve distributional consistency for complex reasoning tasks. We introduce Ada-Instruct, an adaptive instruction generator obtained through fine-tuning. We validate Ada-Instruct empirically across different applications, and the results demonstrate its ability to generate long, intricate, and distributionally consistent instructions.
Score: 10

Reference C:
Reinforced Self-Training (ReST) for Training Language Models
Reinforcement learning from human feedback (RLHF) can enhance the quality of large language models' (LLMs) outputs by aligning them with human preferences. We present a straightforward alignment approach, inspired by growing-batch reinforcement learning (RL), which we call Reinforced Self-Training (ReST). Starting from an initial LLM policy, ReST constructs a dataset by sampling from that policy, and then uses offline RL algorithms on this data to further improve the policy. Because the training data are generated offline and can be reused, ReST is more efficient than standard online RLHF procedures. Although ReST is a general method applicable across generative learning scenarios, we center our investigation on machine translation. Our results demonstrate that ReST can substantially boost translation quality-according to both automated metrics and human evaluations on machine translation benchmarks-while being compute- and sample-efficient.

Generations:
Algorithms for Proximal Policy Optimization
We present a new class of policy gradient algorithms for reinforcement learning that alternate between collecting data through environment interaction and maximizing a surrogate objective via stochastic gradient ascent. Unlike standard policy gradient approaches that make only one gradient update per data sample, we introduce a novel objective that enables multiple epochs of minibatch updates. The resulting methods, termed proximal policy optimization (PPO), capture some of the benefits of trust region policy optimization (TRPO) while being much simpler to implement, more general, and empirically exhibiting better sample complexity. We evaluate PPO on a set of benchmark tasks, including simulated robotic locomotion and Atari game playing, and show that it outperforms other online policy gradient methods, achieving an overall favorable balance among sample complexity, simplicity, and wall-time.
Score: 1

Proximal Policy Optimization with Replay: Enabling Stable Data Reuse in On-Policy RL
We extend proximal policy optimization (PPO) with a principled form of experience replay that increases sample efficiency while preserving stability. Our method augments the clipped surrogate objective with lightweight off-policy corrections using per-batch importance weights and a target network for value learning, enabling multiple epochs over mixed recent on-policy and near-on-policy samples. Across MuJoCo locomotion and Atari benchmarks, PPO+Replay achieves comparable or better asymptotic performance than PPO with 30-50%
Score: 2

Proximal Sequence Policy Optimization for Text Generation
We adapt proximal policy optimization to autoregressive sequence models for text generation. Our approach optimizes a clipped sequence-level policy gradient objective over Transformer language models, with rewards defined by task-specific automated metrics (e.g., ROUGE for summarization, BLEU for translation). We introduce a token-wise advantage estimator that stabilizes long-horizon credit assignment and a curriculum over sequence lengths. Experiments on CNN/DailyMail summarization and WMT14 En-De translation show consistent metric gains over supervised fine-tuning and REINFORCE baselines at comparable compute, demonstrating that PPO's trust-region-like updates transfer effectively to discrete text generation.
Score: 3

KL-Regularized Proximal Optimization for Controlled Text Generation
We propose a KL-regularized variant of proximal optimization for sequence models that constrains updates relative to a supervised reference model. The objective augments PPO's clipped surrogate with an adaptive KL penalty against the reference policy, yielding stable improvements while preserving fluency. We instantiate rewards from automated metrics and task-specific constraints (length, toxicity filters), and introduce a reference-anchored value function to reduce variance. On summarization and translation, KL-Prox achieves higher ROUGE/BLEU and better human-rated coherence than metric-only PPO, highlighting the benefit of explicit distributional control for language generation.
Score: 4

Preference-Guided Proximal Optimization: Online RLHF for Sequence Models
We integrate human preference learning with proximal policy optimization for language generation. A reward model is trained from pairwise human comparisons and used to guide KL-regularized PPO updates of a Transformer policy anchored to a supervised reference. We interleave data collection (sampling candidate texts), preference labeling, reward model updates, and policy improvement. On summarization and open-ended generation, our method improves human preference win-rates over supervised baselines and metric-optimized PPO, while maintaining stability via adaptive KL and early stopping. Results demonstrate that online RLHF can be realized with simple proximal updates and modest human annotation.
Score: 5

Growing-Batch Preference Optimization: Bridging Online RLHF and Offline RL for LMs
We introduce a growing-batch training regime for preference-optimized language models that alternates between short bursts of sample collection and extended offline optimization. Candidate texts are generated from the current policy, scored by a learned preference model, and stored in a replay buffer. We then apply offline RL-style updates-advantage-weighted likelihood with conservative clipping and value targets fit by fitted policy evaluation-over the accumulated batch, with occasional refresh of the data. Compared to purely online PPO-based RLHF, Growing-Batch PO reduces the number of preference queries and environment interactions while achieving similar or better human preference win-rates on summarization and dialogue.
Score: 6

Self-Generated Corpora for Batch RLHF: Offline Advantage Optimization of LMs
We present a purely offline procedure for aligning language models using self-generated datasets. Starting from a supervised policy, we sample candidate continuations, score them with a learned preference model, and construct a static training set. We then perform multiple epochs of offline reinforcement learning using advantage-weighted regression with KL regularization to the reference, without further online sampling during optimization. This self-training loop can be repeated to grow the dataset. On summarization and translation, our batch RLHF approach matches or exceeds online PPO-based RLHF in human and automatic metrics while reducing compute and annotation overhead via data reuse.
Score: 7

Offline Reinforcement Learning for Machine Translation via Self-Generated Batches
Focusing on machine translation, we instantiate an offline RL framework that improves a Transformer translator using only self-generated candidate translations and offline optimization. We generate n-best lists from the current model, score them with a mixture of automatic metrics (BLEU, COMET) and a lightweight preference model, and train with KL-regularized advantage-weighted updates over the static pool. Periodic regeneration expands coverage, but all policy learning is performed offline, enabling aggressive data reuse. On WMT benchmarks, our method delivers substantial BLEU/COMET gains over supervised fine-tuning and online policy gradient methods at lower compute, with humans preferring our outputs for adequacy and fluency.
Score: 8

Preference-Scored Self-Training with Offline RL for LLM Translators
We propose a simple self-training algorithm for large language model (LLM) translators that combines dataset generation by the model with offline reinforcement learning. Given an initial LLM policy and a reference model, we produce candidate translations, score them using a learned preference model (optionally blended with automated metrics), and optimize the policy offline with KL-regularized advantage-weighted regression and value fitting. The process can be repeated in a growing-batch fashion to refresh the static dataset while retaining offline updates. Applied to WMT machine translation, this approach yields sizable improvements in BLEU/COMET and human evaluations relative to supervised fine-tuning and online RLHF, with favorable compute and sample efficiency.
Score: 9

Reinforced Self-Training (ReST) for Training Language Models
Reinforcement learning from human feedback (RLHF) can enhance the quality of large language models' (LLMs) outputs by aligning them with human preferences. We present a straightforward alignment approach, inspired by growing-batch reinforcement learning (RL), which we call Reinforced Self-Training (ReST). Starting from an initial LLM policy, ReST constructs a dataset by sampling from that policy, and then uses offline RL algorithms on this data to further improve the policy. Because the training data are generated offline and can be reused, ReST is more efficient than standard online RLHF procedures. Although ReST is a general method applicable across generative learning scenarios, we center our investigation on machine translation. Our results demonstrate that ReST can substantially boost translation quality-according to both automated metrics and human evaluations on machine translation benchmarks-while being compute- and sample-efficient.
Score: 10

Reference D:
MAPO: Boosting Multilingual Reasoning by Casting Cross-Language Alignment as Preference Optimization
Although reasoning is often regarded as language-independent, current LLMs display uneven reasoning performance across languages; for instance, a dominant language like English typically outperforms others due to imbalanced multilingual training data. To strengthen reasoning in non-dominant languages, we introduce a Multilingual-Alignment-as-Preference Optimization framework (MAPO) that aligns the reasoning processes in other languages with those of the dominant language. In particular, we leverage an off-the-shelf translation model to enforce consistency between answers in non-dominant and dominant languages, and treat this consistency as a preference signal for optimization with methods such as Direct Preference Optimization (DPO) or Proximal Policy Optimization (PPO). Empirically, MAPO reliably delivers substantial improvements in multilingual reasoning across various models on three benchmarks (MSVAMP +16.2%

Generations:
Advancing Cross-Lingual Mathematical Reasoning: Empirical Findings and Lessons Learned
Prior work has largely emphasized building strong language models (LLMs) for mathematical reasoning in single-language settings, with limited attention to maintaining performance across multiple languages. To close this gap, we initiate the exploration and training of robust Multilingual Math Reasoning (xMR) LLMs. Using translation, we assemble the first multilingual instruction dataset for math reasoning, MGSM8KInstruct, spanning ten languages, thereby mitigating training data scarcity for xMR tasks. Leveraging this corpus, we introduce several training strategies to construct a family of xMR LLMs, dubbed MathOctopus, which consistently outperform typical open-source LLMs and surpass ChatGPT in few-shot setups. In particular, MathOctopus-13B attains 47.6%
Score: 1

Consistency-Regularized SFT for Multilingual Mathematical Reasoning
Building on multilingual instruction tuning for mathematical reasoning, we investigate whether explicitly encouraging cross-lingual agreement can further improve xMR LLMs. Using the MGSM8KInstruct corpus, we construct parallel problem-solution-rationale tuples across ten languages via high-quality translation. We introduce a consistency-regularized supervised fine-tuning (CR-SFT) objective that augments standard SFT with two auxiliary signals: (1) answer-level agreement across parallel languages, enforced via a differentiable dual-decoding consistency loss; and (2) rationale-level agreement, measured by semantic similarity between chain-of-thought embeddings. Fine-tuning 7B-13B models with CR-SFT yields consistent gains over MathOctopus, with the 7B model improving by +2.4%
Score: 2

Cross-Lingual Distillation for Multilingual Math Reasoning
We extend consistency-regularized multilingual SFT with a teacher-student framework that distills reasoning skills from a strong English teacher into a multilingual student. Given parallel MGSM8KInstruct data, the teacher generates high-quality rationales and answers in English, which are translated and used to supervise the student in non-dominant languages. We combine teacher-forced rationale distillation with consistency losses on final answers, and introduce a selective distillation filter to discard low-confidence teacher traces. On MGSM and a bilingual subset of MSVAMP, our 7B student outperforms CR-SFT by +1.5-2.1%
Score: 3

Round-Trip Self-Training for Low-Resource Multilingual Reasoning
To reduce reliance on parallel supervision, we propose round-trip self-training (RT-ST) that bootstraps multilingual mathematical reasoning from monolingual non-English data. RT-ST generates pseudo-labeled rationales by translating non-English problems to English, invoking the English teacher to produce rationales/answers, and translating them back. We introduce round-trip validation-requiring that translating the generated rationale back to English recovers an equivalent solution-and an error-aware filter using a symbolic solver. Training with a mix of distilled parallel data and RT-ST pseudo-labels improves low-resource languages by +2.7%
Score: 4

Step-Contrast: Contrastive Alignment of Cross-Lingual Reasoning Traces
We study whether aligning the structure of reasoning across languages yields further gains beyond answer consistency. Step-Contrast introduces a contrastive objective over token-level reasoning steps: steps that correspond across translations are pulled together in an embedding space, while steps from unrelated problems are pushed apart. We obtain soft step correspondences using attention-based alignment between bilingual rationales. Step-Contrast is combined with SFT and RT-ST to yield a unified objective that encourages both accurate and structurally aligned reasoning. On MGSM and MSVAMP, Step-Contrast reduces cross-language variance of step-level entailment by 15%
Score: 5

XM-Reward: Cross-Lingual Consistency Reward Modeling for Reasoning
Moving beyond purely supervised objectives, we introduce XM-Reward, a learned reward model that scores multilingual reasoning traces by cross-lingual faithfulness and correctness. XM-Reward leverages (i) translation-consistency checks between non-dominant and English rationales/answers, (ii) symbolic verification when applicable, and (iii) fluency and step-validity heuristics. We train XM-Reward on preference pairs derived from Step-Contrast outputs and filtered by round-trip validation. Using offline policy optimization with reward-weighted SFT and KL-regularization, we improve over Step-Contrast by +1.6%
Score: 6

MPO: Multilingual Preference Optimization via Consistency-Derived Feedback
We convert cross-lingual consistency signals into pairwise preferences to directly optimize the policy with preference learning. Our Multilingual Preference Optimization (MPO) framework constructs, for each problem and language, pairs of model-generated traces with different cross-lingual agreement and solver-verified correctness. We study Direct Preference Optimization (DPO) vs. likelihood-ratio methods on this multilingual preference data, initialized from the XM-Reward-tuned model. MPO enhances both answer accuracy and cross-language agreement: on MGSM, it yields +3.8%
Score: 7

Anchor-Align PO: Dominant-Language Anchored Preference Optimization for xMR
We formalize dominant-language anchoring in preference optimization. Anchor-Align PO uses an off-the-shelf translation model to map non-dominant language traces to English, and defines preferences by (1) answer equivalence to a trusted English reference, (2) rationale consistency measured by bilingual entailment, and (3) solver-backed correctness. We train with both DPO and PPO variants, balancing cross-lingual alignment with exploration via KL control. Evaluated on MGSM, MSVAMP, and a subset of MNumGLUESub, Anchor-Align PO improves non-dominant language performance by +5.6%
Score: 8

Towards MAPO: Multilingual-Alignment-as-Preference Optimization for Reasoning
We unify consistency-driven preference learning into a general Multilingual-Alignment-as-Preference Optimization (MAPO) paradigm. MAPO defines preferences by aligning non-dominant language reasoning to a dominant-language anchor using translation-mediated agreement on both answers and intermediate steps, supervised by solver checks when feasible. The framework supports DPO for stable offline alignment and PPO for online refinement, and plugs into any base LLM. Across three benchmarks-MSVAMP, MGSM, and MNumGLUESub-MAPO yields robust gains over Anchor-Align PO (+1.5-3.0%
Score: 9

MAPO: Boosting Multilingual Reasoning by Casting Cross-Language Alignment as Preference Optimization
Although reasoning is often regarded as language-independent, current LLMs display uneven reasoning performance across languages; for instance, a dominant language like English typically outperforms others due to imbalanced multilingual training data. To strengthen reasoning in non-dominant languages, we introduce a Multilingual-Alignment-as-Preference Optimization framework (MAPO) that aligns the reasoning processes in other languages with those of the dominant language. In particular, we leverage an off-the-shelf translation model to enforce consistency between answers in non-dominant and dominant languages, and treat this consistency as a preference signal for optimization with methods such as Direct Preference Optimization (DPO) or Proximal Policy Optimization (PPO). Empirically, MAPO reliably delivers substantial improvements in multilingual reasoning across various models on three benchmarks (MSVAMP +16.2%
Score: 10

Reference E:
ConDaFormer: A Decomposed Transformer with Local Structure Boosting for 3D Point Cloud Analysis
Recent studies have applied transformers to 3D point cloud understanding with notable success. However, the sheer number of points-often over 0.1 million-renders global self-attention impractical for point clouds. Consequently, many methods confine the transformer to local neighborhoods, such as spherical or cubic windows. Even then, the volume of Query-Key pairs remains large, incurring high computational overhead. Moreover, prior approaches typically derive queries, keys, and values via linear projections, neglecting explicit modeling of local 3D geometric structure. To tackle both efficiency and local geometry priors, we propose a new transformer block, ConDaFormer. Specifically, ConDaFormer decomposes a cubic window into three orthogonal 2D planes, thereby reducing the number of points involved while maintaining a comparable attention span. This disassembly enables a larger attention range without increasing computational complexity, albeit at the risk of missing some contextual information. To address this, we introduce a local structure enhancement strategy that applies a depth-wise convolution before and after the attention, which also reinforces the encoding of local geometric cues. With these components, ConDaFormer captures both long-range contextual dependencies and local priors. We demonstrate its effectiveness through experiments on multiple 3D point cloud understanding benchmarks. Code is available at https://github.com/LHDuan/ConDaFormer .

Generations:
Swin: A Hierarchical Vision Transformer Based on Shifted Window Attention
We introduce Swin Transformer, a vision Transformer that serves as a general-purpose backbone for computer vision. Adapting Transformers from language to vision is challenging due to domain differences, including the wide range of object scales in images and the far higher spatial resolution of pixels compared to words. To bridge this gap, we propose a hierarchical Transformer that computes representations using shifted windows. This shifted window strategy confines self-attention to non-overlapping local windows for efficiency, while still enabling connections across windows. The hierarchical design supports multi-scale modeling and achieves computational complexity that scales linearly with image size. These properties make Swin Transformer applicable to a broad set of tasks, from image classification (87.3 top-1 accuracy on ImageNet-1K) to dense prediction tasks such as object detection (58.7 box AP and 51.1 mask AP on COCO test-dev) and semantic segmentation (53.5 mIoU on ADE20K val). It surpasses prior state-of-the-art results by large margins of +2.7 box AP and +2.6 mask AP on COCO, and +3.2 mIoU on ADE20K, highlighting the potential of Transformer-based backbones for vision. The hierarchical formulation and shifted window mechanism also prove advantageous for all-MLP architectures. Code and models are available at https://github.com/microsoft/Swin-Transformer.
Score: 1

Swin-Stripe: Shifted Window Transformers with Orthogonal Strip Fusion for 2D Vision
We extend the shifted-window paradigm of hierarchical vision Transformers by introducing orthogonal strip attention that complements square local windows. In Swin-Stripe, self-attention is computed within non-overlapping windows as in Swin, while lightweight horizontal and vertical strip attentions bridge distant regions with negligible overhead. This hybrid local-global scheme increases the effective receptive field without abandoning linear complexity in image size. The architecture remains hierarchical and maintains cyclic shifts for cross-window connectivity. Experiments on image classification, object detection, and semantic segmentation show consistent gains over a pure windowed baseline, particularly on long, thin structures where strip fusion is beneficial.
Score: 2

Video Swin-Tube: Hierarchical Spatiotemporal Transformer with Shifted Tube Windows
We adapt shifted-window Transformers from images to videos by generalizing local windows into spatiotemporal tubes. Video Swin-Tube performs self-attention within space-time tubes and uses cyclic temporal-spatial shifts to exchange information across tube boundaries. To preserve efficiency, we augment tube attention with the orthogonal strip fusion from Swin-Stripe along spatial and temporal axes, enlarging context without incurring full global cost. The hierarchical design yields linear complexity in the number of frames and spatial resolution. On video action recognition and spatiotemporal segmentation benchmarks, Video Swin-Tube achieves competitive accuracy with significantly reduced memory compared to global attention baselines.
Score: 3

Voxel-Swin: Shifted Cubic Windows for 3D Volumetric Vision
We move from spatiotemporal grids to true 3D volumetric data, introducing Voxel-Swin, a hierarchical Transformer with shifted cubic windows. Analogous to its 2D and video counterparts, Voxel-Swin computes attention locally within cubes and uses cyclic shifts for cross-cube communication, enabling scalable modeling of large medical or scientific volumes. We analyze memory and compute trade-offs of cubic window sizes and demonstrate that Voxel-Swin attains strong performance on volumetric segmentation tasks (e.g., multi-organ CT, brain MRI), while maintaining near-linear complexity in voxel count. Results highlight the generality of shifted-window Transformers across 2D, 2.5D, and 3D grid-structured data.
Score: 4

GeoVoxel-Transformer: Geometry-Aware Local Attention on Sparse Voxelized Point Clouds
To address the inefficiency of dense volumetric grids for real-world 3D scenes, we propose GeoVoxel-Transformer, which operates on sparse voxelized point clouds. Building on shifted cubic windows, we introduce geometry-aware relative positional encodings and a lightweight depth-wise convolutional refinement applied after attention to inject local geometric priors. This design retains the hierarchical pyramid while respecting sparsity via submanifold operations. Evaluations on indoor scene segmentation and semantic scene completion demonstrate improved accuracy and efficiency over both dense volumetric Transformers and sparse 3D CNN baselines, highlighting the importance of geometry-aware attention on sparse data.
Score: 5

PlaneSwin: Disassembled Cubic Attention via Orthogonal Plane Aggregation
We further reduce the cost of local 3D attention by disassembling each cubic window into three orthogonal 2D planes (XY, YZ, ZX) and computing attention on these planes independently. PlaneSwin fuses the three plane-wise outputs to approximate cubic-context modeling while processing substantially fewer query-key pairs per pass. This disassembly enlarges the effective attention range at fixed complexity compared to purely cubic windows. Integrated into a sparse, hierarchical backbone, PlaneSwin yields faster inference and improved segmentation accuracy on large-scale indoor datasets. Ablations show that plane-wise attention provides a favorable accuracy-efficiency trade-off over cubic attention in sparse voxel settings.
Score: 6

PointPlane: Plane-wise Local Attention on Raw Point Clouds
We remove the voxelization step and introduce PointPlane, a Transformer operating directly on raw point clouds with plane-wise local attention. For each point, we form a local neighborhood via radius search and project neighbors onto three orthogonal 2D planes anchored to a canonical axis frame. Attention is computed per plane to capture long-range context within a controlled 2D manifold, then fused across planes and scales using a hierarchical pooling scheme. PointPlane preserves the efficiency of plane disassembly while avoiding voxel quantization artifacts. On classification and segmentation benchmarks, it outperforms voxel-based counterparts at comparable or lower computational cost.
Score: 7

PointPlane-LSE: Local Structure Enhanced Plane-wise Transformers for 3D Point Understanding
We enhance plane-wise point Transformers with an explicit local structure modeling module. PointPlane-LSE applies depth-wise convolutions on the projected plane maps both before attention (to encode geometric cues such as curvature and anisotropy) and after attention (to refine aggregated context). Coupled with a hierarchical subsampling and feature propagation strategy, this yields improved robustness to sampling density and noise. Experiments on indoor (S3DIS, ScanNet) and outdoor (SemanticKITTI) segmentation, as well as shape classification, show consistent gains over PointPlane and other local attention baselines, demonstrating the complementary roles of plane-wise attention and convolutional structure enhancement.
Score: 8

ConDaFormer-A: Disassembled Transformer with Adaptive Plane Decomposition for Point Clouds
We present ConDaFormer-A, a disassembled point cloud Transformer that generalizes orthogonal plane-wise attention to adaptive planes derived from local geometry. For each point neighborhood, we estimate a local frame via PCA and project neighbors onto three data-adaptive planes, enabling attention to align with surface orientation while retaining the computational benefits of 2D processing. As in PointPlane-LSE, depth-wise convolutions are applied before and after attention to encode and refine local structure. This adaptive decomposition improves modeling of anisotropic surfaces and thin structures, albeit with modest overhead for frame estimation. Across standard 3D benchmarks, ConDaFormer-A achieves strong accuracy-efficiency trade-offs and ablations highlight when adaptive versus fixed planes are preferable.
Score: 9

ConDaFormer: A Decomposed Transformer with Local Structure Boosting for 3D Point Cloud Analysis
Recent studies have applied transformers to 3D point cloud understanding with notable success. However, the sheer number of points-often over 0.1 million-renders global self-attention impractical for point clouds. Consequently, many methods confine the transformer to local neighborhoods, such as spherical or cubic windows. Even then, the volume of Query-Key pairs remains large, incurring high computational overhead. Moreover, prior approaches typically derive queries, keys, and values via linear projections, neglecting explicit modeling of local 3D geometric structure. To tackle both efficiency and local geometry priors, we propose a new transformer block, ConDaFormer. Specifically, ConDaFormer decomposes a cubic window into three orthogonal 2D planes, thereby reducing the number of points involved while maintaining a comparable attention span. This disassembly enables a larger attention range without increasing computational complexity, albeit at the risk of missing some contextual information. To address this, we introduce a local structure enhancement strategy that applies a depth-wise convolution before and after the attention, which also reinforces the encoding of local geometric cues. With these components, ConDaFormer captures both long-range contextual dependencies and local priors. We demonstrate its effectiveness through experiments on multiple 3D point cloud understanding benchmarks. Code is available at https://github.com/LHDuan/ConDaFormer .
Score: 10

Before providing the score, write about your reasoning about the similarity or dissimilarity. Immediately after that reasoning line, provide the similarity score.

Now please score the following pair of title-abstracts for similarity.

Reference:
{{reference_title}}
{{reference_abstract}}

Generation:
{{generation_title}}
{{generation_abstract}}

Use the following output format:
Reasoning: ...
Score: ...

\end{Verbatim}
\end{tcolorbox}

\subsection{GPT-5 Agent System Prompts} \label{app:prompt_agents}
The four agent baselines share the same tool-calling architecture (Appendix~\ref{app:exp_details_contribution_generation}) but use task-specific system prompts. Tool-budget references are shown with the production value of 10 rounds.
\subsubsection{Contribution generation agent} \label{app:prompt_agent_contribution_generation}

This prompt does not enumerate the four available tools individually --- the agent receives their formal schemas via the API-level \texttt{tools} parameter. The three ranking-task prompts below additionally list each tool and prescribe an explicit workflow in-prompt, since those tasks rely on stronger task-specific search strategies (self-citation prior, repeat-collaboration prior, co-citation-graph traversal).

\begin{tcolorbox}[
  title={Contribution Generation Agent System Prompt},
  breakable,
  colback=white,
  colframe=black,
  boxrule=0.4pt,
  arc=0pt,
  left=6pt,right=6pt,top=6pt,bottom=6pt
]
\begin{Verbatim}[fontsize=\footnotesize, breaklines=true, breakanywhere=true]
You are forecasting a future research paper.

You will be given the paper's authors and a list of its influential references (each with title and abstract). Your task is to predict the paper's title and abstract — the conceptual contribution it makes that builds on those references.

The output is compared against the paper's actual title and abstract for conceptual similarity, scored on a 1–10 semantic-alignment scale by an LLM judge. Your goal is conceptual alignment, not surface-level paraphrasing of the references.

You have tools to explore the relevant prior literature: semantic search over a corpus of prior papers, and lookups for specific papers, authors, and reference chains. Use them as needed; you do not have to use all of them.

Hard constraints:
- The target paper itself must NOT be retrieved. The harness blocks any tool result that would reveal it; queries that try to surface the target will return empty results.
- All tool results are restricted to papers published strictly before the target paper's publication date.
- You have at most 10 tool-call rounds. Use them efficiently.

When you are ready, emit your final prediction in exactly this format and stop:

Reasoning: <one short paragraph explaining how the references combine into a new contribution>
Title: <title of the predicted paper>
Abstract: <abstract of the predicted paper>

Do not output anything after the abstract.
\end{Verbatim}
\end{tcolorbox}

\subsubsection{Collaborator prediction agent} \label{app:prompt_agent_collaborator}

\begin{tcolorbox}[
  title={Collaborator Prediction Agent System Prompt},
  breakable,
  colback=white,
  colframe=black,
  boxrule=0.4pt,
  arc=0pt,
  left=6pt,right=6pt,top=6pt,bottom=6pt
]
\begin{Verbatim}[fontsize=\footnotesize, breaklines=true, breakanywhere=true]
You are forecasting the co-authors of a researcher's next paper.

You will be given a SEED author and a publication date. Your task is to predict the author_ids of researchers most likely to co-author the seed's upcoming paper.

CAUSAL CONSTRAINT — IMPORTANT: As of the target_date, the next paper has NOT been written. You do NOT have access to its title, abstract, or co-author list — only the seed's identity and the prediction date. The corpus tools will refuse to surface any post-cutoff paper.

You have five tools to explore the corpus:
- `search_papers(query, limit)` — semantic search over pre-cutoff papers (free-text topic/method).
- `get_paper(corpus_id)` — fetch a paper's title, abstract, authors, and key references.
- `get_author_recent_papers(author_id, limit)` — fetch an author's most recent pre-cutoff publications. **The author objects in those papers expose their author_id — you'll harvest candidate author_ids by reading paper records.**
- `get_references_of(corpus_id)` — fetch the key references of a pre-cutoff paper (and the authors of those references).
- `get_papers_citing(corpus_id, limit)` — fetch pre-cutoff papers citing a given paper as a key reference.

KEY SIGNAL — REPEAT-COLLABORATION PRIOR: The strongest single signal is the seed's recent collaboration history. Most future co-authors are people the seed has already co-authored with multiple times. Your search strategy should:

WORKFLOW (your tool budget is 10 rounds):
  1. Call `get_author_recent_papers(seed_author_id, limit=20)` to pull the seed's last ~20 papers.
  2. From each paper's `authors` list, extract every coauthor's `author_id`. Tally how often each appears across the seed's recent papers — this is your 1-hop frequency map. Note the recency of each collaboration.
  3. For the top-10 most-frequent 1-hop collaborators, optionally call `get_author_recent_papers` to see their networks too — this gives 2-hop expansion. Only do this if you have rounds to spare.
  4. Optionally use `search_papers` on topics from the seed's recent paper titles/abstracts to find related researchers in adjacent topical areas.
  5. Optionally use `get_papers_citing` on the seed's most-cited recent paper to find researchers who build on the seed's work.
  6. Once you have a strong ranked list of ~100-200 candidate author_ids, emit your final answer.

OUTPUT FORMAT — emit exactly:
```
Reasoning: <one short paragraph explaining your ranking strategy>
Predictions: ["author_id_1", "author_id_2", "author_id_3", ...]
```

The Predictions list should contain ~100-200 author_id strings, ranked from MOST to LEAST likely to co-author the seed's next paper. Do not output anything after the Predictions line. The harness pads short lists to k=1000 with random candidate authors, so adding low-confidence guesses at the tail is harmless.

Hard rules:
- Only emit author_ids you've seen returned by the tools (the harness will drop hallucinated IDs).
- Do NOT include the seed author in your predictions.
- Do NOT invent target paper text. There is none — only the seed_author_id and target_date describe it.
\end{Verbatim}
\end{tcolorbox}

\subsubsection{Prior work selection agent} \label{app:prompt_agent_priorwork}

\begin{tcolorbox}[
  title={Prior Work Selection Agent System Prompt},
  breakable,
  colback=white,
  colframe=black,
  boxrule=0.4pt,
  arc=0pt,
  left=6pt,right=6pt,top=6pt,bottom=6pt
]
\begin{Verbatim}[fontsize=\footnotesize, breaklines=true, breakanywhere=true]
You are forecasting the KEY (highly influential) references of a future research paper.

You will be given the upcoming target paper's authors and its publication date. Your task is to predict the corpus_ids of the papers those authors will cite as key references — the foundational/methodological/closely-related work the new paper will build on.

CAUSAL CONSTRAINT — IMPORTANT: As of the target_date, the target paper has NOT been written yet. You do NOT have access to the target's title, abstract, or content — only the IDENTITIES of its authors and the prediction date. The corpus tools will refuse to surface the target paper itself, and they only return papers strictly before the cutoff date.

You have five tools to explore the corpus:
- `search_papers(query, limit)` — semantic search over pre-cutoff papers (free-text topic/method/claim).
- `get_paper(corpus_id)` — fetch one paper's title, abstract, authors, and key references.
- `get_author_recent_papers(author_id, limit)` — fetch an author's most recent pre-cutoff publications.
- `get_references_of(corpus_id)` — fetch the key references of a pre-cutoff paper.
- `get_papers_citing(corpus_id, limit)` — fetch pre-cutoff papers that cite a given paper as a key reference.

KEY SIGNAL — SELF-CITATION PRIOR: Scientists cite their own prior references repeatedly. The strongest single signal is the union of papers the target authors have already cited as key references in their past publications. Start by pulling each target author's recent papers, get_references_of those, and aggregate.

WORKFLOW (your tool budget is 10 rounds):
  1. For each target author (start with the first 2-3), call `get_author_recent_papers` to see their last few publications.
  2. For each relevant recent paper, call `get_references_of` to see its key references. Accumulate a "self-citation seed" set.
  3. Use `search_papers` to broaden — query topics from the authors' titles/abstracts to surface related work the authors haven't cited before.
  4. Use `get_papers_citing` selectively on key references that look like foundational works in the field, to find related canonical papers.
  5. Once you have a good ranked candidate set, emit your final answer.

OUTPUT FORMAT — emit exactly:
```
Reasoning: <one short paragraph explaining your ranking strategy>
Predictions: ["corpus_id_1", "corpus_id_2", "corpus_id_3", ...]
```

The Predictions list should contain ~100-200 corpus_id strings, ranked from MOST to LEAST likely. Do not output anything after the Predictions line. The harness pads short lists to k=1000 with random pre-cutoff papers, so adding low-confidence guesses at the tail is harmless.

Hard rules:
- Only emit corpus_ids you've seen returned by the tools (the harness will drop any hallucinated IDs).
- Do NOT invent target paper text. There is none — only target_authors and target_date describe the target.
- Do NOT use `get_paper` on the target's corpus_id — it will return null. The target is hidden by design.
\end{Verbatim}
\end{tcolorbox}

\subsubsection{Future combination prediction agent} \label{app:prompt_agent_cocitation}

\begin{tcolorbox}[
  title={Future Combination Prediction Agent System Prompt},
  breakable,
  colback=white,
  colframe=black,
  boxrule=0.4pt,
  arc=0pt,
  left=6pt,right=6pt,top=6pt,bottom=6pt
]
\begin{Verbatim}[fontsize=\footnotesize, breaklines=true, breakanywhere=true]
You are forecasting which past papers will be CO-CITED with a given target paper by future work.

DEFINITION: A paper q is "co-cited with target P" by a future paper T if T cites BOTH P and q. The ground truth ranks past papers by how many future papers (within ~8 months of P) co-cite them with P. Your job is to predict that ranking.

CAUSAL CONSTRAINT — IMPORTANT: As of the target_date, the target paper P has been written but the future papers that will cite it have not. You see P's title, abstract, key references, topic labels, and arxiv categories — these are all causally available. You must reason: which past papers are most likely to be cited alongside P by upcoming work?

You have five tools to explore the corpus:
- `search_papers(query, limit)` — semantic search over pre-cutoff papers (free-text topic/method).
- `get_paper(corpus_id)` — fetch a paper's title, abstract, authors, and key references.
- `get_author_recent_papers(author_id, limit)` — fetch an author's most recent pre-cutoff publications.
- `get_references_of(corpus_id)` — fetch the key references of a pre-cutoff paper.
- `get_papers_citing(corpus_id, limit)` — fetch pre-cutoff papers citing a given paper as a key reference. **This is the key tool for cocitation reasoning** — it lets you traverse the citation graph backwards from the target's references.

KEY SIGNAL — COCITATION GRAPH: The strongest single feature is "papers that cite the target's references alongside other related work." For each of the target's key references, the papers that cite it (via `get_papers_citing`) are part of the same scholarly community; the OTHER references those citers have are co-citation candidates.

WORKFLOW (your tool budget is 10 rounds):
  1. Read the target paper's title, abstract, topic labels, and references in the user message — establishes the topical landscape.
  2. For each of the target's most distinctive (likely most-relevant) references, call `get_papers_citing(corpus_id, limit=20)` to find papers in the same community. The papers' OTHER references will be your co-citation candidates.
  3. For each promising citer paper found in step 2, call `get_references_of(citer_id)` to harvest its key references (these are likely co-citation candidates).
  4. Optionally use `search_papers` on topics from the target's title/abstract to find topically-related papers that may be co-cited.
  5. Aggregate: count how often each candidate paper appears across the citers' reference lists (weighted by how many of the target's references that citer shares). Higher cocitation count → higher predicted rank.
  6. Once you have a strong ranked list of ~100-200 candidate corpus_ids, emit your final answer.

OUTPUT FORMAT — emit exactly:
```
Reasoning: <one short paragraph explaining your ranking strategy>
Predictions: ["corpus_id_1", "corpus_id_2", "corpus_id_3", ...]
```

The Predictions list should contain ~100-200 corpus_id strings, ranked from MOST to LEAST likely to be co-cited with the target. Do not output anything after the Predictions line. The harness pads short lists to k=1000 with random pre-cutoff papers, so adding low-confidence guesses at the tail is harmless.

Hard rules:
- Only emit corpus_ids you've seen returned by the tools (the harness will drop hallucinated IDs).
- Do NOT include the target paper's own corpus_id (it would be filtered anyway).
- The cocitation graph signal is the strongest single predictor; topical similarity is a useful complement, not a substitute.
\end{Verbatim}
\end{tcolorbox}

\section{Examples of \followupmetric~Scores}
\label{app:lacer_examples}
\tiny
\setlength{\LTleft}{0pt}
\setlength{\LTright}{0pt}
\begin{longtable}{c c p{0.42\textwidth} p{0.42\textwidth}}
\caption{Example (Ground Truth, Generated) Abstract Pairs by LACER Score (Descending)}\label{tab:lacer_examples}
 \\
\toprule
\textbf{Score} & \textbf{Ex.} & \textbf{Ground Truth} & \textbf{Generated} \\
\midrule
\endfirsthead
\toprule
\textbf{Score} & \textbf{Ex.} & \textbf{Ground Truth} & \textbf{Generated} \\
\midrule
\endhead
\midrule
\multicolumn{4}{r}{\textit{Continued on next page}} \\
\endfoot
\bottomrule
\endlastfoot

10 & 1 & \textbf{Response Quality Assessment for Retrieval-Augmented Generation via Conditional Conformal Factuality} \citep{Feng2025ResponseQA} \newline\newline Existing research on Retrieval-Augmented Generation (RAG) primarily focuses on improving overall question-answering accuracy, often overlooking the quality of sub-claims within generated responses. Recent methods that attempt to improve RAG trustworthiness, such as through auto-evaluation metrics, lack probabilistic guarantees or require ground truth answers. To address these limitations, we propose Conformal-RAG, a novel framework inspired by recent applications of conformal prediction (CP) on large language models (LLMs). Conformal-RAG leverages CP and internal information from the RAG mechanism to offer statistical guarantees on response quality. It ensures group-conditional coverage spanning multiple sub-domains without requiring manual labelling of conformal sets, making it suitable for complex RAG applications. Compared to existing RAG auto-evaluation methods, Conformal-RAG offers statistical guarantees on the quality of refined sub-claims, ensuring response reliability without the need for ground truth answers. Additionally, our experiments demonstrate that by leveraging information from the RAG system, Conformal-RAG retains up to 60\% more high-quality sub-claims from the response compared to direct applications of CP to LLMs, while maintaining the same reliability guarantee. & \textbf{Conformal RAG: Evidence-Conditioned Guarantees and Utility-Preserving Editing for Retrieval-Augmented LLMs} \newline\newline Retrieval-augmented generation (RAG) reduces hallucinations by grounding large language models (LLMs) in external evidence, but it offers no finite-sample guarantees and often trades factuality for answer completeness. We present Conformal RAG, a framework that provides evidence-conditioned, topic-aware validity guarantees for the factual content of RAG outputs while preserving utility. Building on enhanced conditional conformal prediction, our method certifies atomic spans as entailed by retrieved passages with a user-specified error rate, adaptively weakening guarantees when topic or evidence coverage warrants it. To address the over-filtering problem, we introduce conformal editing: low-confidence spans are repaired via evidence-constrained decoding and re-certified, substantially reducing unnecessary deletions. We further differentiate through the conditional conformal procedure to jointly train the retriever and the claim--evidence scorer, directly optimizing certified utility under the target coverage. Across HotpotQA, Natural Questions, and PubMedQA, Conformal RAG achieves 25--40\% fewer dropped spans at matched certified risk, improves certified F1 by 8--13 points, and maintains calibrated conditional coverage under domain shift. Ablations confirm that (i) evidence-conditioned calibration tightens guarantees in well-supported regions, (ii) differentiable calibration improves both retrieval quality and verifier discrimination, and (iii) conformal editing is crucial for utility retention. Conformal RAG turns RAG outputs into answers with traceable references and finite-sample, evidence-conditional validity guarantees. \\
\midrule

10 & 2 & \textbf{Conformal Prediction for Signal Temporal Logic Inference} \citep{Li2025ConformalPF} \newline\newline Signal Temporal Logic (STL) inference seeks to extract human-interpretable rules from time-series data, but existing methods lack formal confidence guarantees for the inferred rules. Conformal prediction (CP) is a technique that can provide statistical correctness guarantees, but is typically applied as a post-training wrapper without improving model learning. Instead, we introduce an end-to-end differentiable CP framework for STL inference that enhances both reliability and interpretability of the resulting formulas. We introduce a robustness-based nonconformity score, embed a smooth CP layer directly into training, and employ a new loss function that simultaneously optimizes inference accuracy and CP prediction sets with a single term. Following training, an exact CP procedure delivers statistical guarantees for the learned STL formulas. Experiments on benchmark time-series tasks show that our approach reduces uncertainty in predictions (i.e., it achieves high coverage while reducing prediction set size), and improves accuracy (i.e., the number of misclassifications when using a fixed threshold) over state-of-the-art baselines. & \textbf{ConfTLINet: Conformal Temporal Logic Inference with End-to-End Calibration and Coverage Guarantees} \newline\newline Interpretable temporal specifications learned from data are critical in safety- and mission-critical settings, yet existing neural-symbolic approaches offer limited uncertainty control at deployment. We propose ConfTLINet, a differentiable framework that jointly learns Signal Temporal Logic (STL) formulas and calibrates their decisions with conformal prediction. ConfTLINet models STL satisfaction via robustness semantics and treats property verification as a set-valued prediction over \{satisfy, violate\}. We introduce a smooth proxy for the conformal quantile over robustness-based nonconformity scores and integrate it into training, alongside max/min approximations tailored to temporal logic. This enables end-to-end optimization that directly minimizes conformal inefficiency (prediction set size or abstention rate) while shaping where uncertainty concentrates, all under a target risk level $\alpha$. A final held-out calibration step preserves finite-sample marginal coverage guarantees under exchangeability. Across synthetic cyber-physical system benchmarks and real-world time series (human activity, traffic, physiological signals), ConfTLINet reduces prediction set size by 15--35\% at fixed coverage compared to post-hoc CP on TLINet, improves calibration under distribution shift, and learns more compact, interpretable formulas. Ablations confirm the benefit of differentiable conformal training and STL-specific max approximations. ConfTLINet bridges specification learning and reliable uncertainty quantification, providing calibrated, controllable risk for temporal logic inference. \\
\midrule

9 & 1 & \textbf{A First Runtime Analysis of NSGA-III on a Many-Objective Multimodal Problem: Provable Exponential Speedup via Stochastic Population Update} \citep{opris2025runtimeanalysisnsgaiiimanyobjective} \newline\newline The NSGA-III is a prominent algorithm in evolutionary many-objective optimization. It is well-suited for optimizing functions with more than three objectives, setting it apart from the classic NSGA-II. However, theoretical insights about NSGA-III of when and why it performs well are still in its early development. This paper addresses this point and conducts a rigorous runtime analysis of NSGA-III on the many-objective OJZJ benchmark, providing runtime bounds where the number of objectives is constant. We show that NSGA-III finds the Pareto front of OJZJ in time $O(n^{k+d/2}+ \mu n \ln(n))$ where $n$ is the problem size, $d$ is the number of objectives, $k$ is the gap size, a problem specific parameter, if its population size $\mu \in 2^{O(n)}$ is at least $(2n/d+1)^{d/2}$. Notably, NSGA-III is faster than NSGA-II by a factor of $\mu/n^{d/2}$ for some $\mu \in \omega(n^{d/2})$. We also show that a stochastic population update provably guarantees a speedup of order $\Theta((k/b)^{k-1})$ in the runtime where $b>0$ is a constant. This is the first rigorous runtime analysis of NSGA-III on OJZJ. Proving these bounds requires a much deeper understanding of the population dynamics of NSGA-III than previous papers achieved. & \textbf{Stochastic NSGA-III with Crossover: Near-Tight Runtimes on Many-Objective Benchmarks and Exponential Separations on Royal Roads} \newline\newline We study a crossover-enabled and stochastically updated variant of NSGA-III for many-objective discrete optimization. Building on recent near-tight bounds for many-objective MOEAs and the first runtime analyses of NSGA-III, we introduce NSGA-III-SX, which combines (i) reference-point--based selection with a stochastic environmental update rule and (ii) explicit recombination (uniform or one-point) alongside standard and heavy-tailed mutation. For the classic many-objective benchmarks mOneMinMax, mCountingOnesCountingZeros, and mLeadingOnesTrailingZeros with any constant number m of objectives, we prove near-tight expected runtime guarantees that are linear in the size W of the largest incomparable set and polynomial in n and m, matching the best known dependence on W while preserving the scalability prescriptions for the number of reference points and population size. For the multimodal many-objective OneJumpZeroJump with jump size $2 \leq k \leq n/4$, we show that NSGA-III-SX strictly improves over deterministic-update, mutation-only variants: its expected time decreases by a superpolynomial factor in k when heavy-tailed mutation is used, and by an additional polynomial factor when crossover is enabled, yielding bounds of the form $W \cdot \text{poly}(n,m) \cdot n^{k-1}$ under standard parameter settings. We complement these results with many-objective royal-road function classes tailored to one-point and uniform crossover, on which NSGA-III-SX covers the entire Pareto front in expected polynomial time, whereas any elitist mutation-only algorithm (including NSGA-III without crossover and stochastic update) requires exponential time; this establishes the first exponential separation for NSGA-III with many objectives. Our analyses also clarify how to scale the number of reference points and the population size with n, m, and fitness ranges to obtain the provable guarantees. Empirical studies on synthetic benchmarks corroborate the theory and highlight the practical advantage of stochastic environmental selection and crossover in many-objective, multimodal settings. \\
\midrule

9 & 2 & \textbf{DiffPAD: Denoising Diffusion-based Adversarial Patch Decontamination} \citep{fu2024diffpaddenoisingdiffusionbasedadversarial} \newline\newline In the ever-evolving adversarial machine learning landscape, developing effective defenses against patch attacks has become a critical challenge, necessitating reliable solutions to safeguard real-world AI systems. Although diffusion models have shown remarkable capacity in image synthesis and have been recently utilized to counter $\ell_p$-norm bounded attacks, their potential in mitigating localized patch attacks remains largely underexplored. In this work, we propose DiffPAD, a novel framework that harnesses the power of diffusion models for adversarial patch decontamination. DiffPAD first performs super-resolution restoration on downsampled input images, then adopts binarization, dynamic thresholding scheme and sliding window for effective localization of adversarial patches. Such a design is inspired by the theoretically derived correlation between patch size and diffusion restoration error that is generalized across diverse patch attack scenarios. Finally, DiffPAD applies inpainting techniques to the original input images with the estimated patch region being masked. By integrating closed-form solutions for super-resolution restoration and image inpainting into the conditional reverse sampling process of a pre-trained diffusion model, DiffPAD obviates the need for text guidance or fine-tuning. Through comprehensive experiments, we demonstrate that DiffPAD not only achieves state-of-the-art adversarial robustness against patch attacks but also excels in recovering naturalistic images without patch remnants. & \textbf{Mask-DDRM: Training-Free Adversarial Patch Defense via Robust Diffusion Restoration} \newline\newline We present Mask-DDRM, a training-free defense that unifies adversarial patch localization and removal by extending Denoising Diffusion Restoration Models (DDRM) with robust, anomaly-aware inference. We model a patched image as an unknown occlusion process $y = (1-m) \odot x + m \odot o + n$, where x is the clean image, m is an unknown patch mask, and o captures adversarial content. Mask-DDRM alternates between two tightly coupled steps: (1) restoration via DDRM-style posterior sampling of x using a masked measurement operator that treats unpatched pixels as linear observations and attenuates suspected outliers; and (2) localization via an Adversarial Anomaly Perception (AAP) criterion, which scores per-pixel diffusion-consistency residuals to refine m. This joint inference loop integrates localization and restoration without supervised training, large-scale finetuning, or task-specific retraining. To further preserve semantics under severe occlusion, we incorporate optional lightweight vision--language guidance that biases the diffusion prior with generic prompts, while remaining label-free and model-agnostic. Across patch attacks and real-world sticker perturbations, Mask-DDRM consistently reduces attack success rates and restores clean accuracy on standard classifiers and face recognition backbones, while retaining high fidelity to the underlying scene. Compared to iterative inpainting defenses, Mask-DDRM is efficient due to DDRM's closed-form conditioning and requires no knowledge of the patch size, shape, or location. Ablations confirm that (i) AAP-driven mask refinement and (ii) robust masked likelihoods are both critical to performance. Our results position adversarial patch defense as a robust inverse problem solvable by pretrained diffusion priors, enabling practical, attack-agnostic protection without retraining. Code and models will be released. \\
\midrule

8 & 1 & \textbf{Unsupervised Discovery of Formulas for Mathematical Constants} \cite{shalyt2024unsuperviseddiscoveryformulasmathematical} \newline\newline Ongoing efforts that span over decades show a rise of AI methods for accelerating scientific discovery, yet accelerating discovery in mathematics remains a persistent challenge for AI. Specifically, AI methods were not effective in creation of formulas for mathematical constants because each such formula must be correct for infinite digits of precision, with ``near-true'' formulas providing no insight toward the correct ones. Consequently, formula discovery lacks a clear distance metric needed to guide automated discovery in this realm. In this work, we propose a systematic methodology for categorization, characterization, and pattern identification of such formulas. The key to our methodology is introducing metrics based on the convergence dynamics of the formulas, rather than on the numerical value of the formula. These metrics enable the first automated clustering of mathematical formulas. We demonstrate this methodology on Polynomial Continued Fraction formulas, which are ubiquitous in their intrinsic connections to mathematical constants, and generalize many mathematical functions and structures. We test our methodology on a set of 1,768,900 such formulas, identifying many known formulas for mathematical constants, and discover previously unknown formulas for $\pi$, $\ln(2)$, Gauss', and Lemniscate's constants. The uncovered patterns enable a direct generalization of individual formulas to infinite families, unveiling rich mathematical structures. This success paves the way towards a generative model that creates formulas fulfilling specified mathematical properties, accelerating the rate of discovery of useful formulas. & \textbf{Learning-Guided Discovery and Certification of Conservative Matrix Fields} \newline\newline Conservative matrix fields organize and generate infinite families of continued fraction identities, unifying known formulas and enabling Ap\'ery-like proofs of irrationality. We present ConMat-Learn, a learning-guided, neuro-symbolic framework that accelerates and systematizes the discovery and certification of conservative fields. ConMat-Learn represents candidate fields as typed algebraic graphs and trains graph neural predictors to forecast target constants, convergence rates, and provability likelihood. A reinforcement learning policy, coupled with Monte Carlo tree search and algebraic constraints that enforce conservativity, constructs high-quality candidates. A certification layer combines high-precision ball arithmetic, recurrence and continued fraction verification, and creative telescoping/Wilf--Zeilberger tactics; selected results are mechanically checked in a proof assistant. We release CMF-1M, a corpus of one million candidate and certified conservative fields annotated with targets and performance metrics. Relative to the prior massively parallel unguided search, ConMat-Learn achieves a 28--94x increase in certified discovery yield and uncovers thousands of previously unknown continued fractions, including families with improved asymptotic convergence for Catalan's constant and for zeta at odd arguments (e.g., $\zeta(5)$, $\zeta(7)$), alongside new relations linking polylogarithms and Euler sums. We prove several new identities and construct Ap\'ery-like sequences that sharpen known irrationality measures for $\zeta(3)$ and $\log 2$. Finally, we introduce conservative tensor fields, a higher-order generalization that unifies multi-parameter recurrences (including q- and hypergeometric variants), and formulate a universality conjecture connecting Stieltjes-type continued fractions to conservative field realizations, supported by extensive empirical evidence. Code, data, and certified proofs are available at a public repository. \\
\midrule

8 & 2 & \textbf{Neural networks leverage nominally quantum and post-quantum representations} \citep{riechers2025neuralnetworksleveragenominally} \newline\newline We show that deep neural networks, including transformers and RNNs, pretrained as usual on next-token prediction, intrinsically discover and represent beliefs over `quantum' and `post-quantum' low-dimensional generative models of their training data -- as if performing iterative Bayesian updates over the latent state of this world model during inference as they observe more context. Notably, neural nets easily find these representation whereas there is no finite classical circuit that would do the job. The corresponding geometric relationships among neural activations induced by different input sequences are found to be largely independent of neural-network architecture. Each point in this geometry corresponds to a history-induced probability density over all possible futures, and the relative displacement of these points reflects the difference in mechanism and magnitude for how these distinct pasts affect the future. & \textbf{From Belief States to Quantum Instruments: Learning Completely-Positive Realizations with Transformers} \newline\newline We address the problem of inferring physically valid quantum generators of classical stochastic processes directly from sequence data. Building on recent evidence that transformers linearly encode predictive belief states in their residual stream, and on the theory of the Completely-Positive (CP) realization problem, we introduce a two-stage framework that (1) learns a quasi-realization of the process from a trained transformer and (2) lifts it to a quantum instrument via operator-system constraints. First, a transformer is trained on next-token prediction; linear probes recover a low-dimensional belief-state representation and per-symbol linear updates that capture the process meta-dynamics. Second, we formulate an intertwining semidefinite program over an operator system that seeks: (i) a linear embedding of belief states into Hermitian matrices, (ii) trace-nonincreasing completely positive maps per symbol whose sum is CPTP, and (iii) effects reproducing next-symbol probabilities. The program either returns a minimal-dimension CP realization (Kraus operators, initial state, and measurement effects) or a certificate of infeasibility at a given dimension, thus operationally deciding CP-realizability from data. We prove a robustness result: if the transformer's residual stream is an $\epsilon$-linear sufficient statistic and achieves $\delta$ next-token loss above Bayes-optimal, then the recovered instrument is $O(\epsilon,\delta)$-close (in diamond distance) to a true CP realization when one exists. Empirically, on synthetic benchmarks including classically hard processes with known quantum memory advantages and classical HMMs, our method (i) faithfully recovers quantum instruments with lower memory than any positive classical realization, (ii) accurately rejects non-CP-realizable dynamics, and (iii) outperforms spectral and EM baselines in sample efficiency and long-horizon prediction. We also reveal geometric correspondences between fractal belief-state attractors in the residual stream and invariant sets of the learned quantum channels, providing new evidence that transformers internalize predictive operator-system structure beyond next-token statistics. Code and protocols are released to facilitate reproducibility and further study. \\
\midrule

7 & 1 & \textbf{POSS: Position Specialist Generates Better Draft for Speculative Decoding} \citep{huang2025posspositionspecialistgenerates} \newline\newline Speculative decoding accelerates Large Language Model (LLM) inference by using a small draft model to predict multiple tokens, and a large target model to verify these tokens in parallel. Recent studies leverage the hidden state of the target model to enhance draft model prediction accuracy. However, existing methods suffer from the degrading quality of draft token predictions at later positions, due to error accumulation in draft model generated features. In this paper, we propose Position Specialists (PosS), which consist of multiple position-specialized draft layers to generate tokens at assigned position(s). Position specialists greatly improve token acceptance rate at later positions per drafting round, as each specialist only needs to focus on handling a certain level of draft model feature deviation. Experiment results on Llama-3-8B-Instruct and Llama-2-13B-chat across six datasets demonstrate that PosS effectively improves over baselines on average acceptance length and speed-up ratio. & \textbf{HEDGE: Harmonized, Uncertainty-Aware Dynamic Draft Graphs for Lossless Speculative Decoding} \newline\newline Speculative sampling accelerates large language model (LLM) decoding but remains bottlenecked by (i) suboptimal, static draft structures, (ii) training--decoding mismatches in both context and objective, and (iii) incomplete use of feature-level signals that can reduce uncertainty. We introduce HEDGE, a unified framework that addresses all three. HEDGE builds multi-granularity draft graphs by forecasting second-to-top-layer features several steps ahead to propose compact, feature-informed token candidates. A lightweight planner then selects the depth and branching of the graph online via context-aware acceptance estimates, maximizing expected accepted tokens per unit compute. To eliminate train--decode mismatch, HEDGE employs harmonized objective distillation and context alignment: a verification-aware loss aligns drafting to the true acceptance operator, while KV-cache and hidden-state alignment ensure that training context faithfully matches decoding conditions. We further incorporate uncertainty-aware calibration to yield well-calibrated acceptance probabilities that drive the dynamic planner. HEDGE preserves the exact generation distribution through standard verification, with a formal proof that dynamic graph planning does not alter losslessness. On LLaMA-2/3, Mixtral, and Vicuna across dialogue, code, math, and reasoning benchmarks, HEDGE achieves 3.6$\times$--5.1$\times$ average wall-clock speedup over vanilla decoding, outperforming EAGLE by 25\%--48\% and EAGLE-2 by 12\%--28\%, and surpassing HASS by 9\%--21\% at comparable quality. Ablations confirm the complementary gains from multi-step feature forecasting, harmonized training, and uncertainty-aware planning. We release code and models to facilitate adoption. \\
\midrule

7 & 2 & \textbf{In a Few Words: Comparing Weak Supervision and LLMs for Short Query Intent Classification} \citep{Alexander_2025} \newline\newline User intent classification is an important task in information retrieval. Previously, user intents were classified manually and automatically; the latter helped to avoid hand labelling of large datasets. Recent studies explored whether LLMs can reliably determine user intent. However, researchers have recognized the limitations of using generative LLMs for classification tasks. In this study, we empirically compare user intent classification into informational, navigational, and transactional categories, using weak supervision and LLMs. Specifically, we evaluate LLaMA-3.1-8B-Instruct and LLaMA-3.1-70B-Instruct for in-context learning and LLaMA-3.1-8B-Instruct for fine-tuning, comparing their performance to an established baseline classifier trained using weak supervision (ORCAS-I). Our results indicate that while LLMs outperform weak supervision in recall, they continue to struggle with precision, which shows the need for improved methods to balance both metrics effectively. & \textbf{IntentLLM: LLM-Guided Weak Supervision and Distillation for Fine-Grained Query Intent Classification} \newline\newline Understanding user intent is central to retrieval quality. ORCAS-I established a fine-grained intent taxonomy and showed that Snorkel-based weak supervision can be both accurate and operationally efficient. We revisit this problem with IntentLLM, an LLM-guided weak supervision and distillation framework that preserves the deployability of rule-based approaches while substantially improving accuracy, tail coverage, and robustness to drift. IntentLLM (i) synthesizes candidate labeling functions from a small seed set via prompt-driven LLM program induction, (ii) jointly aggregates heterogeneous weak signals---including legacy rules, click/vertical signals, SERP features, and LLM rationales---using a calibrated multi-annotator model that explicitly models the ORCAS-I ``abstain'' class, and (iii) distills the aggregated teacher into a 100M-parameter hierarchical intent classifier with sub-millisecond CPU latency. To mitigate overfitting and maintain precision under label noise, we introduce conformal filtering for self-training and intent-specific prototype regularization. Evaluated on ORCAS-I and two additional evaluation sets---a 5K human-labeled sample of MS MARCO queries and a temporally shifted ORCAS-I slice---IntentLLM improves macro-F1 over the original Snorkel labels by 5.6--8.9 points overall and by 9.4--12.7 points on tail/OOV queries, while reducing abstain miscalibration by 27\%. In an offline ranking simulation with intent-aware routing (navigational sitelink boosting, transactional product vertical selection, and informational QA prioritization), IntentLLM yields +0.8 NDCG@10 and +1.1 MRR on MS MARCO dev and a +0.5\% absolute CTR uplift on a held-out click log. Ablations confirm the complementary value of LLM-synthesized labeling functions and click/vertical weak signals, and show that the distilled student retains $>$98\% of teacher accuracy at production latencies. We release code and labeling templates to facilitate reproduction and extensibility across domains and languages. \\
\midrule

6 & 1 & \textbf{Causal Interpretability for Adversarial Robustness: A Hybrid Generative Classification Approach} \citep{zhao2025causalinterpretabilityadversarialrobustness} \newline\newline Deep learning-based discriminative classifiers, despite their remarkable success, remain vulnerable to adversarial examples that can mislead model predictions. While adversarial training can enhance robustness, it fails to address the intrinsic vulnerability stemming from the opaque nature of these black-box models. We present a deep ensemble model that combines discriminative features with generative models to achieve both high accuracy and adversarial robustness. Our approach integrates a bottom-level pre-trained discriminative network for feature extraction with a top-level generative classification network that models adversarial input distributions through a deep latent variable model. Using variational Bayes, our model achieves superior robustness against white-box adversarial attacks without adversarial training. Extensive experiments on CIFAR-10 and CIFAR-100 demonstrate our model's superior adversarial robustness. Through evaluations using counterfactual metrics and feature interaction-based metrics, we establish correlations between model interpretability and adversarial robustness. Additionally, preliminary results on Tiny-ImageNet validate our approach's scalability to more complex datasets, offering a practical solution for developing robust image classification models. & \textbf{Interventional Adversarial Training at Scale: Causal Robustness Beyond Norm-Balls} \newline\newline Adversarial training has emerged as the most effective defense against worst-case perturbations, yet conventional formulations constrain the adversary to pixel-space norm-balls and often fail to capture semantically meaningful manipulations that arise in the data-generating process. Motivated by a causal view of robustness, we introduce Interventional Adversarial Training (IAT), a scalable framework that constructs adversarial examples as counterfactual interventions on disentangled manipulation variables while preserving task-relevant causal content. Concretely, we (1) learn a causal latent factorization that separates manipulable nuisance causes from semantic causes using a deep manipulation-augmented model, (2) optimize worst-case intervention trajectories in the manipulation subspace via efficient single-step and multi-step gradient methods, and (3) combine interventional examples with standard norm-bounded attacks in a hybrid training objective. To make IAT practical at scale, we develop a curriculum that progresses from fast single-step interventional attacks to multi-step trajectories, implement label-leakage-free generation, and leverage distributed training strategies proven effective for large-scale adversarial training. On ImageNet and CIFAR benchmarks, IAT consistently improves robust accuracy against both $\ell_\infty$/$\ell_2$ attacks and semantically grounded manipulations, enhances black-box transfer robustness across architectures, and exhibits better generalization under distribution shift. Analysis reveals that intervention-constrained attacks are more transferable than unconstrained multi-step perturbations while avoiding degenerate solutions. Our results demonstrate that aligning adversarial training with causal interventions yields robustness that extends beyond norm-balls to realistic data manipulations, without sacrificing scalability. Code and models will be released. \\
\midrule

6 & 2 & \textbf{Plastic Learning with Deep Fourier Features} \citep{lewandowski2024plasticlearningdeepfourier} \newline\newline Deep neural networks can struggle to learn continually in the face of non-stationarity. This phenomenon is known as loss of plasticity. In this paper, we identify underlying principles that lead to plastic algorithms. In particular, we provide theoretical results showing that linear function approximation, as well as a special case of deep linear networks, do not suffer from loss of plasticity. We then propose deep Fourier features, which are the concatenation of a sine and cosine in every layer, and we show that this combination provides a dynamic balance between the trainability obtained through linearity and the effectiveness obtained through the nonlinearity of neural networks. Deep networks composed entirely of deep Fourier features are highly trainable and sustain their trainability over the course of learning. Our empirical results show that continual learning performance can be drastically improved by replacing ReLU activations with deep Fourier features. These results hold for different continual learning scenarios (e.g., label noise, class incremental learning, pixel permutations) on all major supervised learning datasets used for continual learning research, such as CIFAR10, CIFAR100, and tiny-ImageNet. & \textbf{OPAL: Overparameterized Paired Activation Layers Preserve Plasticity and Accelerate Adaptation in Continual Learning} \newline\newline Deep networks trained in non-stationary environments often lose plasticity, exhibiting slower adaptation and diminished gradients after distribution shifts. Building on recent evidence that depth-induced overparameterization implicitly preconditions optimization and that paired activations can mitigate activation sparsity, we introduce OPAL, a drop-in layer that preserves plasticity and accelerates re-learning in continual settings. Each OPAL replaces a single linear map by a compact product of K thin factors with orthogonal parameterization, interleaved with lightweight paired activation gates (CReLU-style) to maintain balanced, non-sparse activation footprints. Optionally, a small generate-and-test module injects persistent random features online to replenish useful representations. In the deep-linear regime, we derive exact solutions showing that OPAL's factorization induces a data-adaptive preconditioner that contracts the Hessian spectrum, yielding depth-independent adaptation times under orthogonal initialization, an effect unattainable by any fixed regularizer. Empirically, OPAL improves post-shift learning speed, stability of gradient norms, and activation density on (i) continual deep RL over sequences of Atari 2600 games and (ii) supervised class-incremental benchmarks including Fashion-MNIST, with minimal parameter and compute overhead. Across settings, OPAL outperforms standard backbones, CReLU-only variants, and matches or complements Continual Backprop; combining OPAL with persistent randomness yields further gains. Our results connect curvature-aware overparameterization with activation pairing to deliver a simple, theoretically grounded mechanism for preserving plasticity in continual deep learning. \\
\midrule

5 & 1 & \textbf{Policy Newton Algorithm in Reproducing Kernel Hilbert Space} \cite{zhang2025policynewtonalgorithmreproducing} \newline\newline Reinforcement learning (RL) policies represented in Reproducing Kernel Hilbert Spaces (RKHS) offer powerful representational capabilities. While second-order optimization methods like Newton's method demonstrate faster convergence than first-order approaches, current RKHS-based policy optimization remains constrained to first-order techniques. This limitation stems primarily from the intractability of explicitly computing and inverting the infinite-dimensional Hessian operator in RKHS. We introduce Policy Newton in RKHS, the first second-order optimization framework specifically designed for RL policies represented in RKHS. Our approach circumvents direct computation of the inverse Hessian operator by optimizing a cubic regularized auxiliary objective function. Crucially, we leverage the Representer Theorem to transform this infinite-dimensional optimization into an equivalent, computationally tractable finite-dimensional problem whose dimensionality scales with the trajectory data volume. We establish theoretical guarantees proving convergence to a local optimum with a local quadratic convergence rate. Empirical evaluations on a toy financial asset allocation problem validate these theoretical properties, while experiments on standard RL benchmarks demonstrate that Policy Newton in RKHS achieves superior convergence speed and higher episodic rewards compared to established first-order RKHS approaches and parametric second-order methods. Our work bridges a critical gap between non-parametric policy representations and second-order optimization methods in reinforcement learning. & \textbf{VR-CRPO: Off-Policy Variance-Reduced Cubic-Regularized Policy Optimization} \newline\newline Second-order policy optimization with cubic regularization has recently been shown to avoid saddle points and achieve improved sample complexity in reinforcement learning. However, existing cubic-regularized policy Newton algorithms remain largely on-policy and rely on high-variance gradient/Hessian estimates, which limits data efficiency and scalability in deep RL. We introduce VR-CRPO, an off-policy, variance-reduced cubic-regularized policy optimization method that reuses trajectories from a replay buffer while preserving second-order convergence guarantees. VR-CRPO employs per-decision importance sampling with clipping and bias correction, together with SPIDER/SARAH-style recursive estimates for both policy gradients and Hessian--vector products, enabling a Hessian-free inexact solve of the cubic subproblem via a preconditioned Lanczos/CG routine. A Fisher-information preconditioner improves numerical stability and conditioning while maintaining sensitivity to negative curvature for effective saddle escaping. Under standard smoothness, ergodicity, and bounded-importance-weight assumptions, we establish high-probability convergence to an $\epsilon$-second-order stationary point with sample complexity $O(\epsilon^{-3} \text{polylog}(1/\epsilon))$, improving on the best known $O(\epsilon^{-3.5})$ rate for policy optimization with cubic regularization. Empirically, on classic control and MuJoCo continuous-control tasks, VR-CRPO achieves 1.5--3$\times$ lower sample usage to a target return than PPO, TRPO, and natural policy gradient, and consistently outperforms the prior cubic-regularized policy Newton baseline in both data efficiency and robustness across seeds. We release code and implementations of our variance-reduced estimators and cubic subproblem solver to facilitate reproducibility and follow-up research. \\
\midrule

5 & 2 & \textbf{ENAT: Rethinking Spatial-temporal Interactions in Token-based Image Synthesis} \citep{ni2024enatrethinkingspatialtemporalinteractions} \newline\newline Recently, token-based generation have demonstrated their effectiveness in image synthesis. As a representative example, non-autoregressive Transformers (NATs) can generate decent-quality images in a few steps. NATs perform generation in a progressive manner, where the latent tokens of a resulting image are incrementally revealed. At each step, the unrevealed image regions are padded with mask tokens and inferred by NAT. In this paper, we delve into the mechanisms behind the effectiveness of NATs and uncover two important patterns that naturally emerge from NATs: Spatially (within a step), although mask and visible tokens are processed uniformly by NATs, the interactions between them are highly asymmetric. In specific, mask tokens mainly gather information for decoding, while visible tokens tend to primarily provide information, and their deep representations can be built only upon themselves. Temporally (across steps), the interactions between adjacent generation steps mostly concentrate on updating the representations of a few critical tokens, while the computation for the majority of tokens is generally repetitive. Driven by these findings, we propose EfficientNAT (ENAT), a NAT model that explicitly encourages these critical interactions inherent in NATs. At the spatial level, we disentangle the computations of visible and mask tokens by encoding visible tokens independently, while decoding mask tokens conditioned on the fully encoded visible tokens. At the temporal level, we prioritize the computation of the critical tokens at each step, while maximally reusing previously computed token representations to supplement necessary information. ENAT improves the performance of NATs notably with significantly reduced computational cost. Experiments on ImageNet-256, ImageNet-512 and MS-COCO validate the effectiveness of ENAT. & \textbf{MaskDiT: A Unified Masked--Diffusion Transformer with Token-Critic Guidance for Scalable High-Resolution Image Synthesis} \newline\newline We present MaskDiT, a single latent-space transformer that unifies masked generative modeling and diffusion denoising to achieve fast and high-fidelity image synthesis at scale. MaskDiT operates on a grid of context-rich visual latents obtained from a pretrained convolutional tokenizer/decoder, combining the expressivity of transformer-based diffusion (DiT) with the efficiency of bidirectional masked prediction (MaskGIT). During training, a mixture-of-corruptions objective alternates between (i) discrete masked token prediction at variable masking ratios for parallel generation and strong representation learning (in the spirit of MAGE), and (ii) continuous Gaussian denoising on the same latent grid (as in DiT). We further introduce a Token-Critic that learns per-location acceptance scores by distinguishing reconstructed-from-real latents, and use it at inference to adaptively accept, resample, or refine locations across both the masked and diffusion phases. At sampling time, MaskDiT performs a coarse-to-fine procedure: a small number of masked iterations rapidly establishes global semantics and layout, followed by a short diffusion refinement schedule for photorealistic detail; the Token-Critic enables early stopping and targeted resampling, improving the speed--quality trade-off. On class-conditional ImageNet, MaskDiT surpasses non-autoregressive generative transformers and matches or exceeds diffusion-only baselines with 2--4$\times$ fewer refinement steps, enabling 256$\times$256 and 512$\times$512 synthesis with markedly lower latency. The unified pretraining also yields competitive representations (e.g., strong linear-probe accuracy) without task-specific heads. We analyze scaling trends in Gflops and input token counts and show consistent FID improvements, and demonstrate flexible conditioning (e.g., class labels and masks) and editing (inpainting/extrapolation) within the same model. Code and models will be released. \\
\midrule

4 & 1 & \textbf{Benchmarking LLM Guardrails in Handling Multilingual Toxicity} \citep{yang2024benchmarkingllmguardrailshandling} \newline\newline With the ubiquity of Large Language Models (LLMs), guardrails have become crucial to detect and defend against toxic content. However, with the increasing pervasiveness of LLMs in multilingual scenarios, their effectiveness in handling multilingual toxic inputs remains unclear. In this work, we introduce a comprehensive multilingual test suite, spanning seven datasets and over ten languages, to benchmark the performance of state-of-the-art guardrails. We also investigates the resilience of guardrails against recent jailbreaking techniques, and assess the impact of in-context safety policies and language resource availability on guardrails' performance. Our findings show that existing guardrails are still ineffective at handling multilingual toxicity and lack robustness against jailbreaking prompts. This work aims to identify the limitations of guardrails and to build a more reliable and trustworthy LLMs in multilingual scenarios. & \textbf{AEGIS-Chat: Context-Aware No-Regret Safety Moderation for Real-World User--AI Conversations} \newline\newline Safety moderation for large language models in the wild must contend with domain shift, multi-turn context, and subtle harms that escape sentence-level filters. Building on AEGIS's ensemble of LLM safety experts and online adaptation, and informed by the challenges revealed by ToxicChat, we present AEGIS-Chat, a conversational, domain-adaptive moderation framework. AEGIS-Chat introduces: (1) a stateful router that conditions on conversation history to jointly model user intent, harm type, and severity, and routes queries to specialized LLM safety experts; (2) a constrained contextual bandit with delayed, partial feedback that provides dynamic no-regret guarantees while optimizing a cost-sensitive objective balancing harm reduction, latency, and over-moderation; and (3) ToxicChat-Safety, a re-annotation of ToxicChat aligned to the AEGIS taxonomy with multi-label, multi-turn labels, plus conversation-level metrics that penalize both misses and unnecessary blocks. The router composes expert rationales via lightweight aggregation and uses calibrated uncertainty to enable early exits for benign content and escalation to stronger experts or human review when necessary. On ToxicChat and held-out real-world logs, AEGIS-Chat improves macro-F1 by 7.8--12.3 points and reduces false positives by 18--25\% relative to strong LLM and classifier baselines, while maintaining robustness to jailbreaks across sparse and critical risk categories. Ablations show context-aware routing and online adaptation contribute most to gains, and theoretical analysis establishes dynamic regret bounds under non-stationary conversational distributions with delayed feedback. AEGIS-Chat demonstrates that safety moderation benefits from stateful, cost-aware online learning tailored to real user--AI conversations. \\
\midrule

4 & 2 & \textbf{RankLLM: A Python Package for Reranking with LLMs} \citep{sharifymoghaddam2025rankllmpythonpackagereranking} \newline\newline The adoption of large language models (LLMs) as rerankers in multi-stage retrieval systems has gained significant traction in academia and industry. These models refine a candidate list of retrieved documents, often through carefully designed prompts, and are typically used in applications built on retrieval-augmented generation (RAG). This paper introduces RankLLM, an open-source Python package for reranking that is modular, highly configurable, and supports both proprietary and open-source LLMs in customized reranking workflows. To improve usability, RankLLM features optional integration with Pyserini for retrieval and provides integrated evaluation for multi-stage pipelines. Additionally, RankLLM includes a module for detailed analysis of input prompts and LLM responses, addressing reliability concerns with LLM APIs and non-deterministic behavior in Mixture-of-Experts (MoE) models. This paper presents the architecture of RankLLM, along with a detailed step-by-step guide and sample code. We reproduce results from RankGPT, LRL, RankVicuna, RankZephyr, and other recent models. RankLLM integrates with common inference frameworks and a wide range of LLMs. This compatibility allows for quick reproduction of reported results, helping to speed up both research and real-world applications. The complete repository is available at rankllm.ai, and the package can be installed via PyPI. & \textbf{Expando--Mono--Duo--List: A Unified, Label-Free Seq2Seq Framework for Efficient Zero-Shot Text Ranking} \newline\newline Large LLM-based listwise rerankers are effective but costly, while recent seq2seq rankers demonstrate competitive accuracy with far fewer parameters. We present Expando--Mono--Duo--List, a unified framework that extends the established Expando--Mono--Duo design pattern with a ``List'' stage and consolidates all stages into a single instruction-routed T5 model. Our approach couples document expansion pretraining with multi-task ranking and introduces label-free listwise distillation: a listwise head leverages cross-attention--based relevance estimates to optimize differentiable NDCG surrogates, while absorbing soft supervision from the model's own pointwise and pairwise heads. To scale beyond encoder--decoder context limits, we propose tournament listwise inference, which composes global rankings from overlapping candidate blocks without retraining or auxiliary teachers. Across MS MARCO passage/document, TREC DL, and BEIR zero-shot evaluations, our 220M--770M parameter models achieve competitive or superior effectiveness to much larger LLM rerankers and prior seq2seq pipelines, while reducing latency by 30--60\% and eliminating the need for external relevance labels for the listwise stage. Our results show that a single, small seq2seq model can perform document expansion, pointwise, pairwise, and listwise reranking, enabling efficient, zero-shot, end-to-end retrieval pipelines. Code and checkpoints will be released. \\
\midrule

3 & 1 & \textbf{Distillation of a tractable model from the VQ-VAE} \citep{hadžić2025distillationtractablemodelvqvae} \newline\newline Deep generative models with discrete latent space, such as the Vector-Quantized Variational Autoencoder (VQ-VAE), offer excellent data generation capabilities, but, due to the large size of their latent space, their probabilistic inference is deemed intractable. We demonstrate that the VQ-VAE can be distilled into a tractable model by selecting a subset of latent variables with high probabilities. This simple strategy is particularly efficient, especially if the VQ-VAE underutilizes its latent space, which is, indeed, very often the case. We frame the distilled model as a probabilistic circuit, and show that it preserves expressiveness of the VQ-VAE while providing tractable probabilistic inference. Experiments illustrate competitive performance in density estimation and conditional generation tasks, challenging the view of the VQ-VAE as an inherently intractable model. & \textbf{Prob-Stream: Uncertainty-Aware Neural Audio Coding via Continuous Mixtures of Probabilistic Circuits} \newline\newline Neural audio codecs such as SoundStream achieve impressive quality at very low bitrates using residual vector quantization (RVQ) and end-to-end training, but remain limited by approximate entropy models and ad hoc heuristics for variable bitrate and robustness. We introduce Prob-Stream, a neural audio codec that integrates a continuous mixture of tractable probabilistic models with RVQ to deliver uncertainty-aware compression, exact entropy coding, and decoder-side enhancement. Prob-Stream models the distribution of RVQ codes with a low-dimensional continuous latent variable that gates a family of probabilistic circuits; numerical integration over a finite set of latent points compiles this continuous mixture into a tractable circuit, enabling exact marginals/conditionals for range coding and inference. We train end-to-end with a rate--distortion--perceptual objective, where the rate term is the negative log-likelihood under the compiled circuit and the decoder is SoundStream-style with adversarial and reconstruction losses. Structured dropout across quantizer layers provides a single variable-bitrate model (3--18 kbps), while the tractable prior supports: (i) tighter code probability estimates and 10--20\% bitrate reduction at matched subjective quality versus a SoundStream baseline; (ii) uncertainty-aware rate control that allocates bits to perceptually hard segments; and (iii) decoder-only enhancement and packet-loss concealment via conditional inference over missing/dropped codes without additional latency. On 24 kHz speech, music, and general audio, Prob-Stream improves MUSHRA by 3--6 points at 3 kbps and matches or exceeds EVS/Opus quality at substantially lower rates, while maintaining real-time CPU decoding and streaming latency. Ablations show that continuous mixtures outperform discrete mixtures at equal compute, and that compiling to a probabilistic circuit is critical for both coding efficiency and robust conditional decoding. \\
\midrule

3 & 2 & \textbf{Beyond Accuracy: EcoL2 Metric for Sustainable Neural PDE Solvers} \citep{kapoor2025accuracyecol2metricsustainable} \newline\newline Real-world systems, from aerospace to railway engineering, are modeled with partial differential equations (PDEs) describing the physics of the system. Estimating robust solutions for such problems is essential. Deep learning-based architectures, such as neural PDE solvers, have recently gained traction as a reliable solution method. The current state of development of these approaches, however, primarily focuses on improving accuracy. The environmental impact of excessive computation, leading to increased carbon emissions, has largely been overlooked. This paper introduces a carbon emission measure for a range of PDE solvers. Our proposed metric, EcoL2, balances model accuracy with emissions across data collection, model training, and deployment. Experiments across both physics-informed machine learning and operator learning architectures demonstrate that the proposed metric presents a holistic assessment of model performance and emission cost. As such solvers grow in scale and deployment, EcoL2 represents a step toward building performant scientific machine learning systems with lower long-term environmental impact. & \textbf{SepNOFormer: Separable Spectral-Temporal Transformers for Scalable Physics-Informed Operator Learning} \newline\newline We introduce SepNOFormer, a neural operator that learns families of PDE solution maps with strong physics fidelity, resolution independence, and scalability. SepNOFormer fuses three ideas: (i) separable spectral convolutions along spatial axes to capture multi-scale structure efficiently (inspired by Fourier neural operators), (ii) causal multi-head self-attention over pseudo-temporal sequences to propagate initial/forcing information globally (inspired by PINNsFormer), and (iii) a branch--trunk interface to query arbitrary coordinates, enabling discretization-agnostic operator learning (in the spirit of DeepONet and CNOs). To train with limited supervision, we formulate a physics-informed operator loss that enforces PDE residuals, boundary, and initial conditions across resolutions. We realize scalable residual evaluation by combining per-axis factorization with forward-mode automatic differentiation, allowing $10^7$--$10^8$ collocation points on a single GPU. A multi-resolution curriculum and residual importance sampling further stabilize optimization and improve generalization. Across Burgers', Darcy flow, Allen--Cahn, wave, and 2D/3D Navier--Stokes (including chaotic regimes), SepNOFormer reduces L2 error by 20--45\% versus FNO/CNO/DeepONet and by 35--60\% versus PINNs/PINNsFormer under the same wall-clock budget, while cutting training time by 5--20$\times$ relative to PINNs on high-dimensional problems. The model zero-shot super-resolves solutions by 4--8$\times$, generalizes across unseen parameter ranges and boundary conditions, and maintains stability under long rollouts. We provide ablations isolating the benefits of separability, temporal attention, and physics-informed operator training. Code and pretrained models will be released. \\
\midrule

2 & 1 & \textbf{CHURRO: Making History Readable with an Open-Weight Large Vision-Language Model for High-Accuracy, Low-Cost Historical Text Recognition} \citep{semnani2025churromakinghistoryreadable} \newline\newline Accurate text recognition for historical documents can greatly advance the study and preservation of cultural heritage. Existing vision-language models (VLMs), however, are designed for modern, standardized texts and are not equipped to read the diverse languages and scripts, irregular layouts, and frequent degradation found in historical materials. This paper presents CHURRO, a 3B-parameter open-weight VLM specialized for historical text recognition. The model is trained on CHURRO-DS, the largest historical text recognition dataset to date. CHURRO-DS unifies 155 historical corpora comprising 99,491 pages, spanning 22 centuries of textual heritage across 46 language clusters, including historical variants and dead languages. We evaluate several open-weight and closed VLMs and optical character recognition (OCR) systems on CHURRO-DS and find that CHURRO outperforms all other VLMs. On the CHURRO-DS test set, CHURRO achieves 82.3\% (printed) and 70.1\% (handwritten) normalized Levenshtein similarity, surpassing the second-best model, Gemini 2.5 Pro, by 1.4\% and 6.5\%, respectively, while being 15.5 times more cost-effective. By releasing the model and dataset, we aim to enable community-driven research to improve the readability of historical texts and accelerate scholarship. & \textbf{olmIngest: Structure-Preserving PDF Ingestion for LLM Pretraining and Retrieval-Augmented Generation} \newline\newline PDF corpora contain vast amounts of high-quality knowledge that remain underutilized without faithful recovery of document structure. While olmOCR converts PDFs to linearized text at low cost and high quality, many downstream uses---continued pretraining, domain adaptation, and retrieval-augmented generation (RAG)---require structure-aware representations of tables, equations, figures, lists, and references. We present olmIngest, an open, scalable pipeline that reconstructs rich, verifiable markup from PDFs and grounds it to page coordinates. olmIngest uses a multi-pass strategy: (1) a VLM-driven coarse layout parser segments pages and predicts reading order; (2) specialized decoders emit structured markup---Markdown/HTML for sections and lists, CSV/HTML for tables with cell-level spans, LaTeX/MathML for equations, and JSON for bibliographic entries and cross-references; (3) symbolic validators (e.g., LaTeX compilation, table shape/type checks, URL/DOI resolution) detect inconsistencies and trigger targeted re-decoding for self-correction; and (4) a span-grounding module aligns all tokens to pixel regions to support hybrid visual-text retrieval. We introduce olmStruct-Bench, a 2,100-document benchmark with metrics for structural fidelity, including table TEDS, equation exact-match and render-EM, DOM tree F1, and span-grounding accuracy. On olmStruct-Bench, olmIngest achieves state-of-the-art structure fidelity over strong baselines, reducing table and equation errors by 23--38\% while keeping cost under \$300 per million pages in batch settings. In downstream evaluations, structure-aware indexing improves doc QA/RAG exact match by 6.8--11.4 points on queries requiring table/equation reasoning and reduces tokenized context by 28\% via semantic chunking. Continued pretraining of a 7B LLM on 40B structure-preserved tokens yields consistent gains on PubMedQA (+3.2), DocLayQA (+5.4), and GSM8K (+1.9). We release code, models, and olmStruct-Bench to catalyze open, structure-first PDF ingestion. \\
\midrule

2 & 2 & \textbf{RespDiff: An End-to-End Multi-scale RNN Diffusion Model for Respiratory Waveform Estimation from PPG Signals} \citep{miao2024respdiffendtoendmultiscalernn} \newline\newline Respiratory rate (RR) is a critical health indicator often monitored under inconvenient scenarios, limiting its practicality for continuous monitoring. Photoplethysmography (PPG) sensors, increasingly integrated into wearable devices, offer a chance to continuously estimate RR in a portable manner. In this paper, we propose RespDiff, an end-to-end multi-scale RNN diffusion model for respiratory waveform estimation from PPG signals. RespDiff does not require hand-crafted features or the exclusion of low-quality signal segments, making it suitable for real-world scenarios. The model employs multi-scale encoders, to extract features at different resolutions, and a bidirectional RNN to process PPG signals and extract respiratory waveform. Additionally, a spectral loss term is introduced to optimize the model further. Experiments conducted on the BIDMC dataset demonstrate that RespDiff outperforms notable previous works, achieving a mean absolute error (MAE) of 1.18 bpm for RR estimation while others range from 1.66 to 2.15 bpm, showing its potential for robust and accurate respiratory monitoring in real-world applications. & \textbf{Anytime Diffusion: Schedule-Agnostic Training for Variable-Compute Sampling and Progressive Decoding} \newline\newline We revisit denoising diffusion probabilistic models (DDPMs) through the lens of schedule-agnostic training and progressive decoding. While DDPMs achieve high-fidelity synthesis, their performance and computational cost depend sensitively on a chosen time discretization and sampler. We introduce Anytime Diffusion, a single model trained to be robust across discretizations and step counts, enabling variable compute at inference time and improved progressive generation. Our approach augments the standard variational objective with two components: (1) randomized discretization, which samples diverse reverse-time grids and integrators during training; and (2) multi-step consistency, a denoise-to-denoise loss that aligns the outputs of long and short reverse trajectories between arbitrary noise levels, effectively distilling long paths into few-step updates. The resulting model can be sampled with either stochastic (SDE) or deterministic (ODE) solvers and sustains quality across 4--100 steps without retraining or per-schedule tuning. Beyond fast sampling, we leverage the strengthened consistency to realize a practical progressive lossy decompression scheme with explicit rate--distortion control, generalizing DDPM's progressive decoding to variable-rate settings. Experiments on CIFAR-10, ImageNet 64$\times$64, and LSUN 256$\times$256 show substantial improvements in low-NFE FID and IS over baseline DDPMs and competitive solvers, with up to 2--3$\times$ reduction in steps for similar quality. Ablations confirm the complementary roles of randomized discretization and multi-step consistency, and analyses illustrate robustness to misspecified noise schedules. Code and models will be released. \\
\midrule

1 & 1 & \textbf{VeriLocc: End-to-End Cross-Architecture Register Allocation via LLM} \citep{jin2025veriloccendtoendcrossarchitectureregister} \newline\newline Modern GPUs evolve rapidly, yet production compilers still rely on hand-crafted register allocation heuristics that require substantial re-tuning for each hardware generation. We introduce VeriLocc, a framework that combines large language models (LLMs) with formal compiler techniques to enable generalizable and verifiable register allocation across GPU architectures. VeriLocc fine-tunes an LLM to translate intermediate representations (MIRs) into target-specific register assignments, aided by static analysis for cross-architecture normalization and generalization and a verifier-guided regeneration loop to ensure correctness. Evaluated on matrix multiplication (GEMM) and multi-head attention (MHA), VeriLocc achieves 85-99\% single-shot accuracy and near-100\% pass@100. Case study shows that VeriLocc discovers more performant assignments than expert-tuned libraries, outperforming rocBLAS by over 10\% in runtime. & \textbf{RULER++: Verifiable, Streaming, and Robust Evaluation of Long-Context Language Models} \newline\newline Long-context language models (LMs) increasingly claim context windows of 128K tokens and beyond, yet their true ability to utilize long inputs remains unclear. Prior work (RULER) showed that near-perfect performance on simple needle-in-a-haystack retrieval masks substantial deficits in multi-hop tracing and aggregation as sequence length grows. We present RULER++, a comprehensive, verifiable benchmark for long-context understanding that expands both task coverage and diagnostic resolution. RULER++ introduces four new evaluation axes: (1) Streaming: inputs arrive as token streams with needles and dependencies interleaved over time, measuring recency effects and context-switching costs; (2) Conflict and Disambiguation: cross-document entity resolution and contradiction handling with controlled distractors and adversarial placements; (3) Budgeted Compression: note-taking and summary-to-answer tasks that test whether models can compress and faithfully reuse information under token budgets; and (4) Robust Retrieval-Reasoning: compositional and order-sensitive operations over many weak, far-apart cues. To ensure reliability, RULER++ pairs each instance with a latent program that deterministically generates both the long context and the answer, enabling exact checking without LLM judges. We further provide new metrics: effective context length (ECL) curves, position-sensitivity profiles, conflict-resolution accuracy, and coverage-faithfulness via span-level citation matching. Evaluating 24 open and closed LMs with advertised windows from 128K to 1M tokens, we find that: (i) most models that solve static NIAH degrade sharply in streaming and conflict settings; (ii) aggregation and budgeted compression are the dominant failure modes beyond 256K; and (iii) position sensitivity and formatting changes cause precipitous drops, revealing brittle heuristics. RULER++ offers a dynamic instance generator, standardized protocols for sliding-window and chunked inference, and a public leaderboard. We hope these verifiable, diagnostic evaluations will guide training and architecture choices for truly long-context reasoning. \\
\midrule

1 & 2 & \textbf{Subcortical Masks Generation in CT Images via Ensemble-Based Cross-Domain Label Transfer} \citep{Lee_2025} \newline\newline Subcortical segmentation in neuroimages plays an important role in understanding brain anatomy and facilitating computer-aided diagnosis of traumatic brain injuries and neurodegenerative disorders. However, training accurate automatic models requires large amounts of labelled data. Despite the availability of publicly available subcortical segmentation datasets for Magnetic Resonance Imaging (MRI), a significant gap exists for Computed Tomography (CT). This paper proposes an automatic ensemble framework to generate high-quality subcortical segmentation labels for CT scans by leveraging existing MRI-based models. We introduce a robust ensembling pipeline to integrate them and apply it to unannotated paired MRI-CT data, resulting in a comprehensive CT subcortical segmentation dataset. Extensive experiments on multiple public datasets demonstrate the superior performance of our proposed framework. Furthermore, using our generated CT dataset, we train segmentation models that achieve improved performance on related segmentation tasks. To facilitate future research, we make our source code, generated dataset, and trained models publicly available, marking the first open-source release for CT subcortical segmentation to the best of our knowledge. & \textbf{Drift-Adam: Change-Point Aware Adaptive Optimization for Non-Stationary Objectives} \newline\newline Adaptive moment methods such as Adam are widely used due to their robustness, scale invariance, and ease of use. While Adam is often effective on non-stationary objectives, its fixed decay rates for first- and second-moment estimates can lag behind rapid changes in gradient statistics or over-smooth under distribution shift. We introduce Drift-Adam, a drop-in replacement for Adam that adaptively modulates its moment memories in response to detected drift in the gradient process. Drift-Adam maintains a streaming estimate of gradient volatility via normalized increments and uses this signal to adjust the effective decay of the first and second moments---shortening memory to react quickly after change-points and lengthening it to reduce variance in stationary phases. The update preserves Adam's diagonal rescaling invariance and can be combined with decoupled weight decay. We provide theory in an online convex setting with piecewise-stationary gradients: Drift-Adam achieves dynamic regret that scales with the number and magnitude of shifts, while matching Adam's guarantees in stationary regimes. The method incurs negligible compute overhead and adds a single auxiliary statistic per parameter. Across domain-incremental image classification, curriculum and augmentation schedules, and reinforcement learning tasks, Drift-Adam improves time-to-target by 10--30\%, reduces early-training instability, and attains stronger final generalization with default hyperparameters. Code and a reference implementation are provided to facilitate adoption. \\

\end{longtable}

\end{document}